\documentclass{article}

 \usepackage[final]{neurips_2025}

\usepackage{listings}
\lstdefinestyle{tt}{
  basicstyle=\ttfamily\small,
  columns=flexible,
  frame=single,
  breaklines=true,
  postbreak=\mbox{$\hookrightarrow$\space},
}

\usepackage[utf8]{inputenc} 
\usepackage[T1]{fontenc}    
\usepackage[hidelinks]{hyperref}       
\usepackage{url}            
\usepackage{booktabs}       
\usepackage{amsfonts}       
\usepackage{nicefrac}       
\usepackage{microtype}      
\usepackage{xcolor}         
\usepackage{enumitem}
\usepackage{graphicx} 
\usepackage{multicol}
\usepackage{multirow}
\usepackage{adjustbox}
\usepackage{wrapfig}
\usepackage{tabularx}
\usepackage{geometry}
\usepackage{pdflscape}
\usepackage{longtable}
\usepackage{appendix}

\newcommand{\set}[1]{\lbrace#1\rbrace} 
\newcommand{\simmodel}{\mathrel|\joinrel\sim}

\usepackage{accents}
\newcommand{\vect}[1]{\accentset{\rightharpoonup}{#1}}


\title{Generics and Default Reasoning in Large Language Models}
\author{%
  James Ravi Kirkpatrick$^{1, 2}$ \quad Rachel Katharine Sterken$^{3}$\\
  $^{1}$University of Oxford, Oxford, UK\\ 
  $^{2}$Magdalen College, Oxford, UK\\ 
  $^{3}$University of Hong Kong, Hong Kong, China\\
  \texttt{james.kirkpatrick@philosophy.ox.ac.uk} \\
  \texttt{sterkenr@hku.hk} \\
}

\begin{document}
\raggedbottom
\maketitle

\begin{abstract}
This paper evaluates the capabilities of 28 large language models (LLMs) to reason with 20 defeasible reasoning patterns involving generic generalizations (e.g., \textit{Birds fly}, \textit{Ravens are black}) central to non-monotonic logic. Generics are of special interest to linguists, philosophers, logicians, and cognitive scientists because of their complex exception-permitting behaviour and their centrality to default reasoning, cognition, and concept acquisition. We find that while several frontier models handle many default reasoning problems well, performance varies widely across models and prompting styles. Few-shot prompting modestly improves performance for some models, but chain-of-thought (CoT) prompting often leads to serious performance degradation (mean accuracy drop -11.14\%, SD 15.74\% in models performing above 75\% accuracy in zero-shot condition, temperature 0). Most models either struggle to distinguish between defeasible and deductive inference or misinterpret generics as universal statements. These findings underscore both the promise and limits of current LLMs for default reasoning.
\vspace{0.2cm}

\end{abstract}

\section{Introduction} \label{sec:intro}

Default reasoning and generic generalizations are central to human cognition. Default reasoning is a form of non-monotonic reasoning: it allows us to draw plausible, defeasible conclusions on the basis of general rules that may have exceptions in situations involving incomplete or uncertain information. It is non-monotonic in the sense that additional information may force us to withdraw earlier conclusions by showing us that the case is exceptional. Default reasoning often relies on generic generalizations (e.g., \textit{Birds fly}, \textit{Ravens are black}) to express defaults, generalizations which express exception-tolerant, law-like regularities. For example, \textit{Birds fly} roughly expresses that birds typically fly.  From this, together with the information \textit{Tweety is a bird}, one may infer \textit{Tweety flies}, though this conclusion should be withdrawn upon learning \textit{Tweety is a penguin} and \textit{Penguins are birds that don't fly}. Such patterns of reasoning underpin a range of cognitive capacities and concepts, including \emph{communication} \citep{asher2003logics, asher2013strategic}, \emph{practical reasoning} \citep{dancy1993moral, dancy2004ethics, horty1994moral}, and \emph{defeasible obligation} \citep{ross1930right, ross1939foundations, nute1997defeasible}.

Default reasoning and generic generalizations have thus been of considerable interest to philosophers \citep[e.g.,][]{asher1991commonsense, asher1995what, asher2013more, pelletier1997generics, leslie2007generics, leslie2008generics, nickel2009generics, nickel2016logic, liebesman2011simple, sterken2015generics}, linguists \citep[e.g.,][]{carlson1977reference, krifka1995genericity, cohen1999generics, cohen2004generics, chierchia1998reference, greenberg2003manifestations, mari2013genericity}, cognitive scientists \citep[e.g.,][]{gelman2000role, hollander2002children, gelman2003essential,  gelman2010generics, prasada2006principled, prasada2009representation, khemlani2009generics, haward2018development, tessler2019language}, and computer scientists \citep[e.g.,][]{mccarthy1980circumscription, mccarthy1986applications, reiter1980logic, moore1984possibleworld, moore1985semantical}. This has led to sophisticated theories of genericity and default reasoning.

Recent advances in natural language processing (NLP) and Artificial Intelligence (AI) has been driven by interest in large language models (LLMs), such as GPT-4 \citep{openai2023gpt4},  Claude 3 \citep{anthropic2024claude}, and Gemini Ultra \citep{team2024gemini}. Trained on large text corpora, these models capture statistical distributions in language, and can generate grammatical, fluent, and seemingly meaningful text. Advancements in computational scaling and training has led to striking gains in the performance of tasks \citep{brown2020language, chen2021evaluating, chowdhery2022palm}, including mathematical reasoning \citep{lewkowycz2022solving, wang2023selfconsistency}, commonsense reasoning \citep{jung2022maieutic, liu2022rainier}, and symbolic reasoning \citep{zhou2022leasttomost, khot2022decomposed}.

\begin{figure}[t!]
\centering
\includegraphics[width=0.7\textwidth]{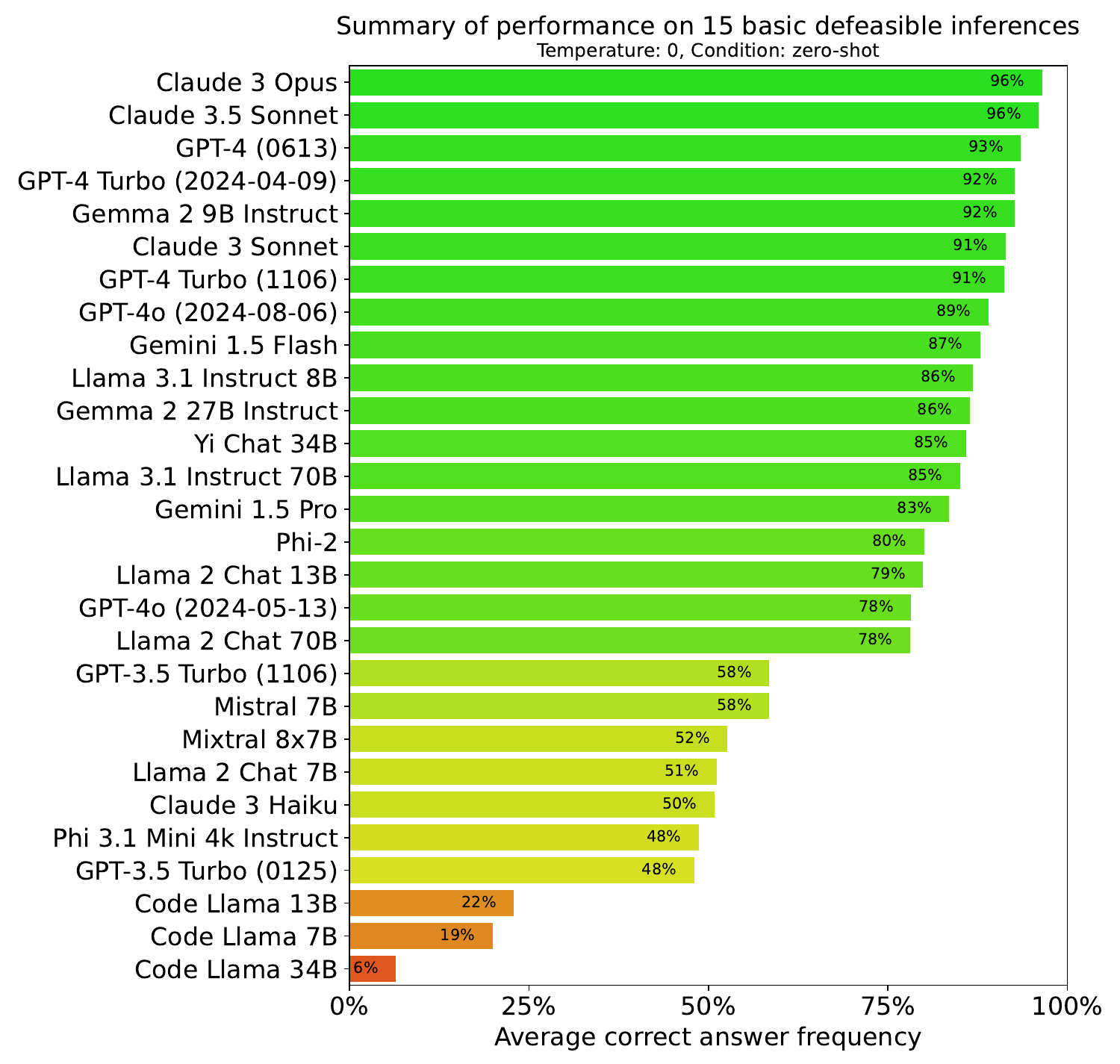}

\caption{Summary of performance of models on 15 uncontroversial default reasoning patterns discussed in \S\ref{sec:discussion} at temperature 0. Guessing accuracy is at 50\%.
}
\label{fig:1}
\end{figure}

However, the capacity of LLMs to handle default reasoning involving generics has been relatively underexplored. This paper addresses this gap by evaluating 28 LLMs for 20 benchmark problems for non-monotonic reasoning. Consider the default reasoning pattern \textit{Defeasible Modus Tollens} (DMT): \textit{$F$s are $G$. $a$ is not $G$. Therefore, $a$ is not $F$}. To test whether LLMs endorse this inference, we prompted models with instances like:

\begin{quote}
    \textbf{User prompt}: From `Adults have driving licenses', together with `Mary doesn't have a driving license', can we defeasibly infer `Mary isn't an adult'? (\textbf{System prompt}: Answer only with `yes' or `no' and nothing else.)\vspace{0.1cm}

    \noindent\textbf{GPT-4}: Yes; \textbf{Mixtral 8x7B}: No; etc.
\end{quote}
Each model was tested on multiple instances of this pattern, and performance was scored for accuracy relative to the defeasible inference. Figure \ref{fig:1} summarizes average performance of models across several inference patterns to be discussed below, while Table \ref{table:1} compares model performance in the zero-shot, few-shot, and zero-shot chain-of-thought conditions. We also test how LLMs use generics in logically deductive inferences and whether they treat generics as universal generalizations, providing insight into how such information is internally represented.

We find that several frontier models perform impressively high on our set of default reasoning patterns, although results vary considerably across models. We also find that the high performance of frontier models is not uniformly robust across prompting strategies, with some performance improvement with few-shot prompting, but significant performance degradation with chain-of-thought (CoT) prompting (mean accuracy drop -11.14\%, SD 15.74\% in models performing above 75\% accuracy in zero-shot condition, temp=0). Finally, we find that most models either fail to recognize the logical distinction between defeasible and deductive reasoning contexts or treat generics as universals. These findings highlight difficulties around default reasoning, as well as the interpretation of generic generalizations.

In sum, our main contributions are:

\begin{itemize}[parsep=0pt]
\item we show that there is considerable variation in performance with default reasoning tasks involving generic generalizations across models, default reasoning patterns, and prompting strategies;
\item we find significant performance degradation in frontier model performance with chain-of-thought (CoT) prompting (mean accuracy drop -11.14\%, SD 15.74\% in models performing above 75\% accuracy in zero-shot condition, temperature 0);
\item we find that language models struggle to distinguish between defeasible and deductive reasoning contexts, as well as between generic and universal generalizations.
\end{itemize}

\section{Background and Related Work} \label{sec:default}

\subsection{Default Reasoning}

In the AI literature, default reasoning has usually been studied under the term `non-monotonic reasoning' (\citealt{mccarthy1980circumscription, mcdermott1980nonmonotonic, reiter1980logic, reiter1987nonmonotonic, moore1985semantical}; see \citet{ginsberg1987readings}, for an overview). Formally, a system is \emph{non-monotonic} if it violates the monotonicity property that governs classically deductive reasoning:

\begin{description}
\item [Monotonicity.] If $\Gamma \simmodel \phi$, then $\Gamma \cup \Delta \simmodel \phi$,
\end{description}
where $\Gamma, \Delta$ are sets of sentences, $\phi$ is a sentence, and $\simmodel$ is an entailment relation. Under Monotonicity, conclusions are robust under information addition: once $\phi$ follows from $\Gamma$, it follows from any superset of $\Gamma$. Non-monotonic frameworks, by contrast, allow for \emph{retraction}: $\phi$ may cease to follow in light of new information. While `non-monotonic reasoning' refers to this feature, we prefer `default reasoning' to remain agnostic about the underlying formal architecture of reasoning. Our focus is on informal reasoning patterns: drawing defeasible conclusions from incomplete information.

Default reasoning should be distinguished from \textit{commonsense reasoning}, which draws on broad background knowledge (e.g., spatial relational, causal, scientific, social) to draw defeasible inferences, like inferring from \textit{John put his hand in fire} that \textit{John's hand is hurt} \citep{speer2018conceptnet}. Although such inferences are often defeasible, commonsense reasoning draws on rich worldly knowledge not given by the premise set. In contrast, we target \emph{pure} default reasoning: inferences drawn strictly from given premises, isolating reasoning patterns for controlled evaluation. To this end, we adapt the experimental design of \citet{holliday2024conditional}, who examine the capacities of LLMs to distinguish between pure logically correct and logically incorrect reasoning patterns involving modals and conditionals.

Research in non-monotonic reasoning has identified many inference patterns that are defeasibly acceptable but deductively invalid, often using generic generalizations to capture default rules tolerant of exceptions (as in Defeasible Modus Tollens introduced in \S\ref{sec:intro}). AI researchers developed non-monotonic logics to define a notion of consequence aimed at capturing these inferences, grounding a symbolic approach to default reasoning \citep{mccarthy1980circumscription, mcdermott1980nonmonotonic, reiter1980logic, reiter1987nonmonotonic, moore1985semantical}, although this notion of consequence should not be confused with classical validity.
Furthermore, while these inference patterns are widely accepted in AI, their psychological status remains debated and human subjects often resist default inferences about individuals known to violate other default rules \citep{elio1993human, pelletier1997what, pelletier2005case}. In this study, we test LLMs on a broad set of such patterns, from the widely accepted to the contested.
We evaluate performance against expert judgments rather than human data, without taking a stand on their normative status.

\subsection{Generic Generalizations} \label{sec:generics}

Linguists distinguish two types of generic sentences: \emph{reference-to-kind} generics, where genericity comes from the determiner phrase as in \ref{ex:rtf}, and \emph{characterizing} generics, where genericity is a feature of the whole sentence as in \ref{ex:cs}.

\begin{enumerate}[label=(\arabic*)]
\item \label{ex:rtf} 
	\begin{enumerate}[label=\alph*.]
	\item The potato was first cultivated in South America.
	\item Potatoes were first cultivated in South America.
	\end{enumerate}
\item \label{ex:cs}
	\begin{enumerate}[label=\alph*.]
	\item Potatoes contain vitamin C.
	\item A potato contains vitamin C.
	\item The potato contains vitamin C.
	\item Sean has potato with dinner.
	\end{enumerate}
\end{enumerate}
This paper focuses on characterizing generics, which express generalizations about kinds or their members without overt quantifiers (e.g., \textit{every}, \textit{some}, \textit{normally}, \textit{typically}, etc.).

A hallmark of generics is their tolerance to exceptions; although identifying systematic patterns in how they tolerate exceptions has been proven challenging \citep{carlson1977reference, krifka1995genericity, pelletier1997generics, cohen1999generics, leslie2007generics, leslie2008generics}. Generic generalizations vary in strength, ranging from exceptionless definitional statements (e.g., \textit{Squares have four sides}, \textit{The lion is mammal}), to majority characteristic statements holding of a majority (e.g., \textit{Birds fly}) or within a restricted subclass (e.g., \textit{Ducks lay eggs}), to striking property generics true only of a small number (e.g., \textit{Pit bulls attack children}). Crucially, their truth is not reducible to majority statistics: \textit{Canadians are right-handed} is false despite being true of most Canadians.

Generics in English are expressed with bare plurals (\textit{potatoes}), indefinite singulars (\textit{a potato}), definite singulars (\textit{the potato}), and mass nouns (\textit{potato}), as seen in \ref{ex:cs}. The expression of genericity varies cross-linguistically, though no known language contains a dedicated phonological marker for genericity \citep{krifka1995genericity, dahl1995marking, chierchia1998reference, gerstner-link1998typological, dayal1999bare, dayal2004number, behrens2005genericity, farkas2007article}. Here we focus on default reasoning with English bare plural generics.

Generics subjects may be natural kinds (\textit{birds}), artifacts (\textit{tables}), and social kind (\textit{Quakers}). 
Generics support stative predicates (\textit{contains vitamin C}) and habituals (\textit{fly}), but not episodics (\textit{are outside my tent}). Some predicates, like \textit{fly}, are ambiguous between `dispositional' and `episodic' readings, as well as between different modes of propulsion (self-powered flight vs being transported by air). This yields ``doubly generic'' interpretations, as in like \textit{Birds fly}, that generalize over both kinds and events.

Following \citet{krifka1995genericity}, we treat characterizing generics as having a tripartite structure involving a variable-binding \emph{generic operator} `Gen', a \emph{restrictor clause} defining the domain, and a \emph{matrix clause} expressing the predicated property. For example, \textit{A lion has a tail} can be formalized as:

\begin{enumerate}[label=(\arabic*), resume]
\item $\textrm{Gen}[x]\big(\textrm{lion}(x); \exists y[\textrm{tail}(y) \wedge \textrm{has}(x, y)]\big)$
\end{enumerate}
which roughly means: if $x$ is a lion, then, typically, $x$ has a tail. More generally, the logical form of a generic sentence is as follows:

\begin{enumerate}[label=(\arabic*), resume]
\item $\textrm{Gen}[x_1,\ldots, x_i] \big(\textrm{Restrictor}(x_1,\ldots,x_i);\exists y_1,\ldots, y_j \textrm{Matrix}[(\set{x_1},\ldots,\set{x_i}, y_1,\ldots,y_j)]\big)$
\end{enumerate}
where $\vect{x}_i$ are bound by \textit{Gen}, $\vect{y}_j$ are existentially bound in the matrix, and $\phi[\ldots\set{x_m}\ldots]$ indicates that $x_m$ may occur free in $\phi$. For readability, we abbreviate this to $\ulcorner Gen (\phi,\psi) \urcorner$, where $\phi$ and $\psi$ are the restrictor and matrix clauses respectively, suppressing variables as necessary. Determining the appropriate logical form of a generic involves partitioning linguistic material between the restrictor and matrix. This partitioning is highly context-sensitive; it is shaped by discourse topic and prosody structure \citep{rooth1985association, rooth1995indefinites, krifka1995focus, cohen1999think}. In the absence of such cues, we assume subjects map to the restrictor and verb phrases to the matrix.

\subsection{Related Work}

Recent work on generics and LLMs has focused on their relation to commonsense knowledge \citep{gajbhiye2022modelling, liu2023vera}, training models for generic sentence generation \citep{bhagavatula2023i2d2}, and constructing knowledge bases of generics, focusing on exceptions and instantiations \citep{bhakthavatsalam2020genericskb, allaway2023penguins, allaway2024exceptions}. Other work evaluates LLMs on classical and non-monotonic reasoning \citep{parmeretal2024logicbench, pateletal2024multistep}, or models the relationship between generics and other forms of quantification \citep{ralethe2022generic, collacciani2024quantifying}. Studies also find show that models' reasoning with generics and exemplars is partly non-monotonic  \citep{allaway2024exceptions, leidinger2024llms}.

Our study differs systematically testing how LLMs handle one-step \emph{logical} defeasible inferences with generics, abstracted from background knowledge or multi-step reasoning. Closely related to our work is \citet{allaway2025evaluating}, which assesses 12 LLMs on five default reasoning rules. In contrast, our study covers a larger number of models and default reasoning rules, multiple prompt settings, and the treatment of generics versus universal generalizations in defeasible and deductive inference contexts.

\section{Default Reasoning Benchmark Problems} \label{sec:benchmark}

In this section, we provide an overview of 20 default reasoning benchmark problems adapted from classic examples in the AI literature on non-monotonic reasoning \citep{lifschitz1989benchmark, pelletier1997generics}. These problems cover a range of argument structures, including default inferences with individuals, inferences involving only generics, and inferences with generics as conclusions. We represent them using the \textit{Gen} notation introduced in \S2.2, omitting variables when unnecessary for comprehension. We adopt the following notational conventions: $A, B, \ldots$ are open sentences; $x$ is a variable; $d, e$ are constant names; $\neg$, $\wedge$, $\rightarrow$, $\forall$, and $\Diamond$ stand for negation, conjunction, material implication, universal quantification, and possibility. Substitution is written as $\ulcorner\xi[d/v]\urcorner$ (i.e., the formula that results from replacing all free occurrences of $v$ in $\xi$ with (free) occurrences of $d$). A default inference $\phi_1,\ldots, \phi_n \simmodel \psi$ licenses $\psi$ from premises $\phi_1,\ldots, \phi_n$ to $\psi$; $\Gamma \simmodel \psi_1,\ldots, \psi_n$ abbreviates $\Gamma \simmodel \psi_1, \ldots, \Gamma \simmodel \psi_n$; and $\Gamma \simmodel\ ?$ means no (non-trivial) defeasible conclusion can be drawn.
Inferences are displayed with premises above a horizontal line and conclusions below.

The first pattern is the argument form \textit{Defeasible Modus Ponens}, which is enshrined in the well-known `Tweety' example from the non-monotonic reasoning literature:

\setlength{\columnsep}{0.0cm}
\begin{multicols}{2}
\begin{enumerate}[label=\arabic*.]
\item \textit{Defeasible Modus Ponens (DMP)}\vspace{0.1cm}

$Gen[x](A;B)$\\
$\underline{A[d/x]}$\\
$B[d/x]$

\textit{Example}\vspace{0.1cm}

Birds fly.\\
\underline{Tweety is a bird.}\\
Tweety flies.
\end{enumerate}
\end{multicols}
\noindent The second pattern is \textit{Defeasible Modus Tollens}:

\begin{multicols}{2}
\begin{enumerate}[label=\arabic*., resume]
\setcounter{enumi}{1}
\item \textit{Defeasible Modus Tollens (DMT)}\vspace{0.1cm}

$Gen[x](A;B)$\\
$\underline{\neg B[d/x]}$\\
$\neg A[d/x]$

\textit{Example}\vspace{0.1cm}

Adults have driving licenses.\\
\underline{Mary doesn't have a driving license.}\\
Mary isn't an adult.
\end{enumerate}
\end{multicols}
\noindent We also test whether models respect the \emph{defeasibility} of these inferences (i.e., whether they still draw a conclusion when DMP or DMT is blocked).

\begin{multicols}{2}
\begin{enumerate}[label=\arabic*., resume]
\setcounter{enumi}{2}
\item \textit{The `Defeat' of DMP/DMT (DDMP)}\vspace{0.1cm}

$Gen[x](A;B)$\\
$A[d/x]$\\
$\underline{\neg B[d/x]}$\\
?

\textit{Example}\vspace{0.1cm}

Birds fly.\\
Tweety is a bird.\\
\underline{Tweety doesn't fly.}\\
?
\end{enumerate}
\end{multicols}
\noindent In such cases, the denial of the expected conclusion in the premise set overrides the default inference, rendering it non-inferable as indicated by the question mark `?'.

The fourth pattern is the \textit{Nixon Diamond}, an example where conflict between incompatible defaults yields contradictory conclusions, precluding any inference from being made:

\begin{multicols}{2}
\begin{enumerate}[label=\arabic*., resume]
\setcounter{enumi}{3}
\item \textit{Nixon Diamond (ND)}\vspace{0.1cm}

$A[d/x]$\\
$B[d/x]$\\
$Gen[x](A; C)$\\
$\underline{Gen[x](B; \neg C)}$\\
?

\textit{Example}\vspace{0.1cm}

Nixon is a Quaker.\\
Nixon is a Republican.\\
Quakers are pacifists.\\
\underline{Republicans are not pacifists.}\\
?
\end{enumerate}
\end{multicols}
\noindent A variant of this pattern is the \textit{Extended Nixon Diamond}, where additional premises allow a conclusion to be drawn via indirect default reasoning. This tests for a model's capacity for chaining defaults and resolving conflicts where conclusions converge.

\begin{multicols}{2}
\begin{enumerate}[label=\arabic*., resume]
\setcounter{enumi}{4}
\item \textit{Extended Nixon Diamond (END)}\vspace{0.1cm}

$A[d/x]$\\
$B[d/x]$\\
$Gen[x](A; C)$\\
$Gen[x](B; D)$\\
$Gen[x](C; E)$\\
$\underline{Gen[x](D; E)}$\\
$E[d/x]$

\textit{Example} \vspace{0.1cm}

Nixon is a Quaker.\\
Nixon is a Republican.\\
Quakers are doves.\\
Republicans are hawks.\\
Doves are politically active.\\
\underline{Hawks are politically active.}\\
Nixon is politically active.
\end{enumerate}
\end{multicols}
\noindent The next two patterns show how information about \emph{subkinds} overrides
defaults about their \emph{superkinds}. The first involves exceptionless taxonomic relations.

\begin{multicols}{2}
\begin{enumerate}[label=\arabic*., resume]
\setcounter{enumi}{5}
\item \textit{Penguin Principle (PP)}\vspace{0.1cm}

$Gen[x](A;C)$\\
$Gen[x](B;\neg C)$\\
$\forall x(B \rightarrow A)$\\
$B[d/x]$\\
$A[d/x]$\\
$\overline{\neg C[d/x]}$

\textit{Example}\vspace{0.1cm}

Birds fly.\\
Penguins don't fly.\\
All penguins are birds.\\
Tweety is a penguin.\\
Tweety is a bird.\\
$\overline{\textrm{Tweety doesn't fly.}}$
\end{enumerate}
\end{multicols}
\noindent When the subkind--superkind link is itself a generic rather than a strict universal, the inference is less secure. This gives rise to the following weakened variant:

\begin{multicols}{2}
\begin{enumerate}[label=\arabic*., resume]
\setcounter{enumi}{6}
\item \textit{Weakened Penguin Principle (WPP)}\vspace{0.1cm}

$Gen[x](A;C)$\\
$Gen[x](B;\neg C)$\\
$Gen[x](B;A)$\\
$B[d/x]$\\
$\underline{A[d/x]}$\\
$\neg C[d/x]$

\textit{Example}\vspace{0.1cm}

Adults are employed.\\
College students are not employed.\\
College students are adults.\\
Sam is a college student.\\
Sam is an adult.\\
$\overline{\textrm{Sam is not employed.}}$
\end{enumerate}
\end{multicols}
\noindent We now consider two inference patterns where the \emph{conclusion} itself is a generic.

\begin{multicols}{2}
\begin{enumerate}[label=\arabic*., resume]
\setcounter{enumi}{7}
\item \textit{Defeasible Transitivity (DT)}\vspace{0.1cm}

$Gen(A;B)\\
\underline{Gen(B;C)}\\
Gen(A;C)$

\textit{Example}\vspace{0.1cm}

Sparrows are birds.\\
Birds fly.\\
$\overline{\textrm{Sparrows fly.}}$
\end{enumerate}
\end{multicols}

\begin{multicols}{2}
\begin{enumerate}[label=\arabic*., resume]
\setcounter{enumi}{8}
\item \textit{Defeasible Strengthening the Antecedent (DSA)}\vspace{0.1cm}

$Gen(A;C)\\
\overline{Gen(A \wedge B;C)}$

\textit{Example}\vspace{0.1cm}

\phantom{o}\\
Birds fly.\\
$\overline{\textrm{White birds fly.}}$
\end{enumerate}
\end{multicols}
\noindent The next pattern formalizes the independence of unrelated defaults:

\begin{multicols}{2}
\begin{enumerate}[label=\arabic*., resume]
\setcounter{enumi}{9}
\item \textit{Independence (I)}\vspace{0.1cm}

$Gen[x](A;B)\\
Gen[x](A;C)\\
\underline{A[d/x], \neg B[d/x]}\\
C[d/x]$

\textit{Example}\vspace{0.1cm}

Dogs are hairy.\\
Dogs have four legs.\\
\underline{Fido is a three-legged dog.}\\
Fido is hairy
\end{enumerate}
\end{multicols}
\noindent The next four patterns illustrate the principle that irrelevant information (e.g., exceptions or extraneous information) should not affect inferences drawn about independent and otherwise exceptional individuals. The first pattern presents the simplest case.

\begin{multicols}{2}
\begin{enumerate}[label=\arabic*., resume]
\setcounter{enumi}{10}
\item \textit{Basic Default Reasoning (BDR)}\vspace{0.1cm}

$Gen[x](A;B)\\
A[d/x], A[e/x]\\
\underline{\neg B[d/x]}\\
B[e/x]$

\textit{Example}\vspace{0.1cm}

Heavy blocks are located on the table.\\
Blocks A and B are heavy.\\
\underline{A is not on the table.}\\
B is on the table.
\end{enumerate}
\end{multicols}
\noindent The second pattern extends this with explicit extraneous information.

\begin{multicols}{2}
\begin{enumerate}[label=\arabic*., resume]
\setcounter{enumi}{11}
\item \textit{Default Reasoning with Irrelevant Information (DRII)}\vspace{0.1cm}

$Gen[x](A;B)\\
A[d/x], A[e/x]\\
\underline{\neg B[d/x], C[e/x]} \\
B[e/x]$

\textit{Example}\vspace{0.1cm}

\phantom{o}\\
Heavy blocks are located on the table.\\
Blocks A and B are heavy.\\
\underline{A is not on the table. B is red.}\\
B is on the table.
\end{enumerate}
\end{multicols}
\noindent The third pattern symmetrically extends the first with several defaults running in parallel.

\begin{multicols}{2}
\begin{enumerate}[label=\arabic*., resume]
\setcounter{enumi}{12}
\item \textit{Default Reasoning with Several Defaults (DRSD)}\vspace{0.1cm}

$Gen[x](A;B)\\
Gen[x](A;C)\\
A[d/x], A[e/x]\\
\underline{\neg B[d/x], \neg C[e/x]}\\
B[e/x], C[d/x]$

\textit{Example} \vspace{0.1cm}

\phantom{o}\\
Heavy blocks are located on the table.\\
Heavy blocks are normally red.\\
Blocks A and B are heavy.\\
\underline{A is not on the table. B is not red.}\\
B is on the table. A is red.
\end{enumerate}
\end{multicols}
\noindent The fourth pattern demonstrates that even when an exception is explicitly stated, default reasoning should remain uneffected for other independent individuals.

\begin{multicols}{2}
\begin{enumerate}[label=\arabic*., resume]
\setcounter{enumi}{13}
\item \textit{Default Reasoning with a Disabled Default (DRDD)}\vspace{0.1cm}

$Gen[x](A;B)\\
A[d/x], A[e/x]\\
\underline{\Diamond\neg B[d/x]}\\
B[e/x]$

\textit{Examples}\vspace{0.1cm}

\phantom{o}\\
Blocks A and B are heavy.\\
Heavy blocks are located on the table.\\
\underline{A is possibly an exception to this rule.}\\
B is on the table.
\end{enumerate}
\end{multicols}
\noindent Two additional patterns further test the impact of irrelevant information on reasoning, distinguishing irrelevant facts from irrelevant defaults:

\begin{multicols}{2}
\begin{enumerate}[label=\arabic*., resume]
\setcounter{enumi}{14}
\item \textit{Fact Irrelevance(FI)}\vspace{0.1cm}

$Gen[x](A;B)\\
A[d/x]\\
\underline{C[d/x]}\\
B[d/x]$

\textit{Example}\vspace{0.1cm}

Birds fly.\\
Tweety is a bird.\\
\underline{Tweety is a sparrow}\\
Tweety flies.
\end{enumerate}
\end{multicols}

\begin{multicols}{2}
\begin{enumerate}[label=\arabic*., resume]
\setcounter{enumi}{15}
\item \textit{Default Irrelevance (DI)}\vspace{0.1cm}

$Gen[x](A;B)\\
Gen[x](C;D)\\
\underline{A[d/x]}\\
B[d/x]$

\textit{Example} \vspace{0.1cm}

Birds fly.\\
Whales swim.\\
\underline{Tweety is a bird.}\\
Tweety flies.
\end{enumerate}
\end{multicols}
\noindent The final four patterns involve different patterns of generic inheritance:

\begin{multicols}{2}
\begin{enumerate}[label=\arabic*., resume]
\setcounter{enumi}{16}
\item \textit{Linear Inheritance (LI)}\vspace{0.1cm}

$Gen(A; \neg C)\\
Gen(B; A)\\
Gen(B; C)\\
Gen(D; B)\\
Gen(D; \neg C)\\
\overline{Gen(A \wedge \neg B; \neg C)}\\
Gen(B \wedge \neg D; C)$

\textit{Example}.\vspace{0.1cm}

Animals do not fly.\\
Birds are animals.\\ 
Birds fly.\\
Ostriches are birds.\\
\underline{Ostriches do not fly. \phantom{birds do not fly}}\\
Animals other than birds do not fly.\\
Birds other than ostriches fly.
\end{enumerate}
\end{multicols}

\begin{multicols}{2}
\begin{enumerate}[label=\arabic*., resume]
\setcounter{enumi}{17}
\item \textit{Tree-Structure Inheritance (TSI)}\vspace{0.1cm}

$Gen(A; \neg C)\\
Gen(B; A)\\
Gen(B; C)\\
Gen(D; A)\\
Gen(D; C)\\
Gen(D; B)\\
Gen(D; \neg C)\\
\overline{Gen(A \wedge \neg B \wedge \neg D ; \neg C)}\\
Gen(B \wedge \neg D; C)$\\
\phantom{i}

\textit{Example}\vspace{0.1cm}

Animals  do not fly.\\
Birds are animals.\\
Birds fly.\\
Bats are animals.\\
Bats fly.\\
Ostriches are birds.\\
\underline{Ostriches do not fly. \phantom{birds and bats do not}}\\
Animals other than birds and bats do not fly.\\
Birds other than ostriches fly.
\end{enumerate}
\end{multicols}

\begin{multicols}{2}
\begin{enumerate}[label=\arabic*., resume]
\setcounter{enumi}{18}
\item \textit{One-Step Multiple Inheritance (OSMI)}\vspace{0.1cm}

$Gen(A;C)\\
Gen(B; \neg C)\\
\overline{Gen(A \wedge \neg B; C)}\\
Gen(B \wedge \neg A; \neg C)$\vfill\null

\item[] \textit{Example}\vspace{0.1cm}

Quakers are pacifists.\\
\underline{Republicans are not pacifists. \phantom{blicans are}}\\
Quakers who are not Republicans are pacifists.\\
Republicans who are not Quakers are not pacifists.
\end{enumerate}
\end{multicols}

\begin{multicols}{2}
\begin{enumerate}[label=\arabic*., resume]
\setcounter{enumi}{19}

\item \textit{Multiple Inheritance (MI)}\vspace{0.1cm}

$Gen(A;C)\\
Gen(B;D)\\
Gen(C;E)\\
Gen(D;E)\\
\underline{Gen(C;\neg D)}\\
Gen(A \wedge \neg B; C)\\
Gen(B \wedge \neg A; \neg C)\\
Gen(A; E), Gen(B; E)\\
Gen(C; E), Gen(D; E)$\vfill\null

\item[] \textit{Example}\vspace{0.1cm}

Quakers are pacifists.\\
Republicans are hawks.\\
Pacifists are politically active.\\
Hawks are politically active.\\
\underline{Pacifists are not hawks.\phantom{akers are pacifists}}\\
Non-Republican Quakers are pacifists.\\
Non-Quaker Republicans are not pacifists.\\
Quakers, Republicans, pacifists and hawks are politically active.\\
\phantom{asdf}
\end{enumerate}
\end{multicols}\vspace{-0.5cm}
\noindent These 20 patterns form a robust suite for testing default reasoning with generics.

\section{Experiments} \label{sec:study1}

\subsection{Models}

We tested 28 LLMs (shown in Figure \ref{fig:1}) encompassing both open- and closed-weight systems (see Appendix A for full list). Closed-weight models (OpenAI, Anthropic, Google) were accessed via their respective APIs. 
Open-weight models ran locally using LM Studio with 6-bit quantized Q6\_K GGUF format (3-bit for Llama 2 70B, Llama 3.1 Instruct 70B, Mixtral 8x7B). Smaller models were executed on a 2021 MacBook Pro (14-inch, 16GB RAM), while larger models were run on a 2023 Mac Studio, 2023 with Apple M2 Ultra chip and 64 GB RAM.
Quantized model versions were sourced from Hugging Face. Code and data is available at: \url{https://github.com/james-kirkpatrick/gen-llm}.

\subsection{Data}

Following the experimental design of \citet{holliday2024conditional}, we created a bank of questions probing defeasible inference capabilities across 21 inference patterns. Each pattern included one handcrafted paradigm instance plus 19 additional instances using Claude 2 and manual curation. For example, for \textit{Defeasible Modus Ponens}, our paradigm instance was: ``From `Dogs have a tail' together with  `David is a dog', can we reasonably infer `David has a tail'?'' All LLM-generated prompts were reviewed by hand and adjustments made where necessary. These standard versions are denoted with the prefix `di' together with the prompt class abbreviation (e.g., diDMP). We also tested a negative version of \textit{One-Step Multiple Inheritance} (diOSMIn) (e.g., ``From `Quakers are pacifists' together with `Republicans are not pacifists', can we reasonably infer `Quakers who are Republicans are pacifists'?''; \emph{Correct answer}: no). Overall, 21 inference patterns were tested in total (20 patterns from \S\ref{sec:benchmark} plus diOSMIn).

To test reliance on world knowledge gained from training data, we created 20 matched instances of each inference pattern using nonsense kind terms (e.g., ``From `Zarpies live in the ocean' together with `Sue is a zarpie', can we reasonably infer `Sue lives in the ocean'?''). These versions are denoted with an `x' suffix (e.g., diDMPx).
To test order effects in two-premise inferences, we reversed premise order.  These versions are denoted with an `o' suffix (e.g., diDMPo). Combined conditions (both nonsense terms and switched premise order) used `ox' suffix (e.g., diDMPox). (See Table \ref{table:prefsuf} in Appendix B for a summary of the prefix--suffix system with examples.)

To test sensitivity to defeasible vs. deductive reasoning contexts and generic vs. universal generalizations, we compared three matched versions of 8 patterns (BDR, DDMP, DI, DMP, DMT, DRDD, DRII, and FI): a standard defeasible version with generics, a deductive version with generics (e.g., ``From `Cats are cute' together with `Chuck is a cat', does it deductively follow `Chuck is cute'?''), and a deductive version with universals (e.g., ``From `All cats are cute' together with `Chuck is a cat', does it deductively follow `Chuck is cute'?''). The latter two versions are denoted by a `v' and `va' prefix (e.g., vDMP and vaDMP) respectively. None of the reasoning patterns are deductively valid when their generic premises are interpreted correctly, but they \textit{become} deductively valid if the generic premise is erroneously interpreted as a universal generalization. Conversely, if the model is misinterpreting the generic premise as a universal one, we would expect it to answer `yes'. Therefore, models sensitive to these distinctions should answer `no' to the v- variants, but `yes' to the va-variants.

\subsection{Evaluation}

For each of the 20 instances of a given inference pattern and for each LLM, we posed each instance under three conditions --- zero-shot, few-shot, or zero-shot chain-of-thought --- at temperature 0 and 1 for zero-shot and few-shot conditions (see Appendix C for condition samples). At temperature 1, we asked the question repeatedly to obtain an empirical distribution for `yes' and `no' answers. We used an early stopping rule (10 consecutive identical answers or maximum 20 repetitions; for expensive models --- Claude 3 Opus, Claude 3.5 Sonnet, Gemini 1.5 Pro, GPT-4o (2024-08-06), GPT-4o (2024-05-13), GPT-4 Turbo (2024-04-09), GPT-4 Turbo (1106), GPT-4 (0613), and GPT-4 (0314) --- the repetition range was reduced to a minimum of 5 and a maximum of 10 times).

To establish a baseline of performance, we evaluated model performance on 15 uncontroversial inference patterns (DMP, DMT, DDMP, ND, PP, DT, DSA, I, BDR, DRII, DRDD, FI, DI, OSMI, and its negation OSMIn, as well as the x, o, and ox variants of these inferences). Responses were generated with greedy decoding (at temperature = 0) and constrained to the single-token outputs `yes' or `no', ensuring chance-level performance corresponds to 50\% accuracy.

To validate the adequacy of using 20 instances of each inference pattern, we tested sensitivity of model responses to the form of reasoning pattern by looking at the Pearson correlation coefficients between the `yes'-response frequencies on the 20 sensical instances of each of the 15 core inference patterns vs the matched nonsense versions (average correlation 0.95). This confirmed sensitivity to logical structure over lexical associations. An ablation study randomly sampled 5, 10, 15, and 20 items per inference pattern (with matched x versions), computed the correlation in model `yes'-response frequencies across these pairs, and repeated this procedure 100 times per sample size. The average correlation increased steadily with more items (5 items: M=0.716, SD=0.076; 10 items: M=0.717, SD=0.057; 15 items: M=0.718, SD=0.051; 20 items: M=0.723, SD=0.044), but never approached the correlation we observed with the full dataset. This suggests 20 items per pattern were necessary for stable correlations.

We also tested robustness under word variations (e.g., `can we defeasibly conclude/deduce', `is it reasonable to conclude', `can we reasonably infer') and premise presentations (conditional; string of sentences) for BDR, using appropriate prefix/suffix labels (Table \ref{table:prefsuf}, Appendix B).

\section{Results} \label{sec:discussion}

Performance was measured against expert judgements of defeasible consequence as reported in the literature.
Aggregate results are shown in Table \ref{table:1} (temperature 0) and Figure \ref{fig:1} (see \S\ref{sec:intro}). Performance varies significantly across models and sizes, revealing important trends about how language models handle generics in default reasoning.

\begin{table}[t!]
\centering
\caption{Model performance on 15 uncontroversial default reasoning inferences under zero-shot, few-shot, and zero-shot chain-of-thought conditions, temperature 0}
\label{table:1}
\resizebox{0.6\textwidth}{!}{%
\begin{tabular}{lccc}
\toprule
\textbf{Model} & \textbf{0-shot} & \textbf{Few-shot} & \textbf{0-shot CoT} \\
(Temp = 0) & \textbf{Accuracy \%} & \textbf{Delta} & \textbf{Delta} \\ 
\midrule
Claude 3 Opus           & 96.51 & -6.81 & -7.64 \\
Claude 3.5 Sonnet       & 96.01 & 0.66 & -0.33 \\
GPT-4 (0613)            & 93.52 & 1.99 & -12.62 \\
GPT-4 Turbo (2024-04-09)& 92.69 & 1.99 & -2.82 \\
Gemma 2 9B Instruct     & 92.69 & 2.33 & -11.79 \\
Claude 3 Sonnet         & 91.36 & 1.33 & -16.28 \\
GPT-4 Turbo (1106)      & 91.19 & 2.82 & -2.16 \\
GPT-4o (2024-08-06)     & 89.03 & -0.66 & 4.98 \\
Gemini 1.5 Flash        & 87.87 & -3.99 & -18.77 \\
Llama 3.1 Instruct 8B   & 86.88 & -10.29 & -15.61 \\
Gemma 2 27B Instruct    & 86.37 & 8.13 & -18.44 \\
Llama 3.1 Instruct 70B  & 85.05 & -0.50 & 3.49 \\
Yi Chat 34B             & 85.88 & 6.31 & -33.22 \\
Gemini 1.5 Pro          & 83.55 & 4.15 & 0.50 \\
Phi-2                   & 80.07 & 3.49 & -4.65 \\
Llama 2 Chat 13B        & 79.90 & 2.82 & -41.03 \\
GPT-4o (2024-05-13)     & 78.24 & 16.27 & 17.44 \\
Llama 2 Chat 70B        & 78.07 & -1.00 & -41.52 \\
Mistral 7B              & 58.47 & 12.62 & -22.09 \\
GPT-3.5 Turbo (1106)    & 58.47 & 10.63 & -7.48 \\
Mixtral 8x7B            & 52.66 & 16.11 & -1.00 \\
Llama 2 Chat 7B         & 51.16 & 28.74 & 1.99 \\
Claude 3 Haiku          & 50.83 & 25.08 & 5.48 \\
Phi 3.1 Mini 4k Instruct& 48.67 & 7.48 & 12.29 \\
GPT-3.5 Turbo (0125)    & 48.00 & 13.46 & -1.50 \\
Code Llama 13B          & 22.92 & 17.44 & 3.98 \\
Code Llama 7B           & 19.93 & 0.83 & 38.03 \\
Code Llama 34B          & 6.48  & 13.79 & 62.79 \\
\bottomrule
\end{tabular}
}
\end{table}

\begin{itemize}
\item \textbf{Overall performance on uncontroversial inferences is impressively high for frontier models}.
Claude 3 Opus and Claude 3.5 Sonnet achieve near-ceiling performance (96+\%) in the zero-shot condition, closely followed by GPT-4 variants and Gemma 2 9B Instruct (90-93\%). Gemma 2 9B Instruct is the only open-weight model exceeding 90\% accuracy. A second tier of 11 models performed 78--89\%, while 7 models performed at near chance and 3 models performed significantly below chance. In addition, we observed:

\begin{itemize}[leftmargin=0.5cm]
\item \textit{Model tier/size relationships are non-linear.}
Despite general trends favouring larger models, notable exceptions exist. Gemma 2 9B outperforms Gemma 2 27B and both Gemini models; Gemini 1.5 Flash outperforms Gemini 1.5 Pro; smaller Llama models outperform their larger counterparts.

\item \textit{High volatility exists within model families.}
Code Llama models perform poorly in zero-shot (6.48\% for the 34B model), but show dramatic improvements under CoT (>62\%). Conversely, Yi Chat 34B and Llama 2 Chat 70B experience steep CoT performance drops, indicating substantial differences in how models encode default reasoning patterns.
\end{itemize}

\item \textbf{Chain-of-thought prompting degrades performance.} Surprisingly, many of the top-performing models show significant performance degradation under CoT prompting (mean accuracy drop -11.14\%, SD 15.74\% in models performing above 75\% accuracy in zero-shot condition, temperature 0). Claude 3 Sonnet drops >16\% and GPT-4 (0613) drops >12\%, with similar or greater declines observed across many other models. This contradicts expectations that CoT enhances reliability and accuracy. 
Importantly, however, CoT prompting does not always lead to performance degradation; GPT-4o models improve under CoT (GPT-4o (2024-05-13) 17.44\%; GPT-4o (2024-08-06) 4.98\%).

\item \textbf{Few-shot prompting yields heterogenous effects.} Weaker models substantially benefit from few-shot examples (GPT-4o (2024-05-13): +16.27\%; Llama 2 Chat 7B, +28.74\%; Claude 3 Haiku, +25.08\%), suggesting that contextual anchoring calibrates inferential behaviour. However, high-performing models show minimal effects, the exception being Claude 3 Opus whose performance dropped in few-shot condition (-6.81\%). These results show that high-performing models do not benefit much from additional context; indeed, additional examples may introduce noise that reduces performance. But other models that performed worse may benefit from additional context that clarifies the notion of default reasoning through examples.

\item \textbf{Prompt wording robustness varies considerably.} Testing alternative formulations of BDR revealed high correlations for minor changes (aBDRdc: 0.99; aBDRdd: 0.99; aBDRsdi: 0.88), but moderate-to-weak correlations for others (aBDRc: 0.54; aBDRnq: 0.43; BDR: 0.47). The aBDRnqc variant showed no correlation (-0.05). These results highlight how subtle shifts in phrasing significantly impact LLM behaviour the importance of careful prompt formulation.
\end{itemize}
In the remainder of this section, we highlight further points of interest from our study.

\subsection{Chain-of-Thought Performance Degradation}

\textbf{Qualitative error analysis under CoT prompting.} To better understand CoT performance degradation, we manually inspected incorrect responses from three representative models with varying CoT deltas (Claude 3.5 Sonnet: -0.33; GPT-4 (0613): -12.62; Gemini 1.5 Flash: -18.77). We identified four common error patterns (see Table \ref{table:cot-errors}, Appendix D for a summary with examples).
\begin{enumerate}
    \item \emph{Generic $\rightarrow$ Universal}: Treating generic claims as universals (e.g., explicitly treating exceptions as counterexamples).
    \item \emph{Defeasible vs. Deductive Confusion}: Reasoning deductively where defeasible reasoning applies, misconstruing concept of defeasibility, or insisting on deductive logic.
    \item \emph{Overly Statistical}: Using frequency/statistical information as decisive factors, even where defeasible endorsement of a generic would be appropriate (e.g., rejecting acceptable generics because they may not be overwhelmingly true).
    \item \emph{Step-Consistent, Bad Conclusion}: Locally coherent intermediate reasoning, but conflicting final conclusions.
\end{enumerate} 
These errors suggest two compatible explanations for CoT performance degradation.

\begin{itemize}
\item \textit{CoT prompting causes ``overthinking''.} Research shows verbal thinking and deliberation can hurt performance on certain tasks in humans \citep[\textit{inter alia}]{schooler1990verbal, dijksterhuis2004think, vandenbos2008intentional}, with similar effects observed in LLMs \citep{liu2025mind}. CoT prompting may encourage overly analytic processing that misrepresents the defeasible character of default reasoning and generic generalizations. In turn, this could lead to decreased performance in our tasks.

\item \textit{CoT prompting reveals conceptual competence gaps.} Our tasks required models to distinguish between defeasible and deductive consequences of premise sets and recognize the exception-permitting behaviour of generics. Error analysis shows models sometimes erroneously applying these concepts, failing to recognize whether it is reasoning defeasibly or deductively, and misinterpreting generic premises as universal generalizations. We examine these behaviours further in \S\ref{sec:genasuni}. 
\end{itemize}

\subsection{Default Reasoning With Irrelevance}

We tested five inference patterns examining how LLMs handled default reasoning with irrelevant information (e.g, information about exceptional individuals or extraneous independent information). (See Appendix E for model performance on these inferences in zero-shot condition.)

\begin{itemize}[leftmargin=0cm]
\item[] \textbf{Basic Default Reasoning.} Exceptional individuals should not affect inferences about independent alternatives (e.g., from \textit{Heavy blocks are located on the table}, \textit{Blocks A and B are heavy}, \textit{A is not on the table}, one should defeasibly infer \textit{B is on the table}). Most newer/larger models performed well on this pattern in zero-shot (temperature 0). Notable exceptions were: GPT-4o (2024-05-13) performed at chance but improved in few-shot condition; Gemini 1.5 Pro performed significantly below chance, while Gemini 1.5 Flash achieved at 75\%. Some models (Claude 3 Opus, Gemini 1.5 Flash) declined in few-shot, while others (Llama 2 7B Chat) improved.

\item[] \textbf{Default Reasoning with Irrelevant Information.} Irrelevant information (\textit{B is red}) should not interfere with basic default inferences. Top models performed well in zero-shot (temperature 0), most improving slightly in few-shot except Claude 3 Opus and Gemini 1.5 Flash. Llama 2 Chat 7B showed significant few-shot improvement at temperature 0, but regressed completely at temperature 1.

\item[] \textbf{Default Reasoning with a Disabled Default.} Mentioning possibly exceptional independent individuals (\textit{A is possibly an exception to this rule}) should not affect inferences about other individuals. Most top models performed well in zero-shot (temperature 0).
GPT-4o (2024-05-13) again performed only at chance, improving in zero-shot (temperature 1) and few-shot (temperature 0). Several top models performed worse in few-shot, complicating any simple interpretations of performance.

\item[] \textbf{Fact Irrelevance.} Irrelevant facts about individuals should not block default inferences (e.g., \textit{Tweety is a sparrow} should not block the inference from \textit{Birds fly} and \textit{Tweety is a bird} to \textit{Tweety flies}). Top models performed well in zero-shot (temperature 0); smaller models (Claude 3 Haiku, Mistral 7B) achieved $\sim$75\%. GPT 3.5 (1106), Claude 3 Haiku, and Llama Chat 7B improved substantially in few-shot condition (75--100\%).

\textbf{Default Irrelevance.} Irrelevant defaults (`Whales swim') should not interfere with separate default inferences (`Tweety flies'). Top models succeeded in zero-shot (temperature 0). Others performed at/below chance with modest gains at temperature 1 and significant improvements in few-shot (except Llama Code 7B).
\end{itemize}
\noindent In summary, across patterns involving irrelevant facts or defaults, newer/larger models generally succeeded at disregarding exceptional yet irrelevant information in zero-shot (temperature 0)  conditions, though with  notable exceptions. Performance patterns varied: some models worsened in few-shot or at temperature 1, while others achieved competence only with additional prompting. These results suggest that top-tier LLMs often handle irrelevance effectively, while smaller/older models require explicit few-shot guidance.

\subsection{Logical Complexity and Length of Arguments}

LLMs have been found to struggle with negation contextually complex single inference rules \citep{parmeretal2024logicbench}. Our results partly align with these findings, but suggest a more nuanced picture.

\begin{itemize}[leftmargin=0cm]
\item[] \textbf{Negation.} Among models with overall performance <90\% (as shown in Figure \ref{fig:1}), performance on negation-involving inferences (ND, DMT, DDMP, BDR, DRII, DRDD, OSMI) was highly variable in zero-shot conditions. Code Llama models performed well on ND and DMT, but fell below chance performance on DDMP, BRD, DRII, DRDD, and OSMI. Gemini 1.5 Flash model (87.87\% overall) performed well on DMT, DDMP, DRDD, and OSMI, but poorly on BRD and DRII. These inconsistencies suggest model-specific idiosyncrasies rather than a general failure tied to negation. In contrast, models with >90\% performance generally handled negation robustly in both zero- and few-shot conditions (temperature 0), with exceptions: GPT-4o models dropped to at/below chance on ND, BDR, and DRDD in zero-shot; Claude 3 Opus showed decreased performance on BDR despite otherwise being a top performer.  

\item[] \textbf{Argument Length.} Inference patterns with more than 3 premises (ND, BDR, DRII, DRDD) did not elicit systematically lower performance among models with >90\% accuracy. These models remained stable across zero-shot and few-shot conditions. Lower-performing models (<90\%) showed variable performance on longer patterns, but no consistent degradation was observed from argument length alone.
\end{itemize}
In summary, neither negation nor argument length systematically undermined performance across all models. While lower-performing models showed fluctuating performance across these dimensions, top-tier models maintained stable performance.

\subsection{Premise Order Effects}

For two-premise inferences (DMP, DMT, DT), we investigated whether premise order affected performance.

\begin{itemize}[leftmargin=0cm]
\item[] \textbf{Defeasible Modus Ponens.} Zero-shot performance typically improved when specific premises were presented first. This trend reversed in few-shot, where models often performed better when generic premises were presented first.

\textbf{Defeasible Modus Tollens.} Results varied considerably. In zero-shot (temperature 0), GPT-4o models, Code Llama 13B, and Mistral 7B performed better with specific premise first (the latter two models improving from below chance). However, GPT-4 Turbo models showed opposite trends: performance dropped with specific-first, with GPT-4 Turbo (1106) falling below chance. Similar patterns occurred in few-shot conditions.

\textbf{Defeasible Transitivity.} Zero-shot performance typically improved when the specific premise came first.
Code Llama 13B's performance improved from 0\% to $\sim80\%$. GPT-4 Turbo again declined under this ordering. While standard prompt variants improved from zero-shot to few-shot, no such improvement was observed for any model family in the reordered (o) variant, which actually declined  slightly in few-shot compared to the standard variants.
\end{itemize}
\noindent These order effects are unexpected: order should not affect the validity or plausibility of an inference, yet surface presentation clearly guides LLM inference. This suggests that internal representations are sensitive to input structure in ways that diverge from logical norms.

\subsection{Reasoning with Contradictory Information}

Previous research reports that LLMs struggle to reason with inconsistent or contradictory information \citep{kazemi2024boardgameqa}.
Two of our inference patterns involved potentially inconsistent or contradictory information: ND and DDMP.
Performance with ND varied considerably among high-performing models in zero-shot (temperature 0). Claude family models performed well (>80\%), while the GPT-4 family models varied from below chance (GPT-4 Turbo (1106) to 100\% (GPT-4o (2024-05-06). In contrast, performance with DDMP was uniformly high. This suggests that high-end models do have some capacity for correct default reasoning with generic and contradictory information.

\subsection{Do LLMs Interpret Generics as Universals?} \label{sec:genasuni}

To examine whether LLMs properly interpret generics as exception-permitting generalizations and distinguish generics from universal quantification in deductive contexts, we compared aggregate model performance on 8 basic default reasoning patterns (BDR, DDMP, DI, DMP, DMT, DRDD, DRII, FI) across three variants:

\begin{itemize}
    \item \textbf{Base}: ``From `Adults have driving licenses' \ldots can we defeasibly infer\ldots''. (Expected answer: `Yes')
    \item \textbf{v variant}: ``From `Adults have driving licenses'\ldots does it deductively follow that\ldots''. (Expected answer: `No')
    \item \textbf{va variant}: ``From `All adults have driving licenses'\ldots does it deductively follow that\ldots''. (Expected answer: `Yes')
\end{itemize}
\begin{itemize}[leftmargin=0cm]
\item[] \textbf{Defeasibly vs. Deductive Distinction.} To examine whether LLMs interpret generics as exception-permitting generalizations, we analyzed Pearson correlation coefficients between aggregate correct answer percentages over the 28 models on the 20 sensical instances of each base patterns vs the matched v variants. Since the defeasible inferences are not deductively valid, we expected models to answer `yes' for the base (defeasible inference) variants and `no' for the v (deductive entailment) variants.

Analysis revealed unexpectedly strong negative correlations between to the base and v variants 6 out of 8 patterns in aggregate model responses (r < -0.6, all correlations statistically significant). Only DDMP showed expected positive correlation (r=0.90). Table \ref{table:prompt_level_correlations} and Figure \ref{fig:comparisons} present these correlations. These negative correlations indicate that systematic failure to distinguish defeasible from deductive contexts with generics, suggesting models either (i) fail to recognize logical distinctions between these reasoning contexts, or (ii) treating generics as universals, making both contexts deductively valid.

\begin{table}[t]
\centering
\caption{\label{table:prompt_level_correlations} Pearson correlations among prompt level performance over 28 models for 8 baseline inference patterns involving only one generic premise}
\resizebox{0.6\textwidth}{!}{%
\begin{tabular}{lrrrrrrrr}
\toprule
 & \multicolumn{2}{c}{\textbf{base vs v}} & \multicolumn{2}{c}{\textbf{v vs va}} & \multicolumn{3}{c}{\textbf{mean}} \\
\multicolumn{1}{l}{\textbf{Prompt}} & \multicolumn{1}{c}{\textbf{corr}} & \multicolumn{1}{c}{\textbf{p}} & \multicolumn{1}{c}{\textbf{corr}} & \multicolumn{1}{c}{\textbf{p}} & \multicolumn{1}{c}{\textbf{base}} & \multicolumn{1}{c}{\textbf{v}} & \multicolumn{1}{c}{\textbf{va}} \\
\midrule
BDR   & -0.49 & <.01 & -0.50 & <.01 & 59.20 & 61.16 & 29.92 \\
DDMP  & 0.90  & <.01 & -0.75 & <.01 & 86.96 & 90.20 & 6.98 \\
DI    & -0.84 & <.01 & -0.67 & <.01 & 72.99 & 30.06 & 65.42 \\
DMP   & -0.68 & <.01 & -0.55 & <.01 & 80.50 & 25.06 & 74.77 \\
DMT   & -0.61 & <.01 & -0.25 & .19 & 80.30 & 34.48 & 69.30 \\
DRDD  & -0.69 & <.01 & -0.50 & <.01 & 62.13 & 56.86 & 38.29 \\
DRII  & -0.80 & <.01 & -0.66 & <.01 & 60.63 & 54.96 & 31.54 \\
FI    & -0.80 & <.01 & -0.73 & <.01 & 75.77 & 31.23 & 67.67 \\
\bottomrule
\end{tabular}
}
\end{table}

\begin{figure}[t]
    \centering
  \includegraphics[width=1\textwidth,]{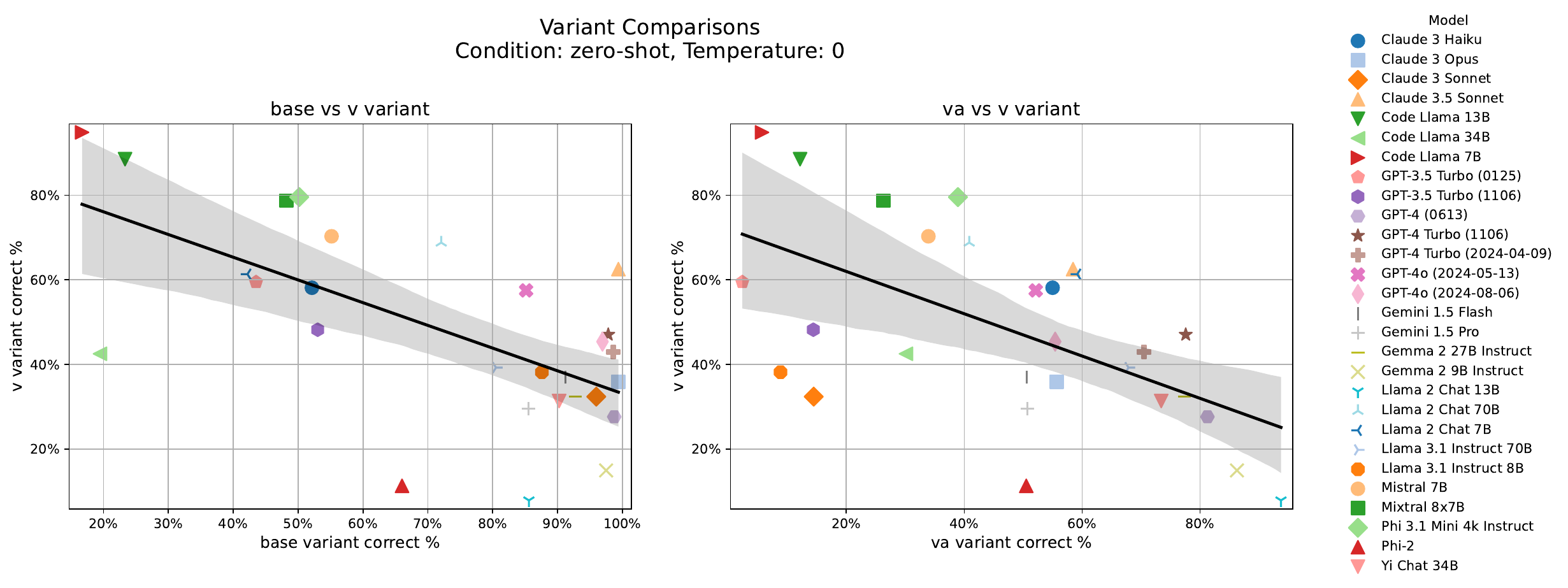}
    \caption{Correlations of zero-shot performance on base vs. v, v vs va. The correlations are -0.647 and -0.584, respectively, The p-values are both <.001.}
    \label{fig:comparisons}
\end{figure}

\item[] \textbf{Generic vs. Universal Distinction.} To examine whether LLMs distinguish generics from universal generalizations in deductive inference contexts, we analy`ed correlations between aggregate correct answer percentages for v versus va variants. Since the inferences with generic premises are not deductively valid but the inferences with universal premises are, we expected models to answer `no' for v (generic--deductive entailment) variants and `yes' for va (universal--deductive entailment) variants.

Analysis revealed strong negative correlations between v and va variants for nearly all patterns (Table \ref{table:prompt_level_correlations}), indicating systematic failure to distinguish generics from universals in deductive contexts. 

\item[] \textbf{Performance Asymmetries.} Models demonstrate superior performance on base defeasible inferences (59--87\%) compared to rejecting deductive entailment from generics (25--61\%).
Performance on va variants exhibits highly variability (7--75\%), suggesting inconsistent treatment of universal quantification.
Some patterns (DI, DMP, FI) show high performance on both base and va variants but poor v performance, indicating that these models may be defaulting to `yes' responses regardless of logical validity.
\end{itemize}
These findings raise concerns about the capacity of current LLMs for non-monotonic reasoning.  The systematic failure to distinguish defeasible from deductive contexts when processing generics suggests fundamental limitations in understanding the logical properties of natural language quantification and the context-sensitive nature of natural language inference.

\subsection{Relation to Other Benchmarks}

\begin{figure}[t]
    \centering
    \includegraphics[width=\textwidth]{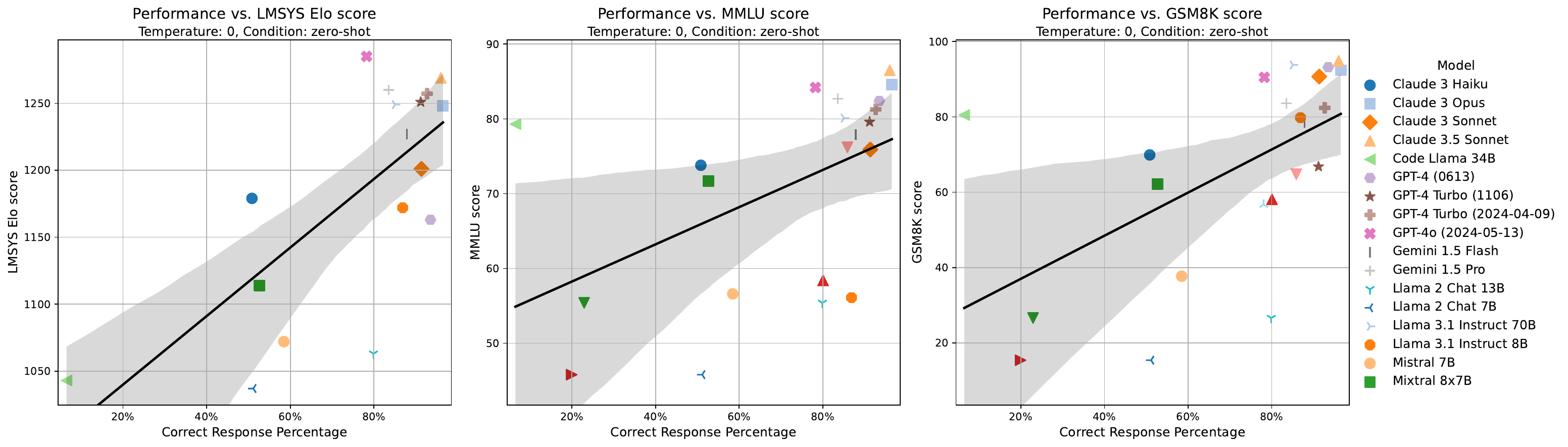}
    \caption{Correlations of zero-shot performance on default reasoning benchmark vs. LMSYS Elo ratings, MMLU scores, and GSM8k scores. The correlations are 0.72, 0.50, and 0.57, respectively. The p-values are 0.0012, 0.02, and 0.005, respectively.}
    \label{fig:ranking}
\end{figure}

To orient our default reasoning tasks in relation to other LLM evaluations, we compare model performance on our benchmark of uncontroversial instances with three popular benchmarks for general LLM performance: ChatBot Arena \citep{chiang2024chatbot}, MMLU \citep{hendrycks2021measuring}, and GSM8K \citep{cobbe2021training}. Figure \ref{fig:ranking} shows the results. Our results are strongly correlated with LMSYS Elo, suggesting that default reasoning capabilities correlate with general model performance tasks, although there is only moderate correlation with MMLU and GSM8K, suggesting that default reasoning performance is less correlated with domain knowledge and mathematical reasoning.

\section{Conclusions and Future Work} \label{sec:conc}

The ability to defeasibly draw conclusions from generic generalizations is crucial for real-world applications of language models. We evaluated 28 large language models on 20 benchmark default reasoning problems across two temperature settings (0 and 1) and three prompting conditions (zero-shot, few-shot, zero-shot CoT). Performance varied widely across and within model families, though several models (Claude 3 Opus, Claude 3.5 Sonnet, GPT-4, GPT-4 Turbo, Gemma 2 9B Instruct) exceeded 90\% accuracy in zero-shot condition. Few-shot prompting sometimes improved performance, but chain-of-thought prompting often led to serious performance degradation. Most models struggled to maintain a distinction between defeasible and deductive reasoning contexts, and between generics and universals, raising concerns about the capacity for non-monotonic reasoning.

Our findings also bear on the broader contrast between connectionist and symbolic systems, specifically whether connectionist systems replicate or diverge from inference patterns traditionally modelled by the symbolic systems. Whereas symbolic systems implement rules as explicit instructions, connectionist systems embed them as patterns of activation and weighted connections within a neural network. Rules are not stored or hard coded as discrete statements, but emerge as learned dispositions to generate certain outputs given certain inputs. While frontier models sometimes display dispositions consistent with expert default reasoning, these behaviours are fragile under prompting conditions and styles. This highlights the limitations of connectionist systems as substitutes for symbolic reasoning.

There are several natural paths for future work. First, LLM behaviour should be compared with actual human behaviour on the inferences we tested. We used reported expert human judgements as a benchmark for performance, although there is some evidence that experimental human subjects draw different inferences \citep{elio1993human, pelletier1997what, pelletier2005case}. Second, there are many more default inferences involving generics and other logical operators to explore. Third, reasoning with generics is connected with probabilistic and inductive inference, and so this could provide another perspective on LLM reasoning capabilities. Fourth, future evaluations of default reasoning should test new models (OpenAI's o1, o3, and GPT-5; Anthropic's latest Claude models; Google's Gemini 2.5 Pro; Deepseek's V3). Testing these models for default reasoning capabilities would shed light on the kinds of reasoning tasks in which they excel. Finally, future work should look at the performance of hybrid models that blend connectionist and symbolic approaches.

\section{Limitations}

Local open-weight models ran in quantized form, which may have reduced their performance.

Our 15 baseline inference patterns excluded social and artefact kinds, due to LLM bias and guardrails against stereotyping, and naming artefacts is unnatural in English. However, results showed no bias between sensical vs. non-sensical kind terms, which suggests this restriction was not limiting. We also focused on default reasoning with bare plural generics. Future work should test social and artefact generics, as well as indefinite or definite singular generics.

Inference patterns were based on non-monotonic reasoning benchmark problems, raising possible data contamination. Although we minimized risks (e.g., using pure inference pattern prompts), we cannot rule out overlap with training data.

All experiments were conducted in English. 
Future research should evaluate default reasoning and generics in other languages.

\section*{Acknowledgments}
For helpful feedback, we should like to thank Wesley Holliday, Philipp Koralus, Matthew Mandelkern, Kyriacos Shiarlis, Alejandra Vergara-Lope, and especially Ryan Kearns, as well as the audience at a research seminar in the Institute for Ethics in AI, University of Oxford.

\renewcommand{\bibsection}{\section*{References}}
\setlength{\bibsep}{0.0pt}
\bibliographystyle{plainnat}
\bibliography{main}

\pagebreak
\appendix

\section{List of models}

In this section, we cite all LLMs used to conduct our experiments: the OpenAI GPT-4o models \citep{hurst2024gpt} GPT-4o (2024-08-06) and GPT-4o (2024-05-13); the OpenAI GPT-4 model family \citep{openai2023gpt4}, specifically, GPT-4 Turbo (2024-04-09), GPT-4 Turbo (1106), and GPT-4 (0613);
the OpenAI GPT 3.5 model family \citep{brown2020language}, specifically GPT-3.5 Turbo (0125), and GPT-3.5 Turbo (1106);
the Anthropic Claude 3 family models\footnote{\url{https://www.anthropic.com/news/claude-3-family}}, Claude 3 Opus, Claude 3 Sonnet, and Claude 3 Haiku, and Claude 3.5 Sonnet\footnote{\url{https://www.anthropic.com/news/claude-3-5-sonnet}}; 
the Gemini 1.5 model family \citep{reid2024gemini}, specifically, Gemini 1.5 Pro and Gemini 1.5 Flash; 
Code Llama 7B, Code Llama 13B, Code Llama 34B \citep{roziere2024code};
Llama 2 Chat 7B, Llama 2 Chat 13B, Llama 2 Chat 70B \citep{touvron2023llama}; 
Llama 3.1 Instruct 8B, Llama 3.1 Instruct 70B, 
Mistral 7B \citep{jiang2023mistral}; Mixtral 8x7B \citep{jiang2024mixtral}; 
Gemma 2 9B Instruct and Gemma 2 27B Instruct \citep{team2024gemma};
Yi Chat 34B \citep{ai2024yi}; 
Phi-2 7B \citep{javaheripi2023phi}; 
and Phi 3.1 Mini 4k Instruct \citep{abdin2024phi3}.

\section{Summary of Prefix--Suffix System} \label{app:prefix}

\begin{table}[h]
\centering
\caption{\label{table:prefsuf} Guide to Prefix--Suffix System for Prompt Class Names}
\resizebox{\textwidth}{!}{
\begin{tabular}{lllp{5in}p{2.5in}}
\toprule
\multicolumn{5}{c}{\textbf{Core Default Reasoning Inferences}}\\
\midrule
Prefix & Prompt & Suffix & Example & Notes\\
\midrule
di     & DMP         &        & ``From `Dogs have a tail' together with `David is a dog', can we defeasibly infer `David has a tail'?''                                                            & ``defeasibly infer'', DMP, sensical, generic                                \\
di     & DMP         & o      & ``From `David is a dog' together with `Dogs have a tail', can we defeasibly infer `David has a tail'?''                                                            & ``defeasibly infer'', DMP, non-sensical, reversed premise order, generic    \\
di     & DMP         & x      & ``From `Zarpies live in the ocean' together with `Sue is a zarpie', can we defeasibly infer `Sue lives in the ocean'?''                                            & ``defeasibly infer'', DMP, non-sensical, generic                            \\
di     & DMP         & ox     & ``From `Sue is a zarpie' together with `Zarpies live in the ocean', can we defeasibly infer `Sue lives in the ocean'?''                                            & ``defeasibly infer'', DMP, reversed premise order, non-sensical, generic    \\
\midrule
\multicolumn{5}{c}{\textbf{OSMI variant (o, x, ox variants omitted)}}\\
\midrule
di & OSMI & & ``From `Quakers are pacifists' together with `Republicans are not pacifists', can we defeasibly infer `Quakers who are not Republicans are pacifists'?'' & ``defeasibly infer'', OSMI, sensical, generic \\
di & OSMI & n & ``From `Quakers are pacifists' together with `Republicans are not pacifists', can we defeasibly infer `Quakers who are Republicans are pacifists'?'' & ``defeasibly infer'', OSMI, sensical, generic, n-variant \\
\midrule
\multicolumn{5}{c}{\textbf{Core Deductive Reasoning Inferences (Generic)}}\\
\midrule
v      & DMP         &        & ``From `Dogs have a tail' together with `David is a dog', does it deductively follow that `David has a tail'?''                                                    & ``deductively follow'', DMP, sensical, generic                              \\
v      & DMP         & x      & ``From `Zarpies live in the ocean' together with `Sue is a zarpie', does it deductively follow that `Sue lives in the ocean'?''                                    & ``deductively follow'', DMP, non-sensical, generic                          \\
\midrule
\multicolumn{5}{c}{\textbf{Core Deductive Reasoning Inferences (Universal)}}\\
\midrule
va     & DMP         &        & ``From `All dogs have a tail' together with `David is a dog', does it deductively follow that `David has a tail'?''                                                & ``deductively follow'', DMP, sensical, universal                            \\
va     & DMP         & x      & ``From `All zarpies live in the ocean' together with `Sue is a zarpie', does it deductively follow that `Sue lives in the ocean'?''                                & ``deductively follow'', DMP, non-sensical, universal                        \\
\midrule
\multicolumn{5}{c}{\textbf{Alternative Prompt Phrasings}}\\
\midrule
       & BDR         &        & ``From `Dogs have a tail' together with `David is a dog' and 'Chuck is a dog' and 'Chuck does not have a tail', can we reasonably infer `David has a tail'?''      & ``reasonably infer'', BDR, sensical, generic                                \\
a      & BDR         & c      & ``From `Dogs have a tail' together with `David is a dog' and 'Chuck is a dog' and 'Chuck does not have a tail', is it reasonable to conclude `David has a tail'?'' & ``reasonable to conclude'', BDR, sensical, generic                          \\
a      & BDR         & dc     & ``From `Dogs have a tail' together with `David is a dog' and 'Chuck is a dog' and 'Chuck does not have a tail', can we defeasibly conclude `David has a tail'?''   & ``defeasibly conclude'', BDR, sensical, generic                             \\
a      & BDR         & dd     & ``From `Dogs have a tail' together with `David is a dog' and 'Chuck is a dog' and 'Chuck does not have a tail', can we defeasibly deduce `David has a tail'?''     & ``defeasibly deduce'', BDR, sensical, generic                               \\
a      & BDR         & nq     & ``If dogs have a tail and David is a dog and Chuck is a dog and Chuck does not have a tail, can you defeasibly infer that David has a tail?''                      & ``defeasibly infer'' (conditional version), BDR, sensical, generic          \\
a      & BDR         & nqc    & ``If dogs have a tail and David is a dog and Chuck is a dog and Chuck does not have a tail, is it reasonable to conclude that David has a tail?''                  & ``reasonable to conclude'' (conditional version), BDR, sensical, generic    \\
a      & BDR         & sdi    & "Dogs have a tail. Also David is a dog. Also Chuck is a dog. Also Chuck does not have a tail. Can we defeasibly infer that David has a tail?''                    & ``defeasibly infer'' (sequence of premises version), BDR, sensical, generic\\
\bottomrule
\end{tabular}
}
\end{table}

\newpage
\section{Prompts used in experiments}

Here are specific examples of the prompts for an inference pattern in zero-shot, few-shot, and zero-shot chain-of-thought conditions in the experiments.

\subsection{Zero-shot prompt} 

\begin{lstlisting}[style=tt]
Answer only with `yes' or `no' and nothing else.

From `Dogs have a tail' together with `David is a dog' and `Chuck is a dog' and `Chuck does not have a tail', can we reasonably infer `David has a tail'?
\end{lstlisting}

\subsection{Few-shot prompt} 

\begin{lstlisting}[style=tt]
Consider the following examples:

From `Birds fly' together with `Penguins don't fly' and `Penguins are birds' and `Tweety is a penguin', can we defeasibly infer that `Tweety doesn't fly'? Correct answer: Yes.

From `Animals do not fly' together with `Birds are animals' and `Birds fly' and `Ostriches are birds' and `Ostriches do not fly', can we defeasibly infer that `Animals other than birds do not fly'? Correct answer: Yes.

From `Birds fly' together with `Tweety is a bird', can we defeasibly infer that `Tweety can fly'? Correct answer: Yes.

From `Nixon is a Quaker' together with `Nixon is a Republican' and `Quakers are doves' and `Republicans are hawks' and `Doves are politically active' and `Hawks are politically active', can we defeasibly infer that `Nixon is politically active'? Correct answer: Yes.

Now here is a question for you,  which you should answer with only `yes' or `no' and nothing else:

From `Dogs have a tail' together with `David is a dog' and `Chuck is a dog' and `Chuck does not have a tail', can we reasonably infer `David has a tail'?
\end{lstlisting}

\subsection{Chain-of-thought prompting}

\begin{lstlisting}[style=tt]
In response to the following question, think step by step and explain your reasoning, starting your response with `Explanation: '; then *after* explaining your reasoning, when you are ready to answer, simply write `Answer: ' followed by `yes' or `no' and nothing else. Please make sure to format your response by first explaining your reasoning and then writing `Answer:' followed by `yes' or `no' at the very end.

From `Dogs have a tail' together with `David is a dog' and `Chuck is a dog' and `Chuck does not have a tail', can we reasonably infer `David has a tail'?
\end{lstlisting}

\newpage

\section{Common Errors in Chain-of-Thought Prompting} \label{app:error}

\begin{table}[h]
\centering
\caption{\label{table:cot-errors}Descriptions and Examples of Common Errors in Chain-of-Thought Defeasible Reasoning with Generics}
\resizebox{\textwidth}{!}{
\begin{tabular}{p{0.9in}p{1.2in}p{5in}}
\toprule
\textbf{Error Pattern} & \textbf{Description} & \textbf{Example} \\
\midrule
\multirow[t]{4}{0.9in}{\textit{Generic $\rightarrow$ \phantom{oo} Universal}} & \multirow[t]{4}{1.2in}{Generic treated as universal; exceptions explicitly treated as counterexamples} & \emph{Model}: GPT-4 (0613) \\ 
& & \emph{Prompt Class}: diOSMInx \\
& & \emph{Prompt}: From `Snerfs are happy' together with `Quandles are not happy', can we defeasibly infer `Snerfs who are quandles are happy'?\\
& & \emph{Response}:
  Explanation: The statement `Snerfs are happy' implies that all Snerfs, regardless of any other characteristics they may have, are happy. The statement `Quandles are not happy' implies that all Quandles are not happy. However, the question asks about `Snerfs who are Quandles'. This is a subset of Snerfs that also have the characteristic of being Quandles. Given the information provided, we have a contradiction. On one hand, if they are Snerfs, they should be happy. On the other hand, if they are Quandles, they should not be happy. Therefore, we cannot make a definitive inference about the happiness of Snerfs who are also Quandles based on the information provided. Answer: no
\\
\multirow[t]{4}{0.9in}{\textit{Defeasible vs. Deductive Confusion}} & \multirow[t]{4}{1.2in}{Deductive reasoning used, not defeasible reasoning; concept of defeasibility misconstrued or misapplied} & \emph{Model}: Gemini 1.5 Flash \\
&& \emph{Prompt Class}: diDRII \\
&& \emph{Prompt}: From `Hummingbirds drink nectar' together with `Henry is a hummingbird' and `Frank is a hummingbird' and `Frank doesn't drink nectar' and `Henry is quick', can we defeasibly infer `Henry drinks nectar'?\\
&&\emph{Response}: Explanation: We are given that hummingbirds drink nectar.  We know Henry is a hummingbird.  However, we are also told that Frank is a hummingbird but Frank *doesn't* drink nectar. This contradicts the general statement that hummingbirds drink nectar.  The existence of Frank, a hummingbird who doesn't drink nectar, weakens the defeasible inference that all hummingbirds drink nectar.  Therefore, while it's *likely* Henry drinks nectar based on the general rule, the exception presented by Frank makes the inference defeasible -- meaning it could be false.  The fact that Henry is quick is irrelevant to whether he drinks nectar. Answer: no\\
\multirow[t]{4}{0.9in}{\textit{Over-Statistical}} & \multirow[t]{4}{1.2in}{Frequency or statistical information taken as decisive, even where defeasible endorsement of a generic would be appropriate} & \emph{Model}: Gemini 1.5 Flash \\
&& \emph{Prompt Class}: diDSA \\
&& \emph{Prompt}: From `Spiders spin webs', can we defeasibly infer `Brave spiders spin webs'?\\\
&& \emph{Response}: Explanation: The statement ``Spiders spin webs'' describes a characteristic behavior of spiders.  The statement ``Brave spiders spin webs'' adds an attribute -- bravery -- to the spiders that spin webs.  While it's possible that spiders exhibit bravery in spinning webs (perhaps facing dangers while doing so), the original statement doesn't provide any information about their bravery.  The inference is not logically guaranteed.  We can't defeasibly infer bravery from the act of web-spinning.  The connection is not established.  Therefore, the inference is invalid.\\
\multirow[t]{4}{0.9in}{\textit{Step-Consistent, Bad Conclusion}} & \multirow[t]{4}{1.2in}{Intermediate reasoning is locally coherent but conclusion contradicts those steps} & \emph{Model}: Gemini 1.5 Flash\\ 
& & \emph{Prompt Class}: diDSA \\
& & \emph{Prompt}: From `Hummingbirds drink nectar', can we defeasibly infer `Happy hummingbirds drink nectar'?\\
& & \emph{Response}: Explanation: The statement ``Hummingbirds drink nectar'' describes a general characteristic of hummingbirds.  It doesn't specify anything about their emotional state.  The statement ``Happy hummingbirds drink nectar'' adds a condition (happiness) that is not present in the original statement. While it's *possible* that happy hummingbirds drink nectar, it's not necessarily true.  A hummingbird could be unhappy and still drink nectar.  Therefore, the inference is defeasible; it could be proven false by finding a counterexample (a sad hummingbird drinking nectar).\\
\bottomrule
\end{tabular}
}
\end{table}

\newpage
\section{Additional results}

\subsection{Individual inferences}

In this Appendix, we show the zero-shot performance of LLMs on the 21 inference patterns we tested, at temperatures 0 and 1.

\raggedbottom

\subsubsection{Defeasible Modus Ponens (DMP)}

\begin{center}
\includegraphics[scale=.25]{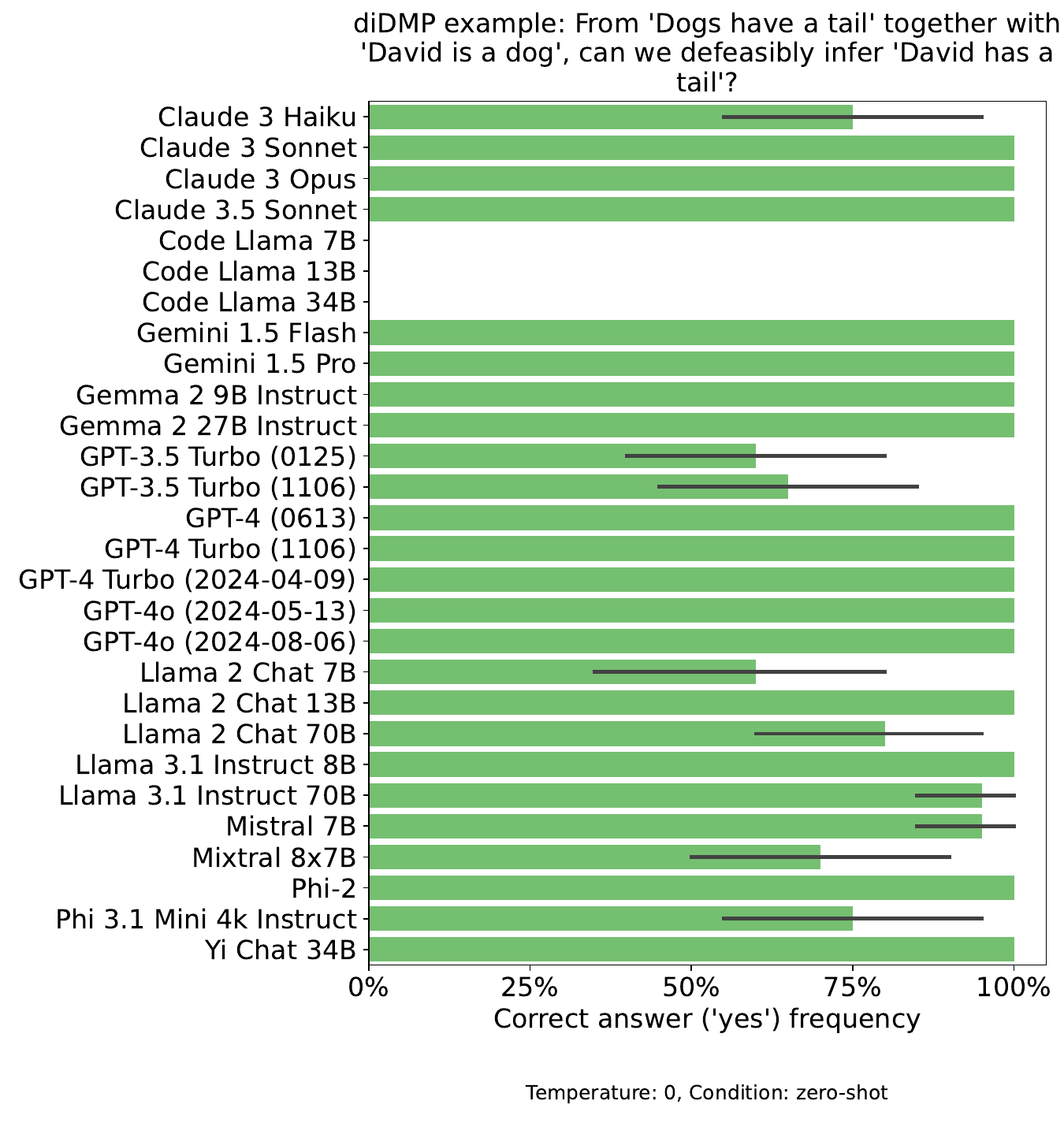}
\includegraphics[scale=.25]{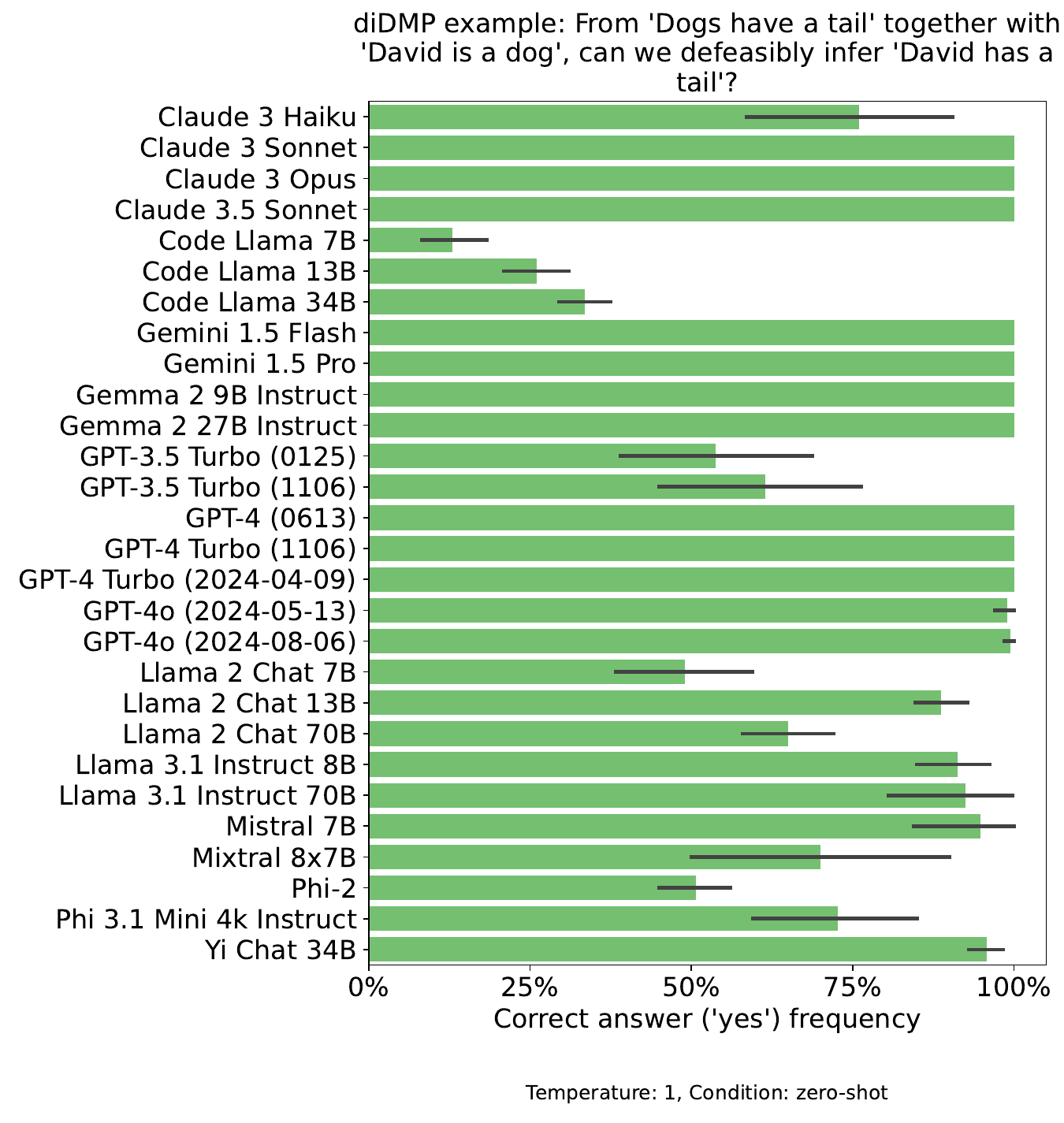}
\end{center}

\subsubsection{Defeasible Modus Tollens (DMT)}

\begin{center}
\includegraphics[scale=.25]{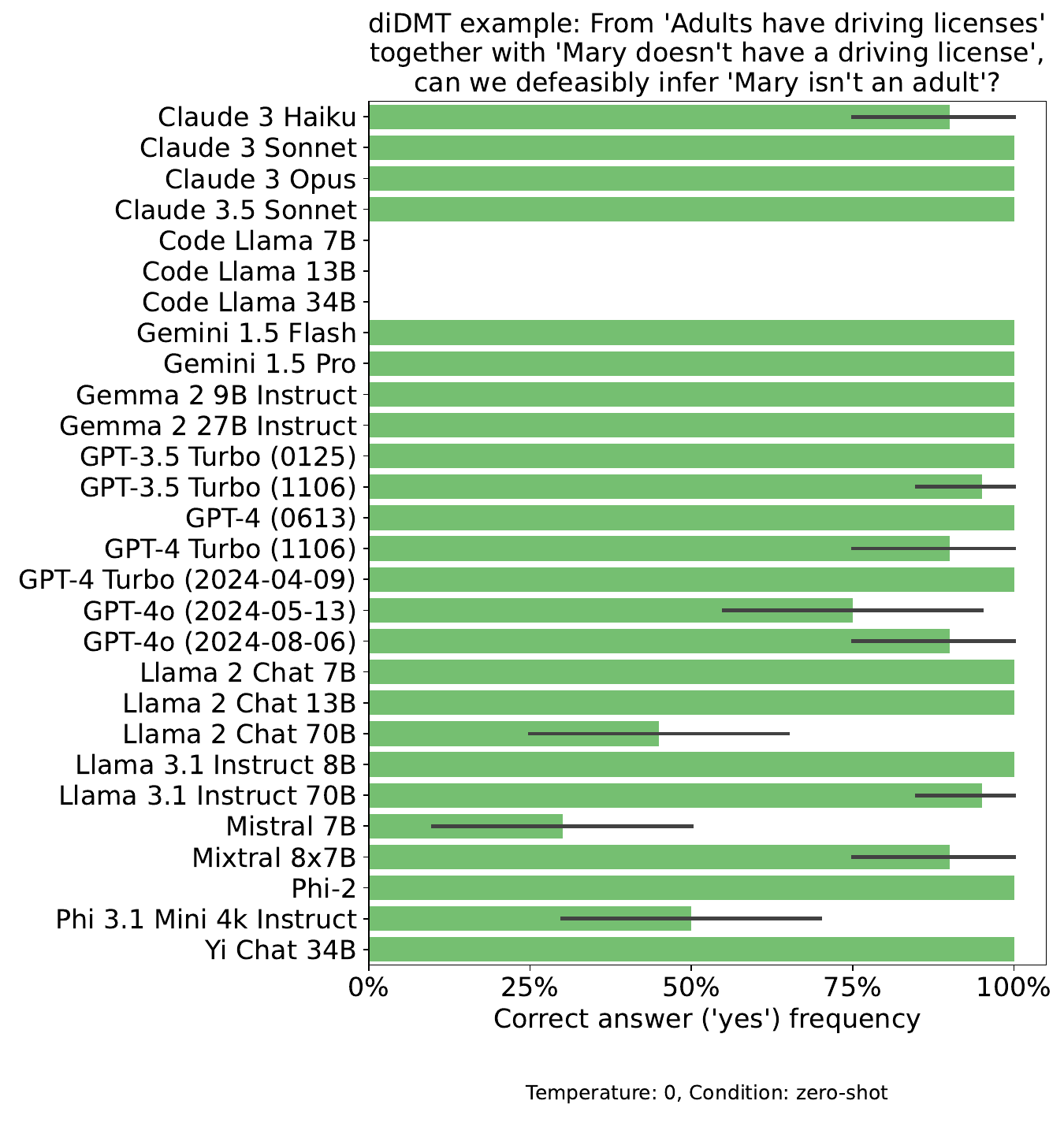}
\includegraphics[scale=.25]{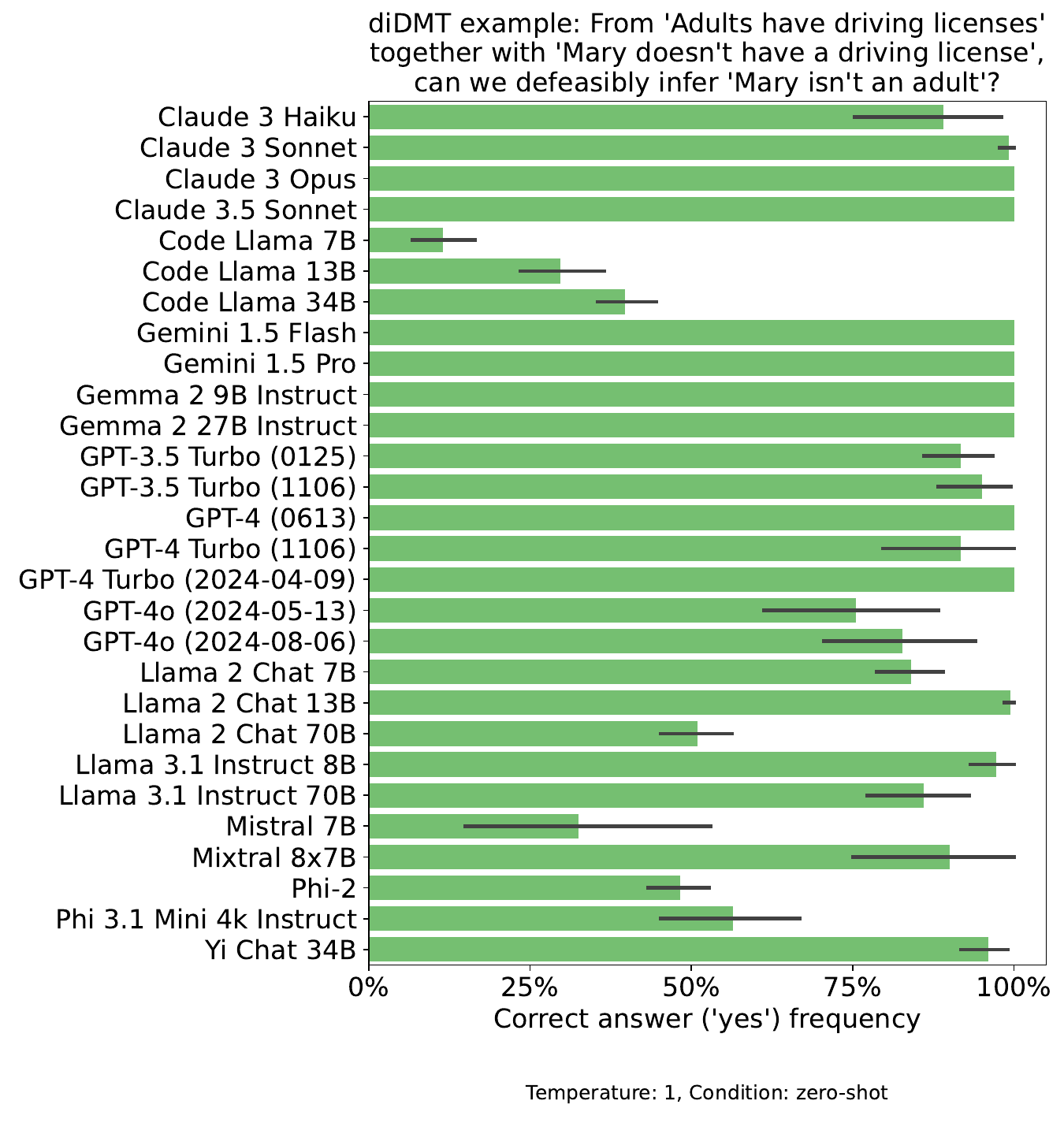}
\end{center}

\pagebreak
\subsubsection{`Defeat' of DMP/DMT (DDMP)}

\begin{center}
\includegraphics[scale=.25]{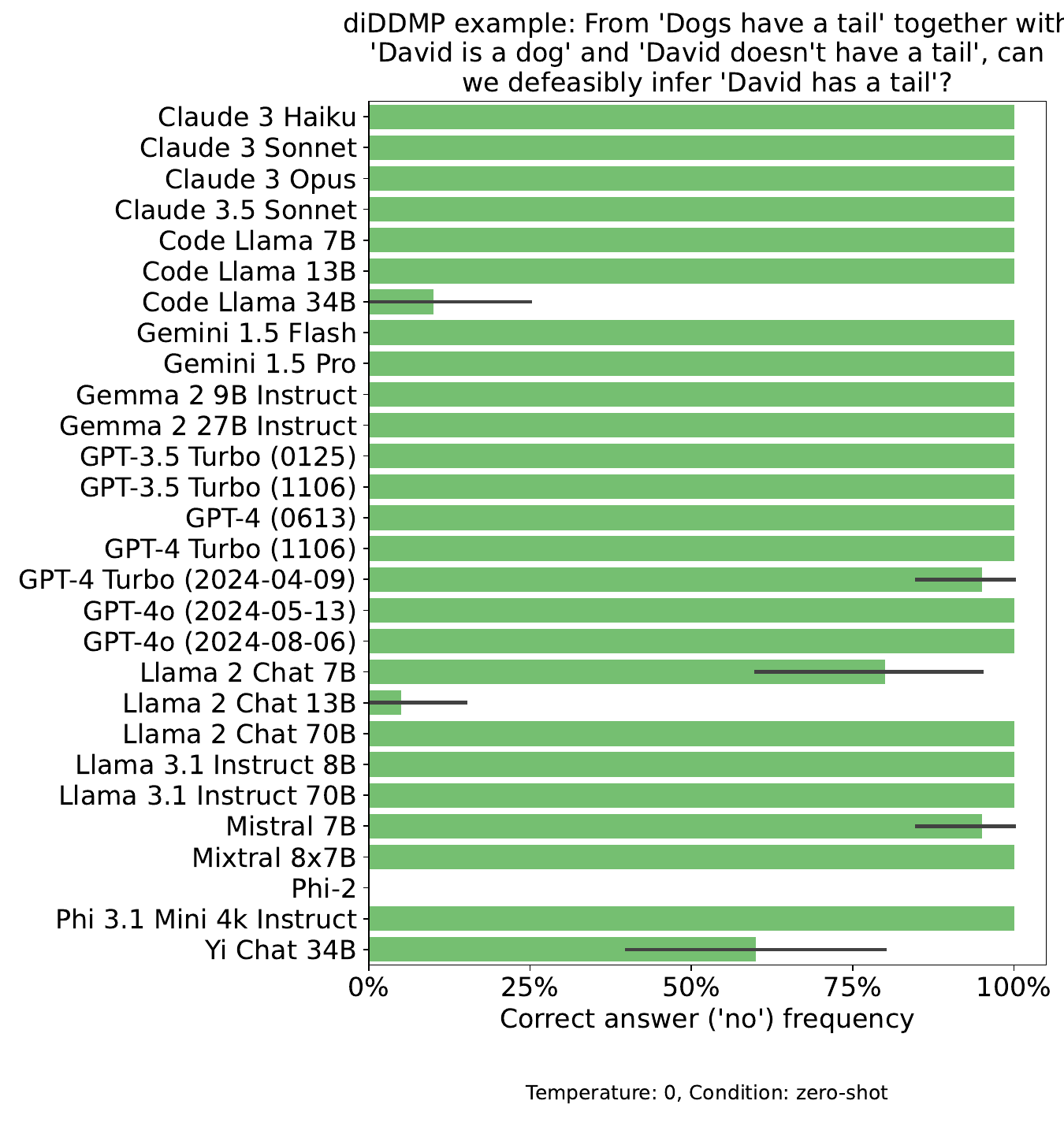}
\includegraphics[scale=.25]{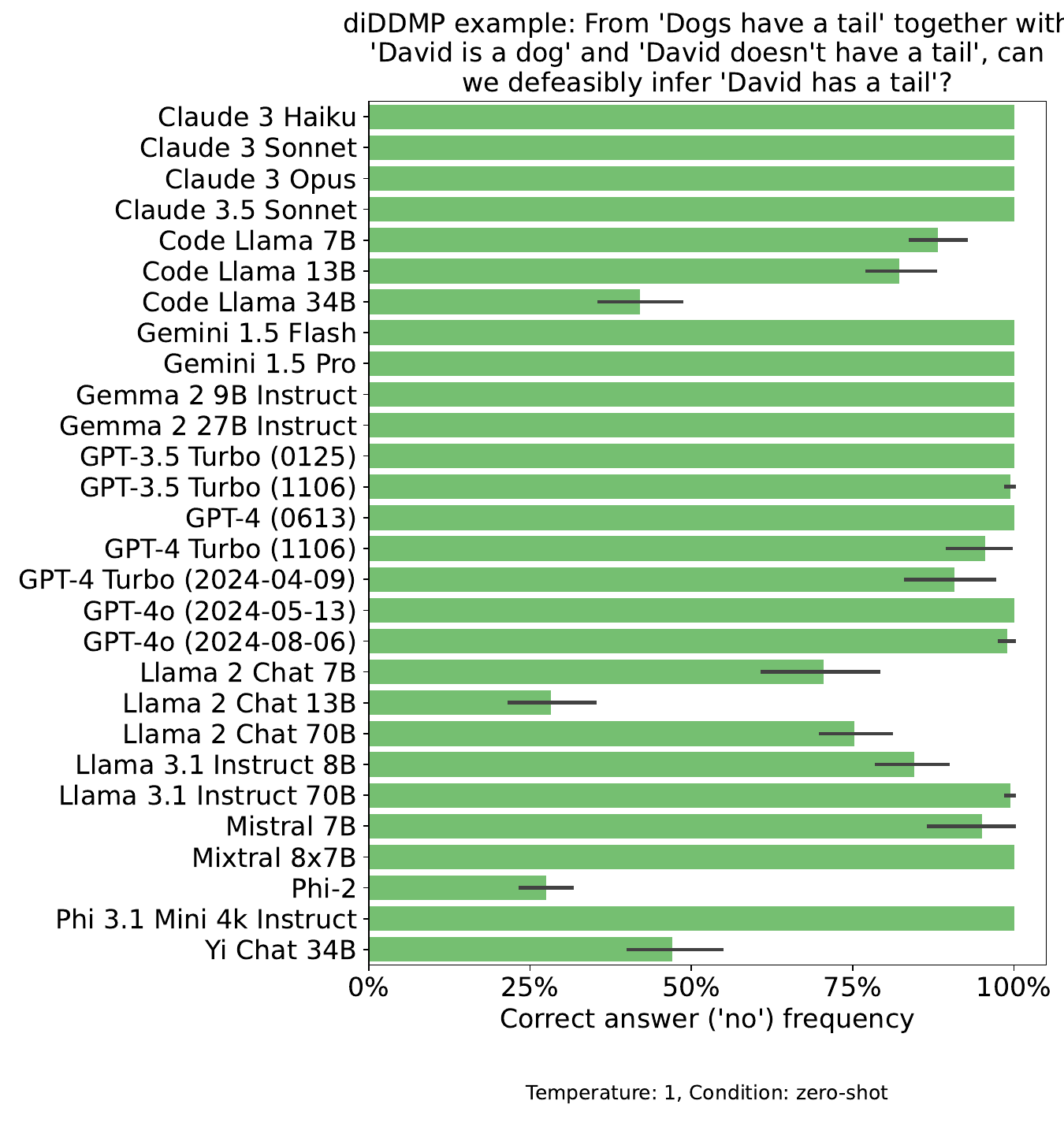}
\end{center}

\subsubsection{Nixon Diamond (ND)}

\begin{center}
\includegraphics[scale=.24]{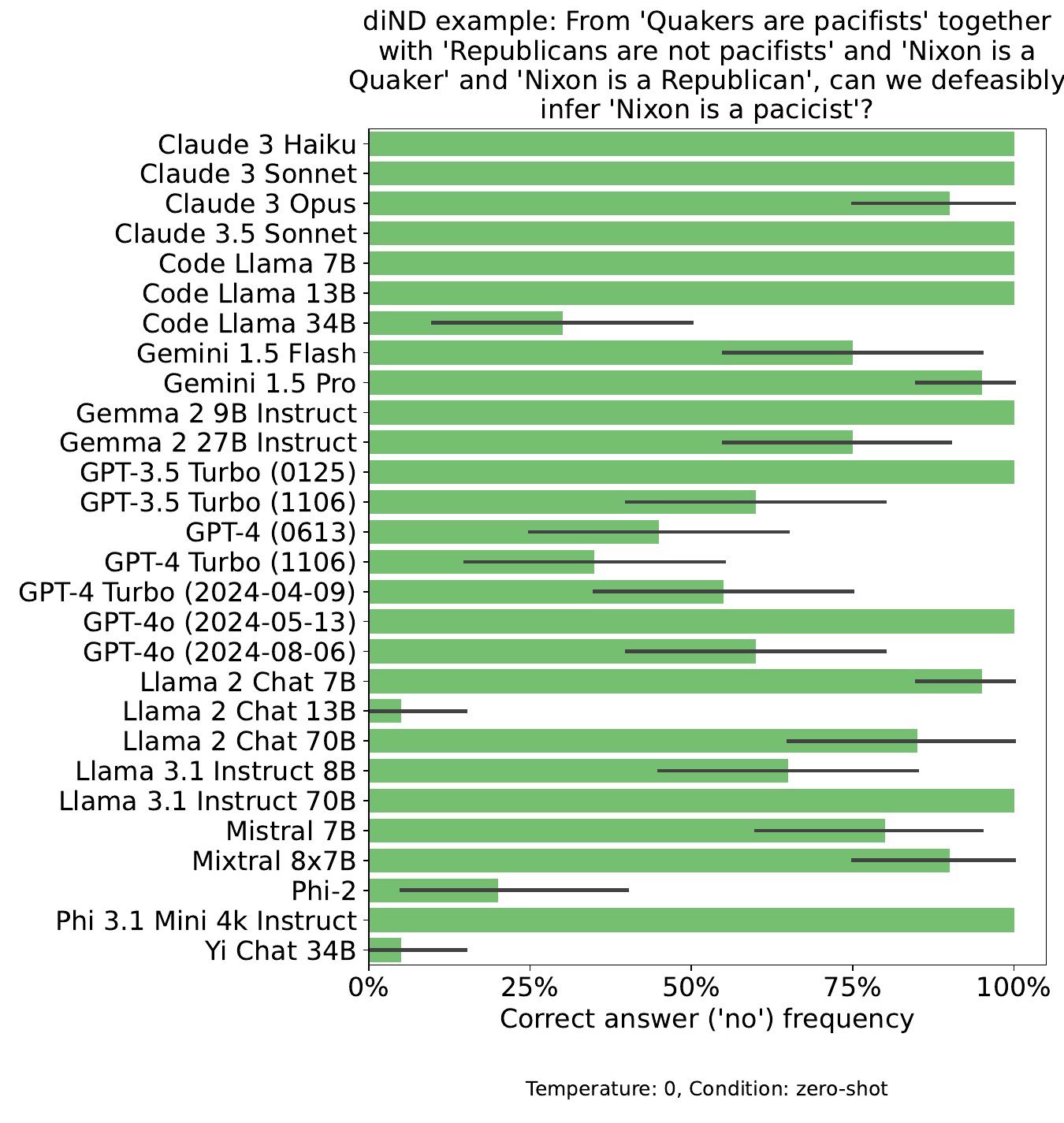}
\includegraphics[scale=.24]{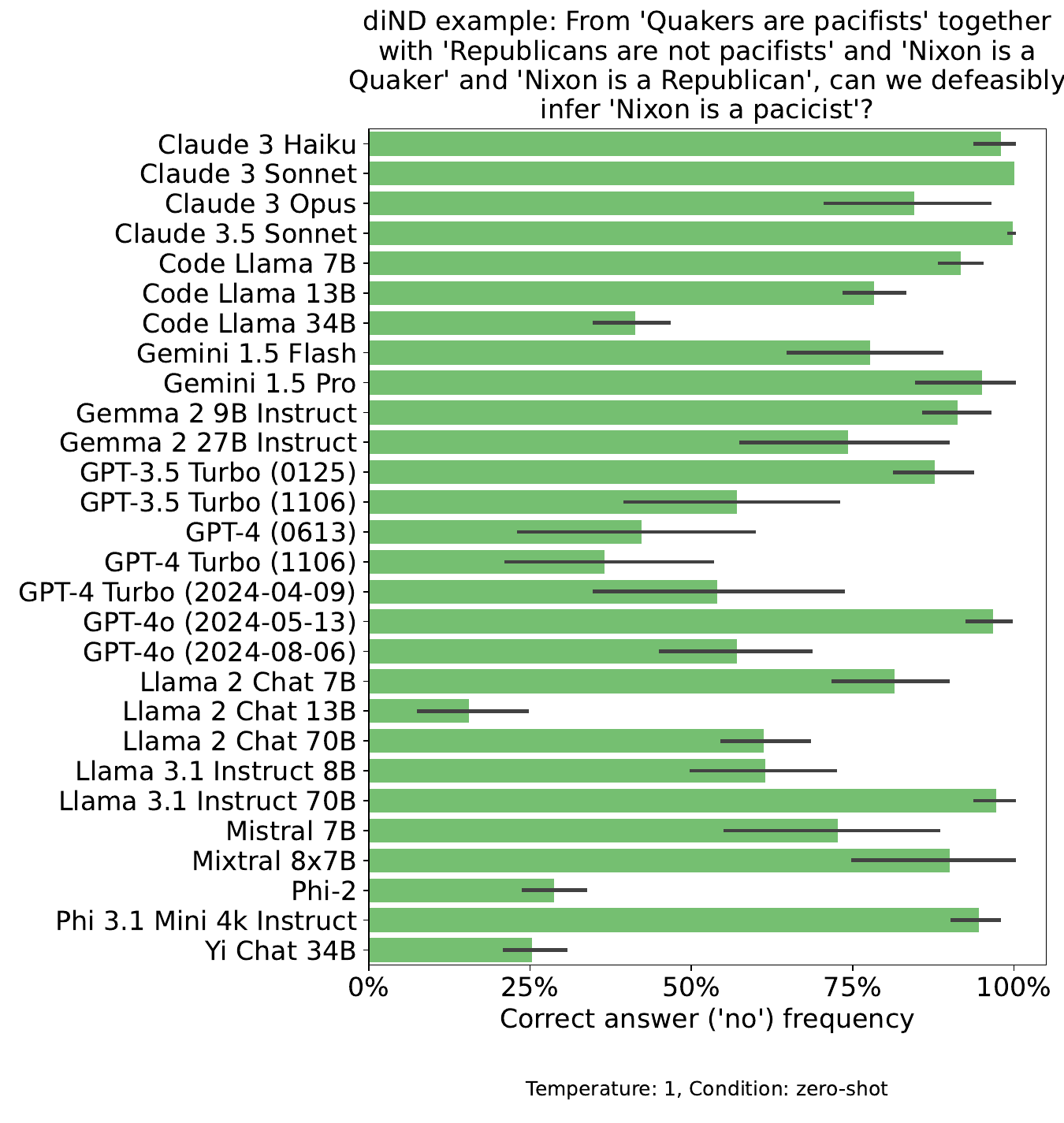}
\end{center}

\subsubsection{Extended Nixon Diamond (END)}

\begin{center}
\includegraphics[scale=.24]{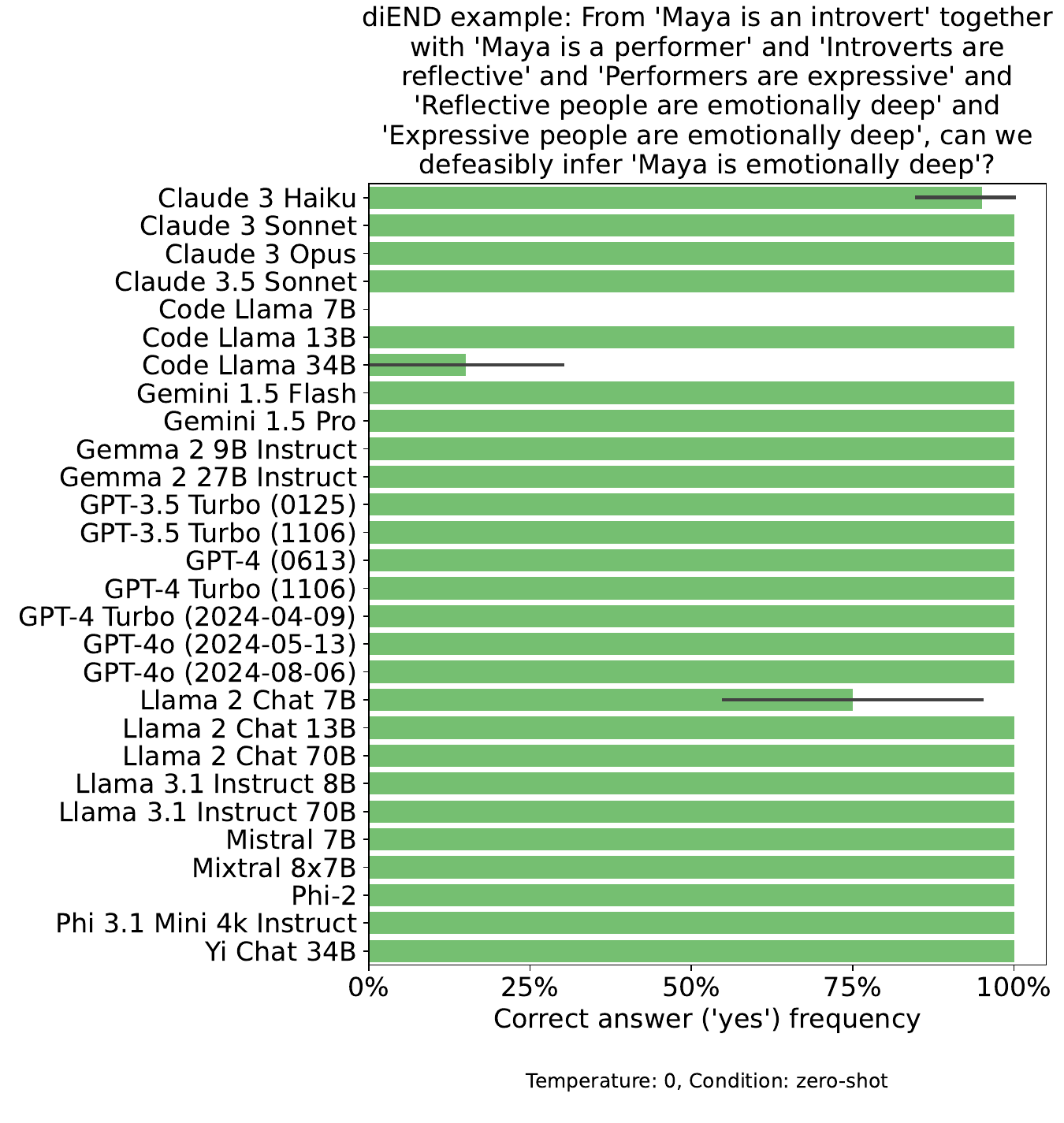}
\includegraphics[scale=.24]{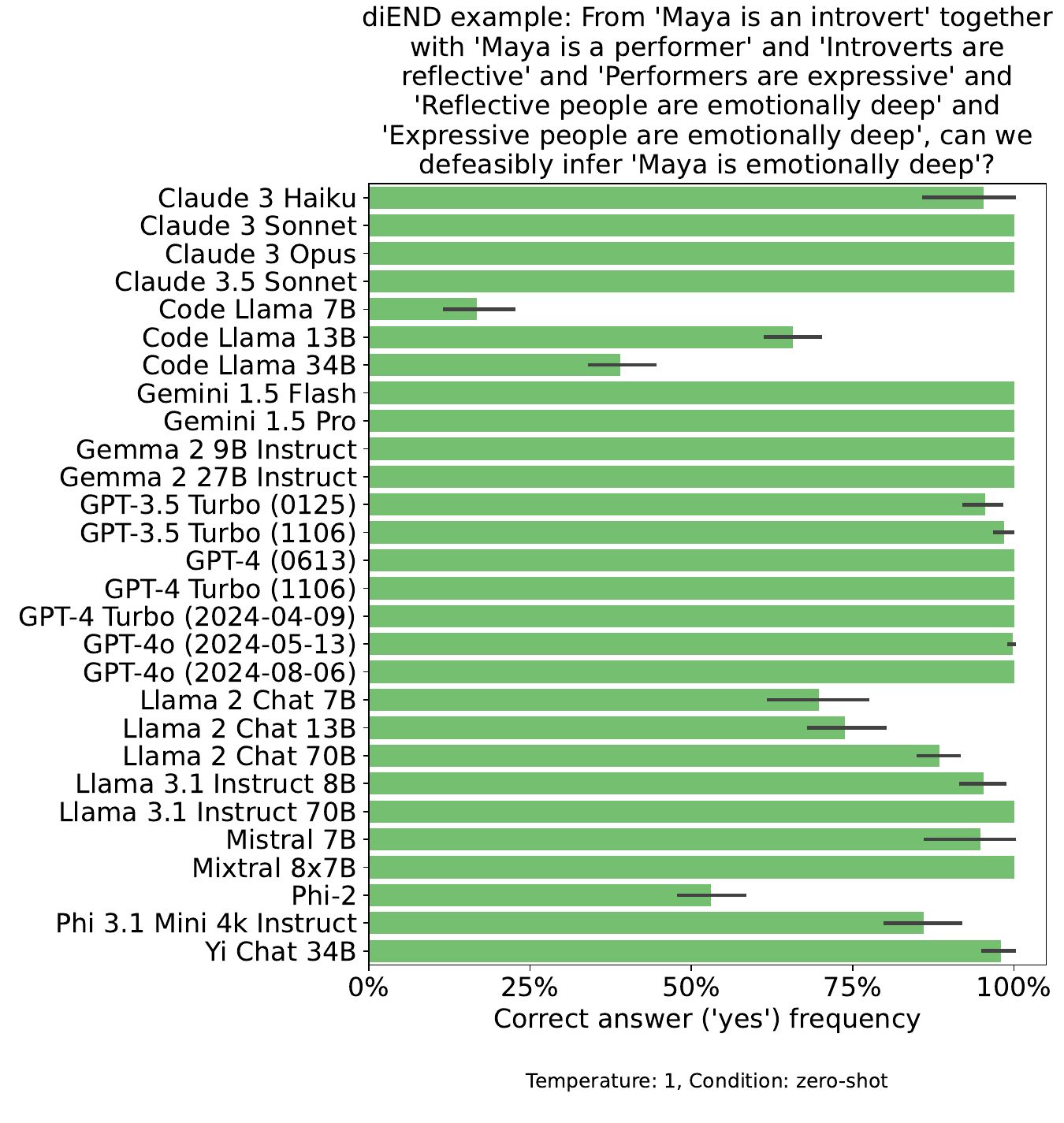}
\end{center}

\subsubsection{Penguin Principle (PP)}

\begin{center}
\includegraphics[scale=.24]{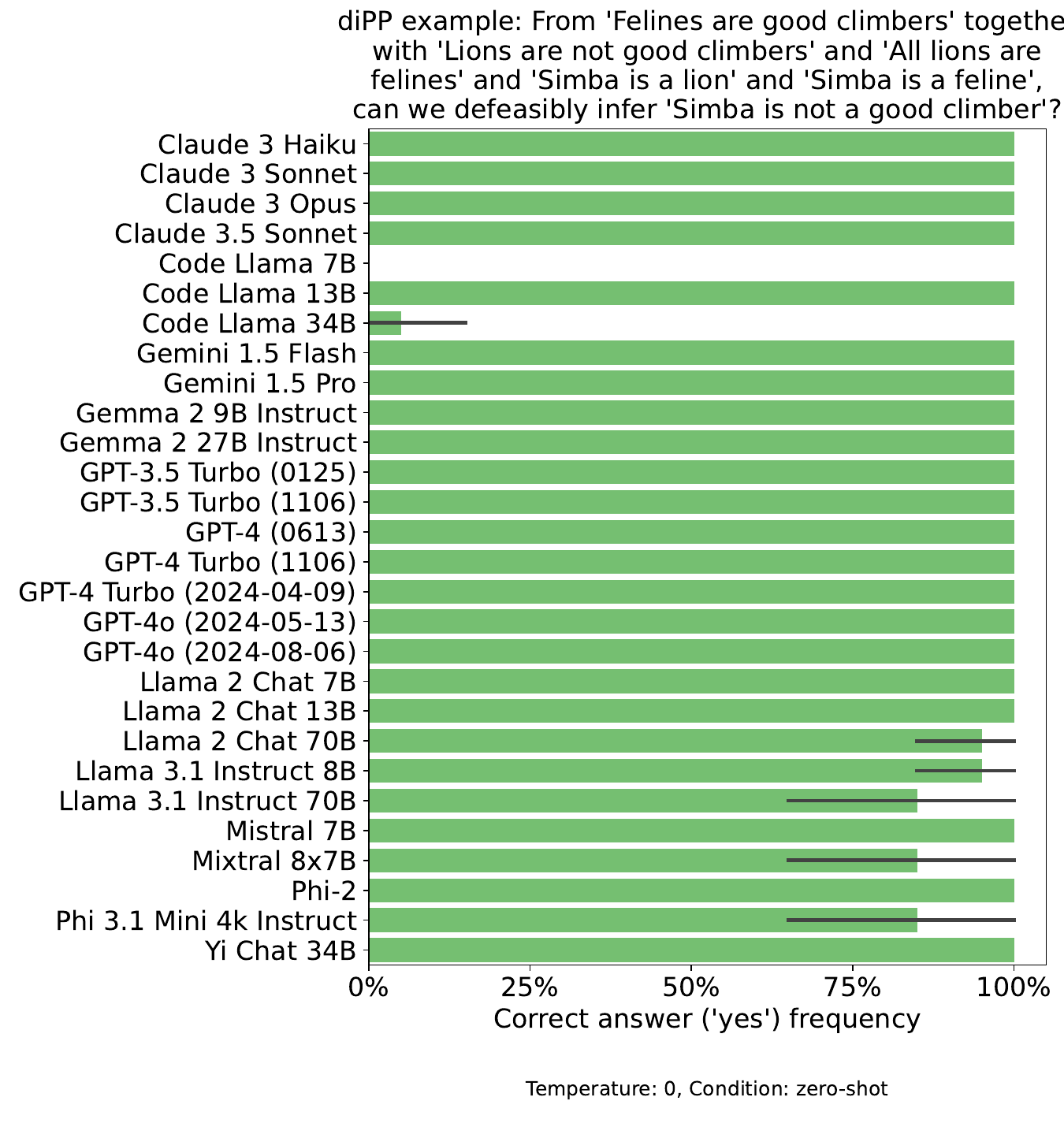}
\includegraphics[scale=.24]{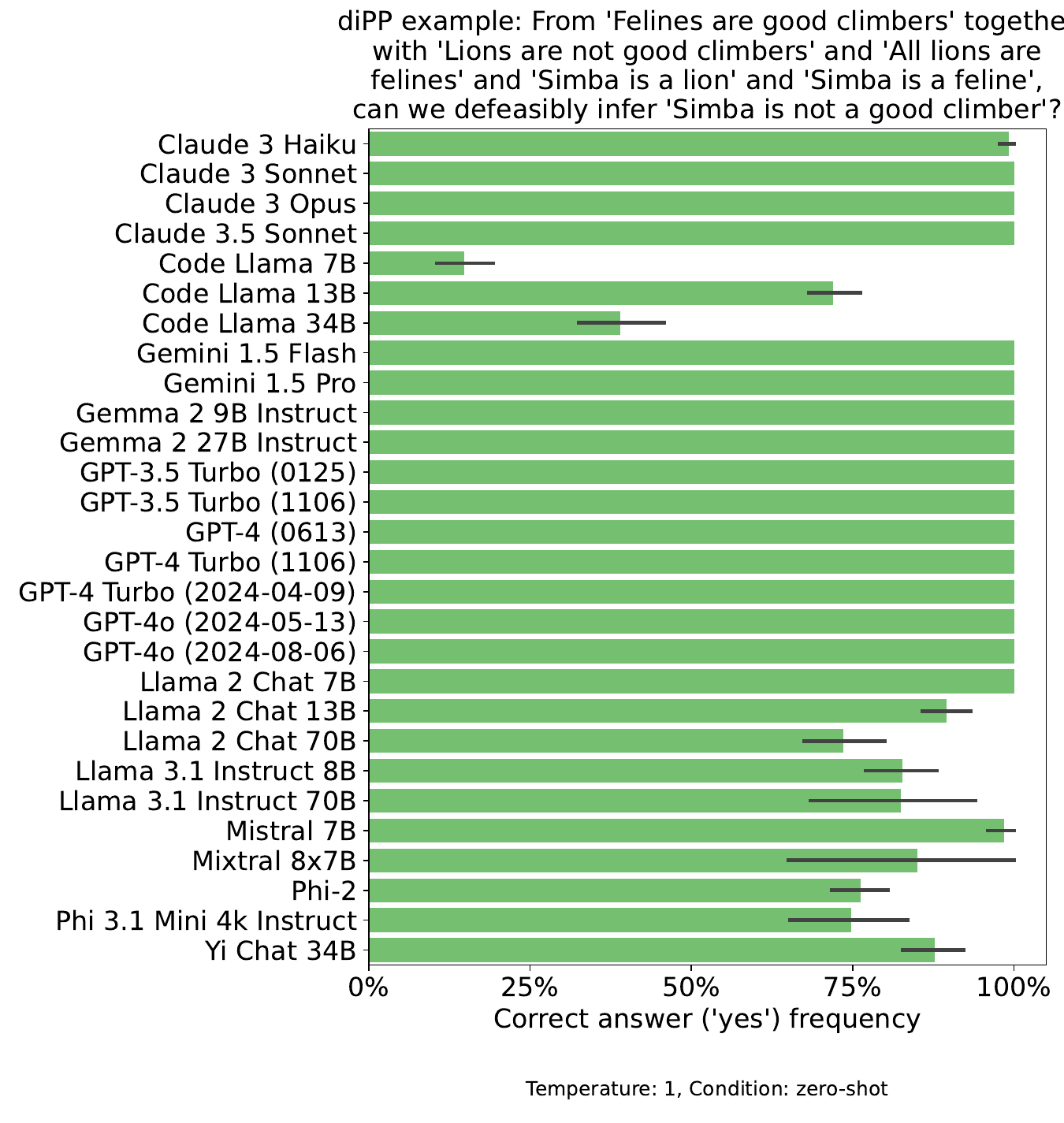}
\end{center}

\subsubsection{Weakened Penguin Principle (WPP)}

\begin{center}
\includegraphics[scale=.24]{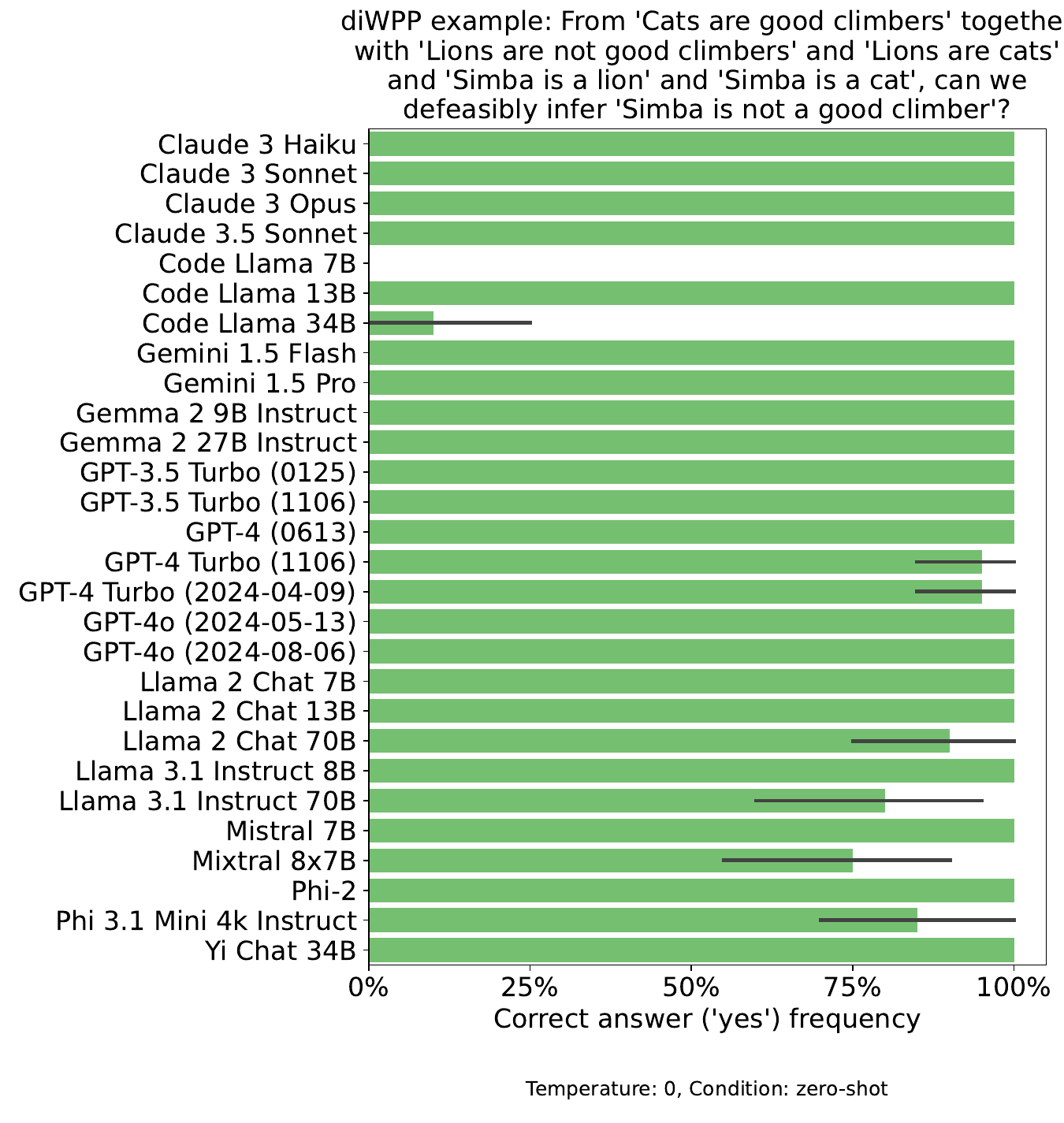}
\includegraphics[scale=.24]{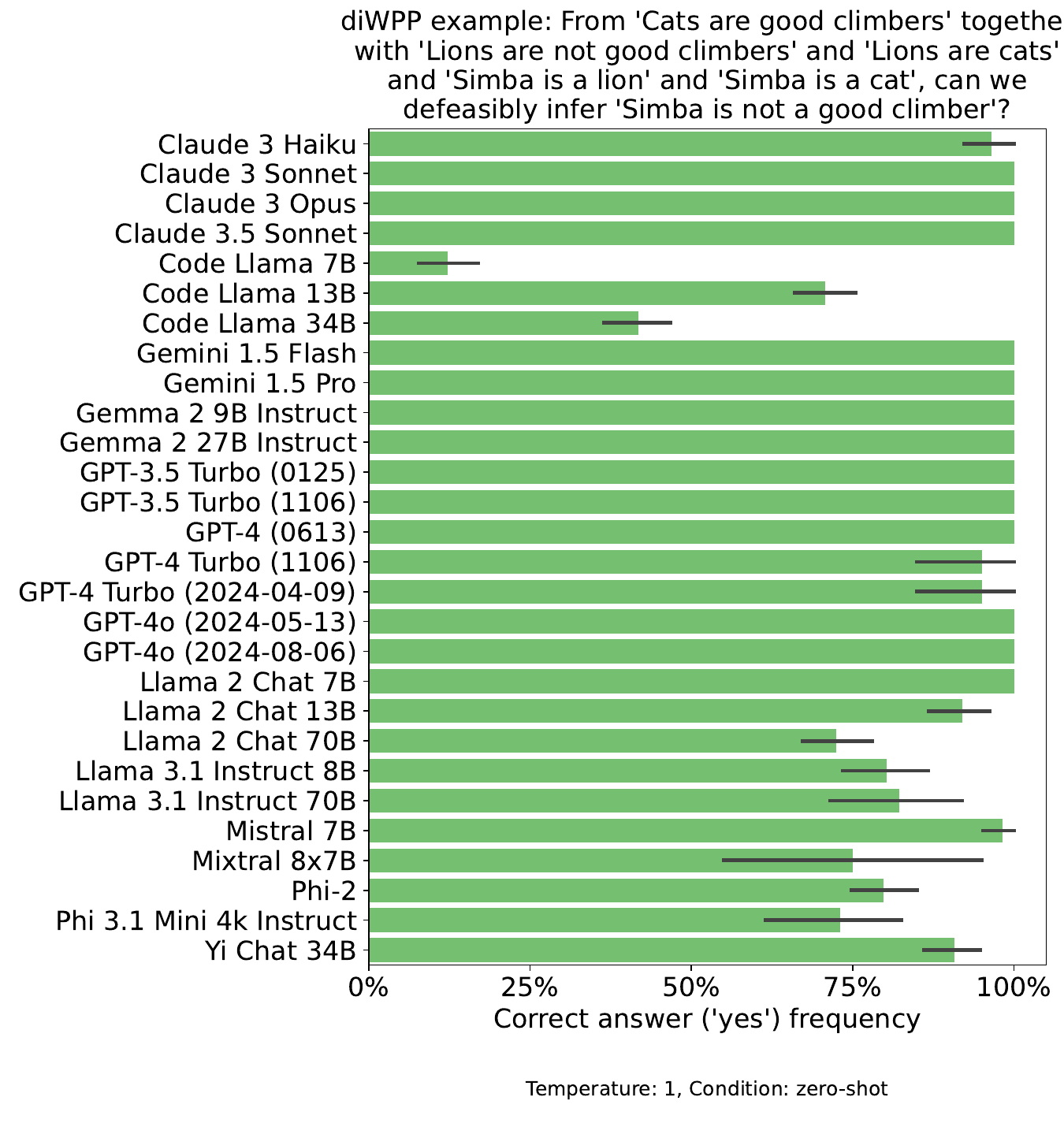}
\end{center}

\subsubsection{Defeasible Transitivity (DT)}

\begin{center}
\includegraphics[scale=.24]{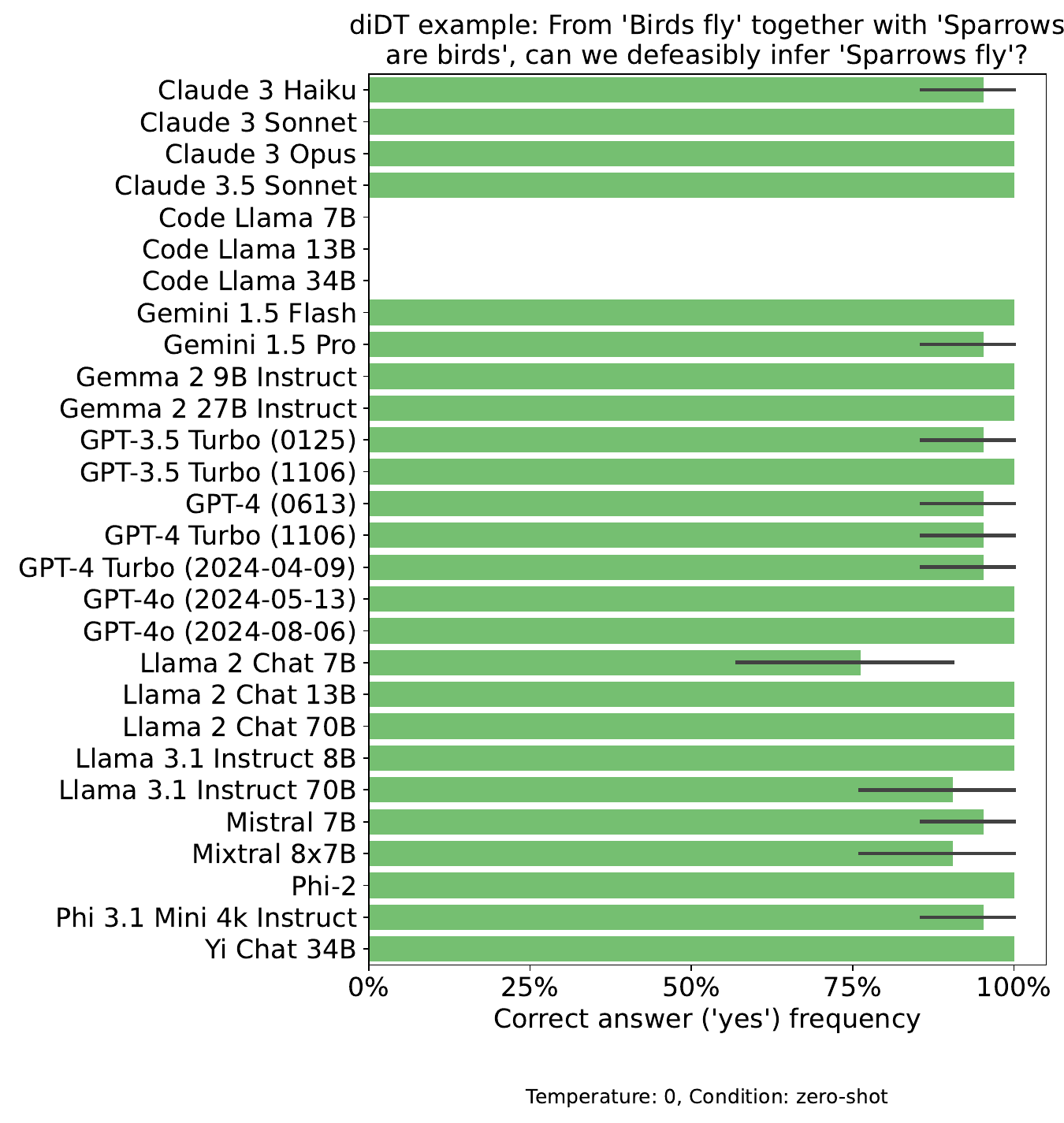}
\includegraphics[scale=.24]{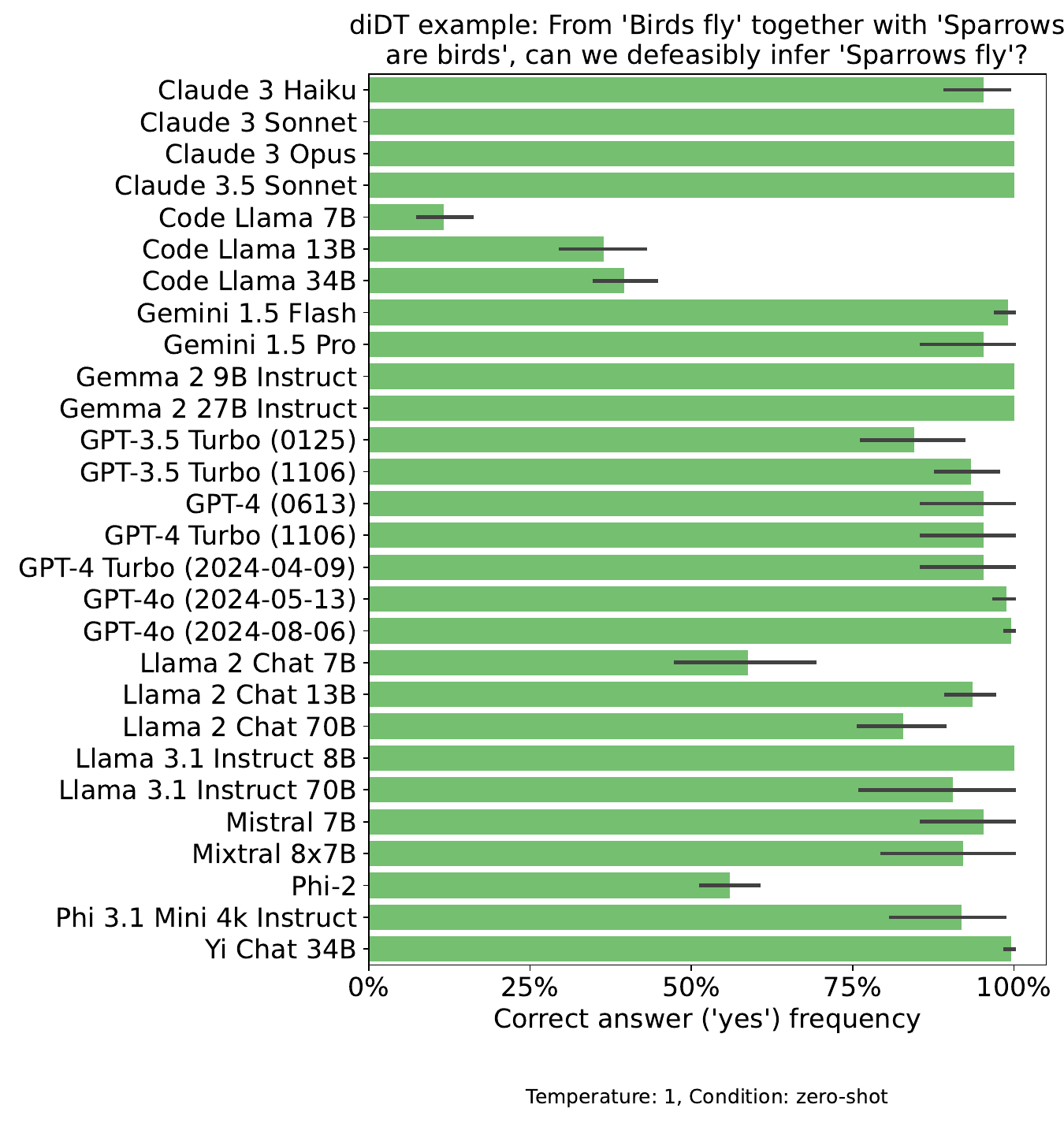}
\end{center}

\subsubsection{Defeasible Strengthening the Antecedent (DSA)}

\begin{center}
\includegraphics[scale=.24]{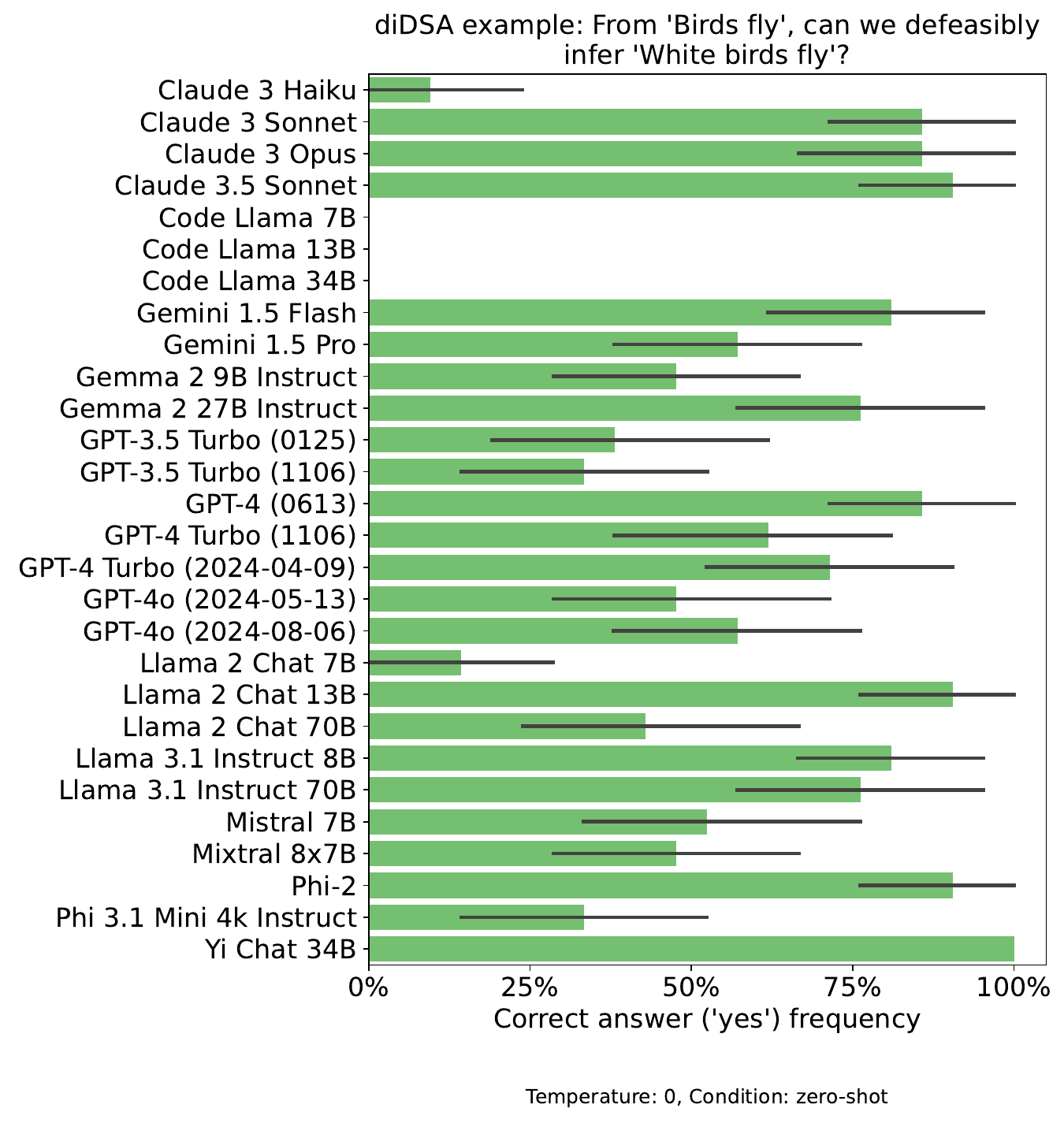}
\includegraphics[scale=.24]{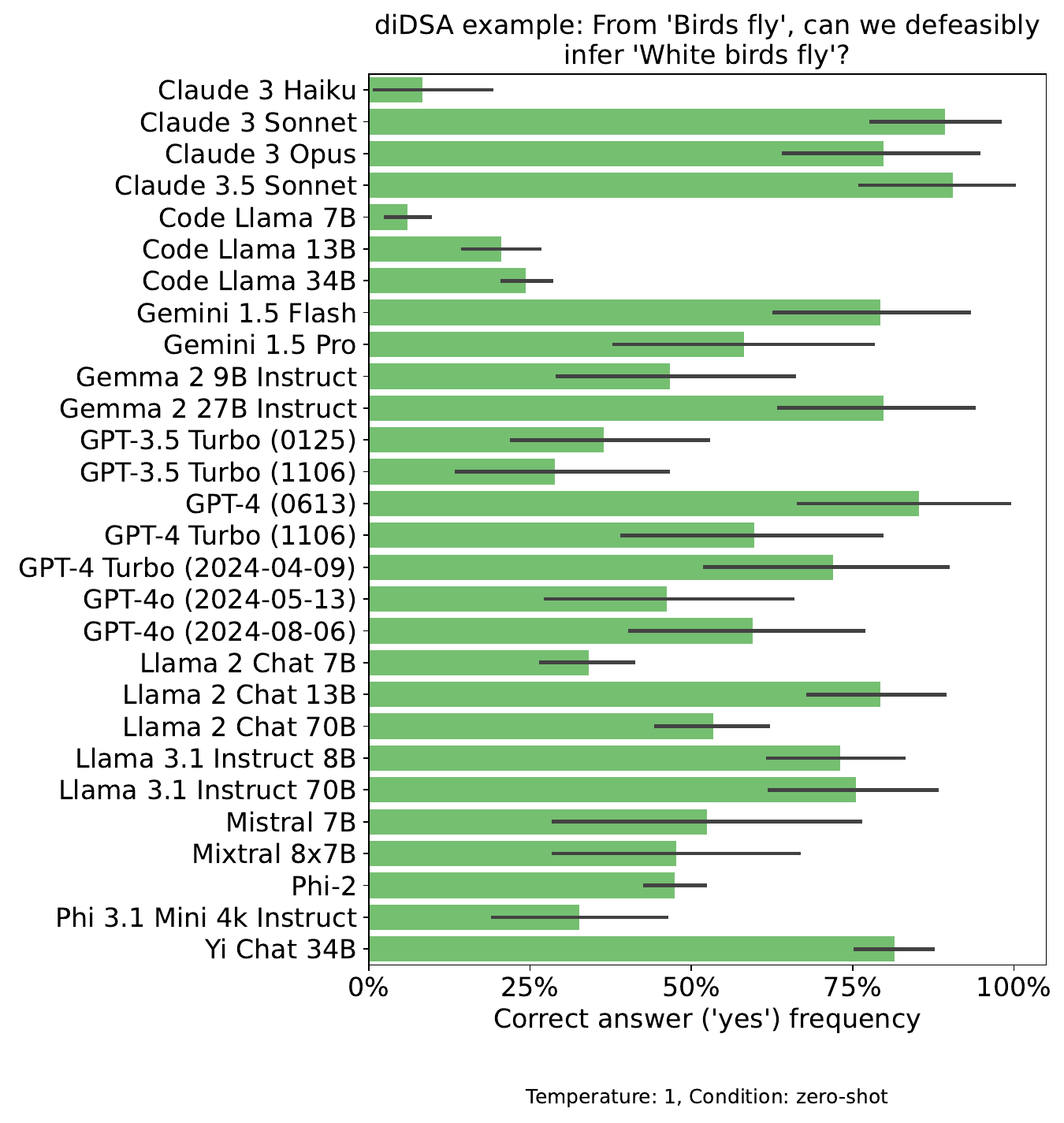}
\end{center}

\subsubsection{Independence (I)}

\begin{center}
\includegraphics[scale=.24]{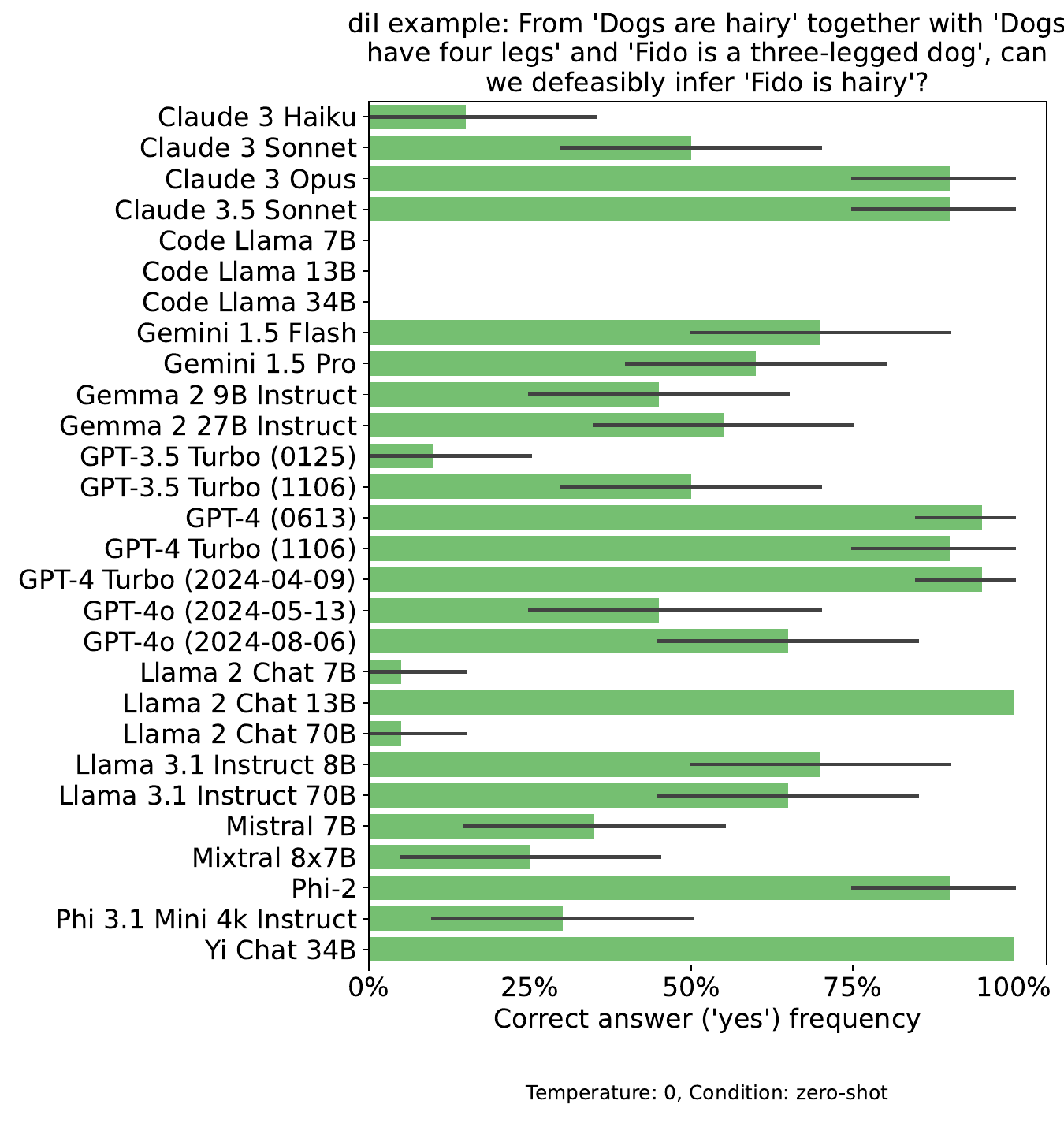}
\includegraphics[scale=.24]{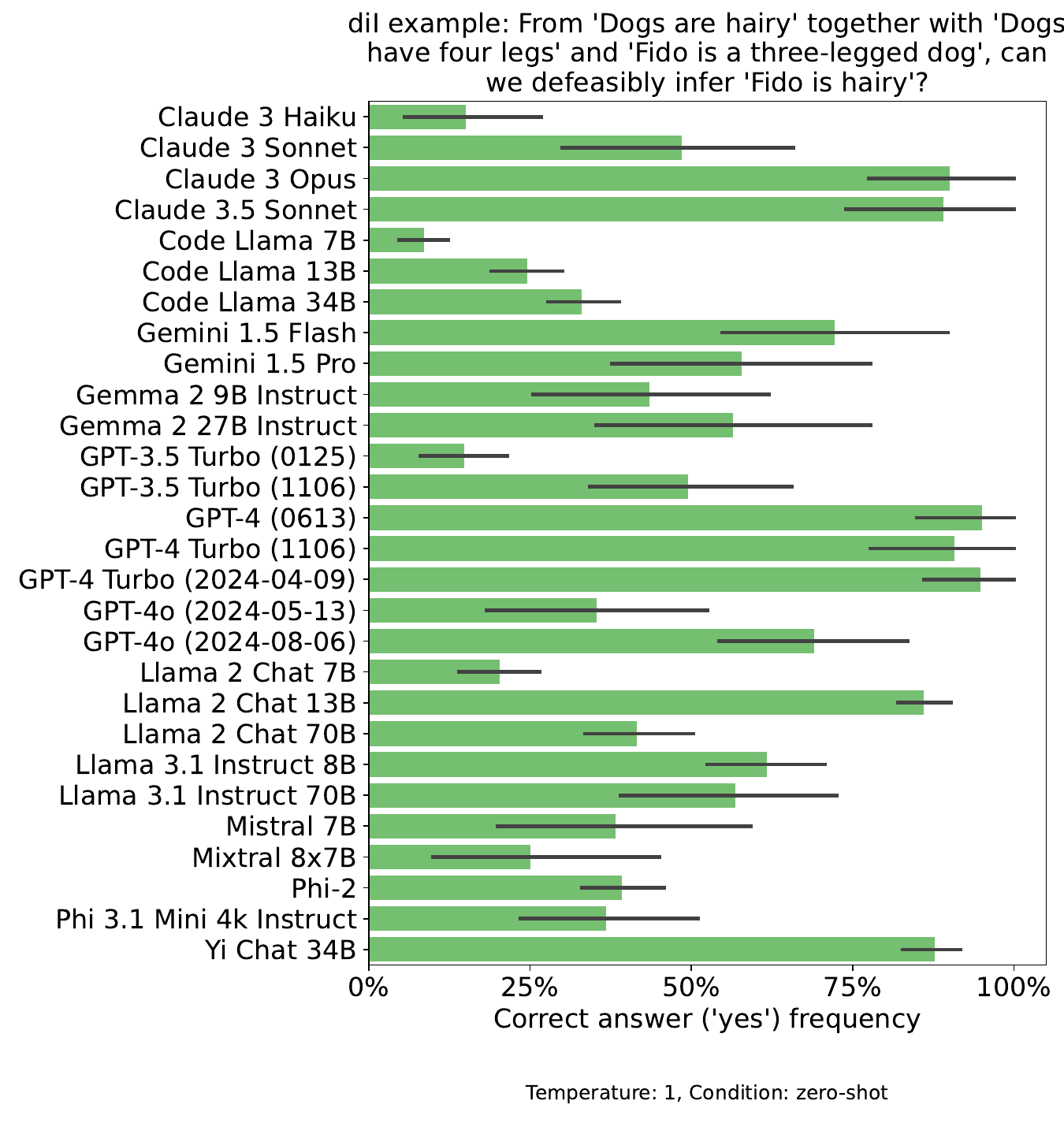}
\end{center}

\subsubsection{Basic Default Reasoning (BDR)}

\begin{center}
\includegraphics[scale=.24]{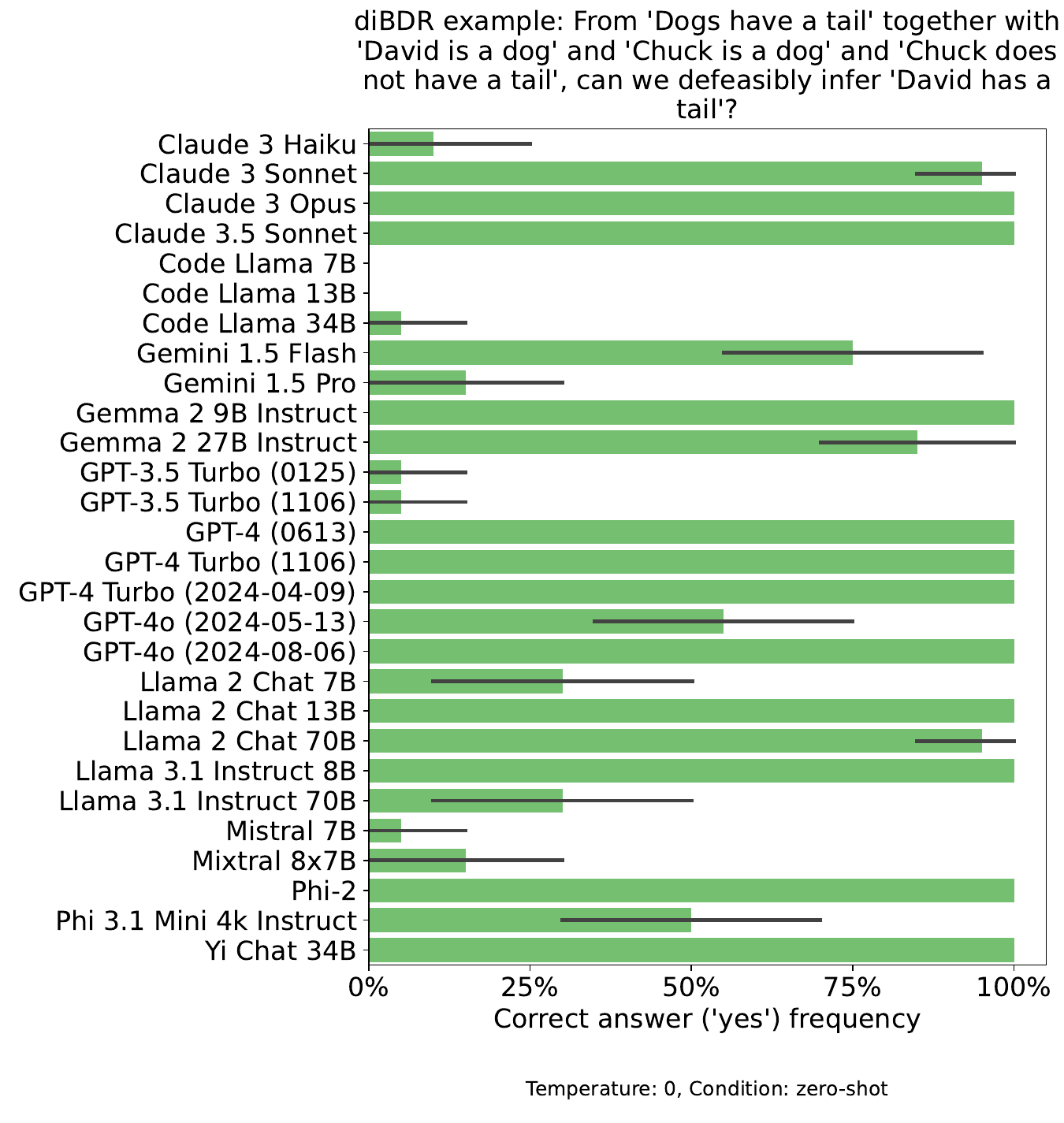}
\includegraphics[scale=.24]{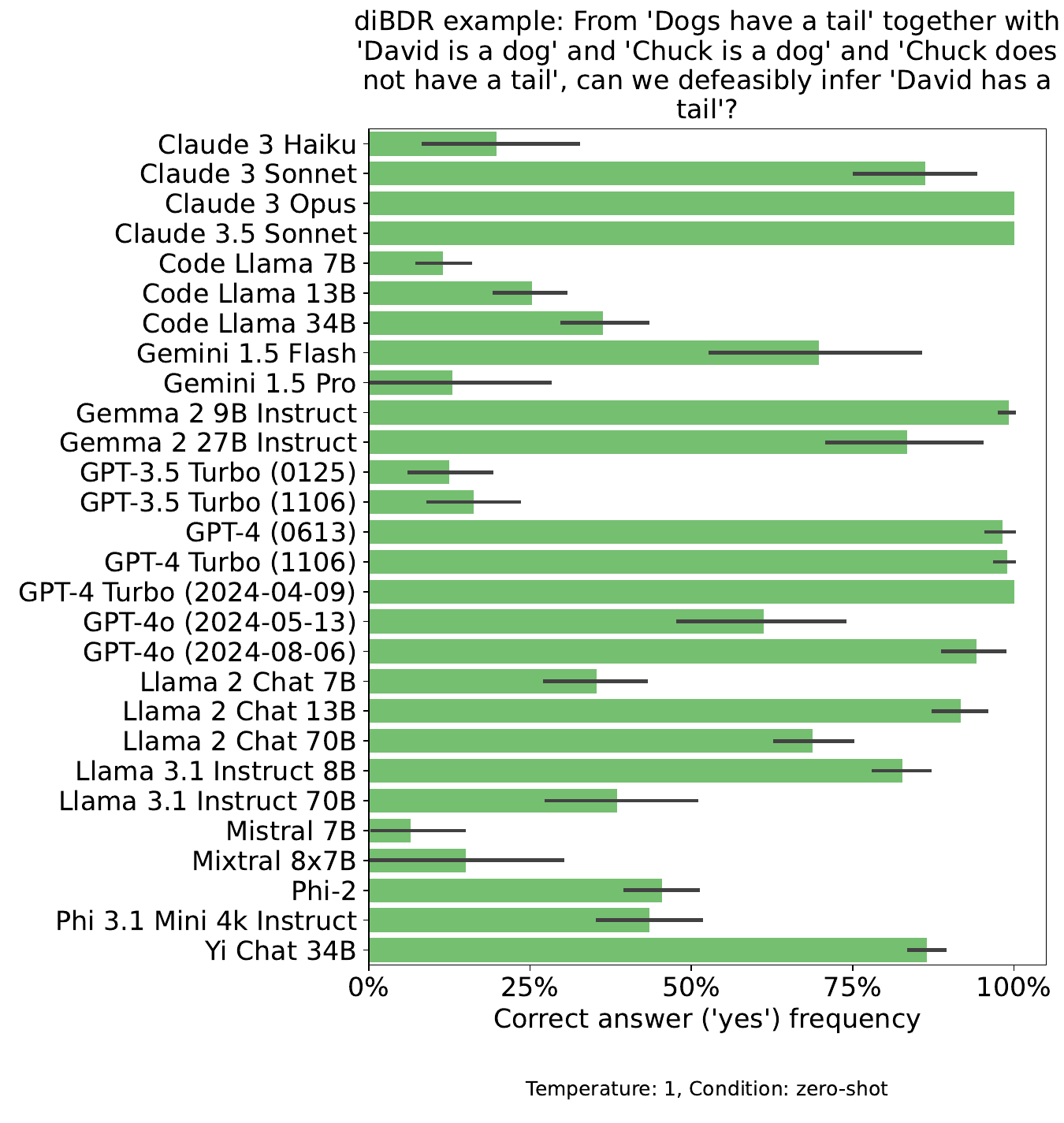}
\end{center}

\subsubsection{Default Reasoning with Irrelevant Information (DRII)}

\begin{center}
\includegraphics[scale=.24]{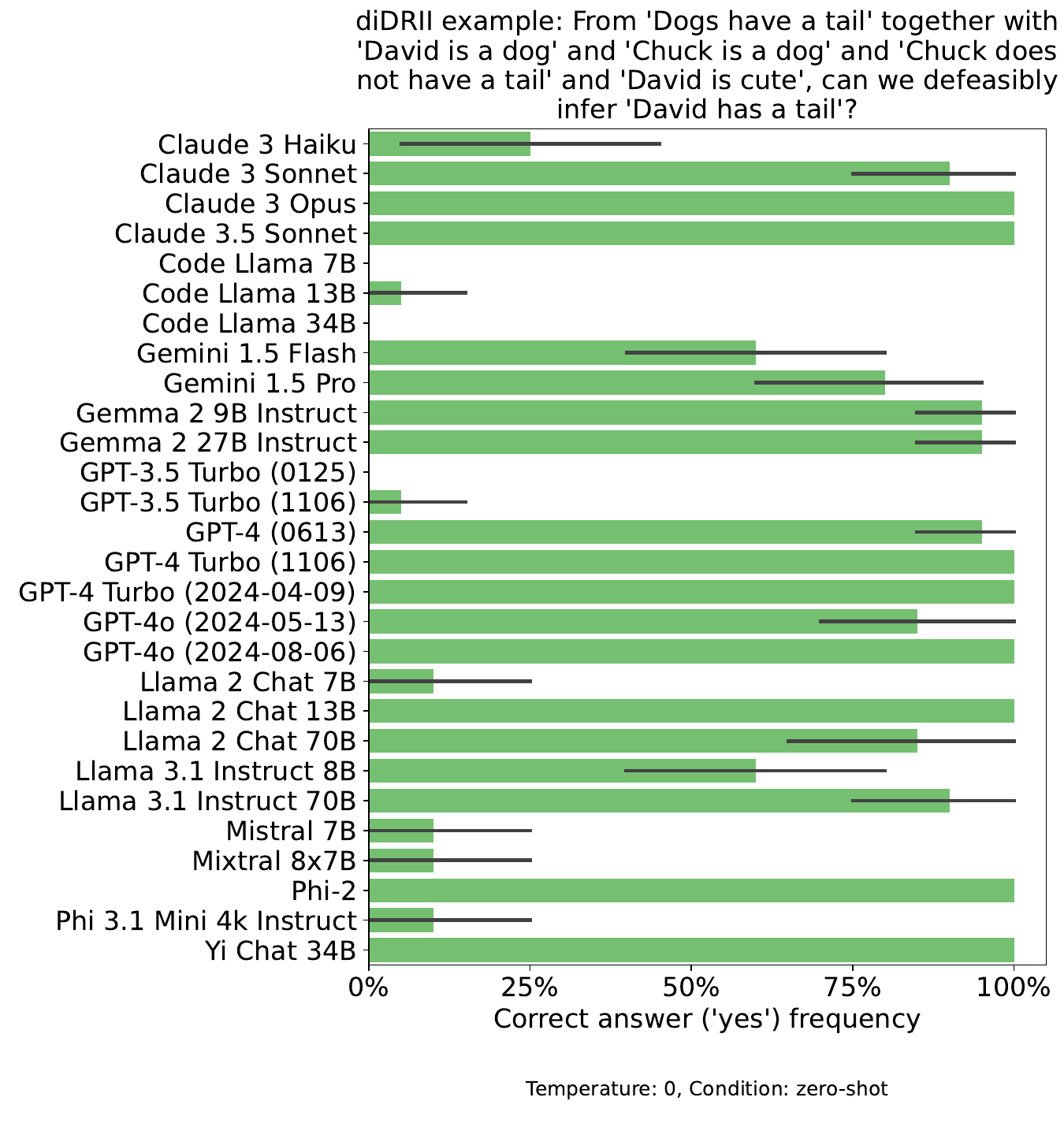}
\includegraphics[scale=.24]{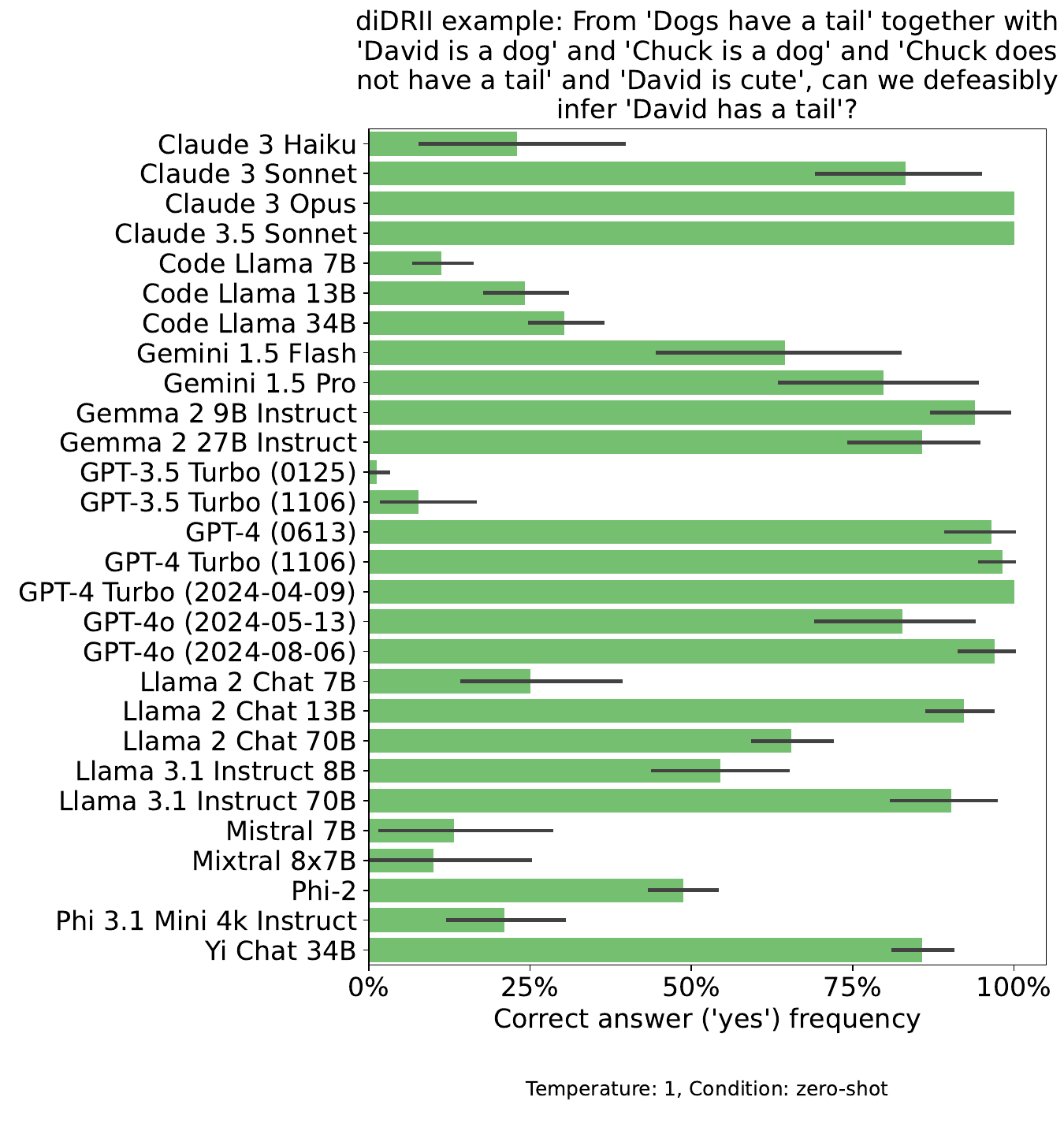}
\end{center}

\subsubsection{Default Reasoning with Several Defaults (DRSD)}

\begin{center}
\includegraphics[scale=.24]{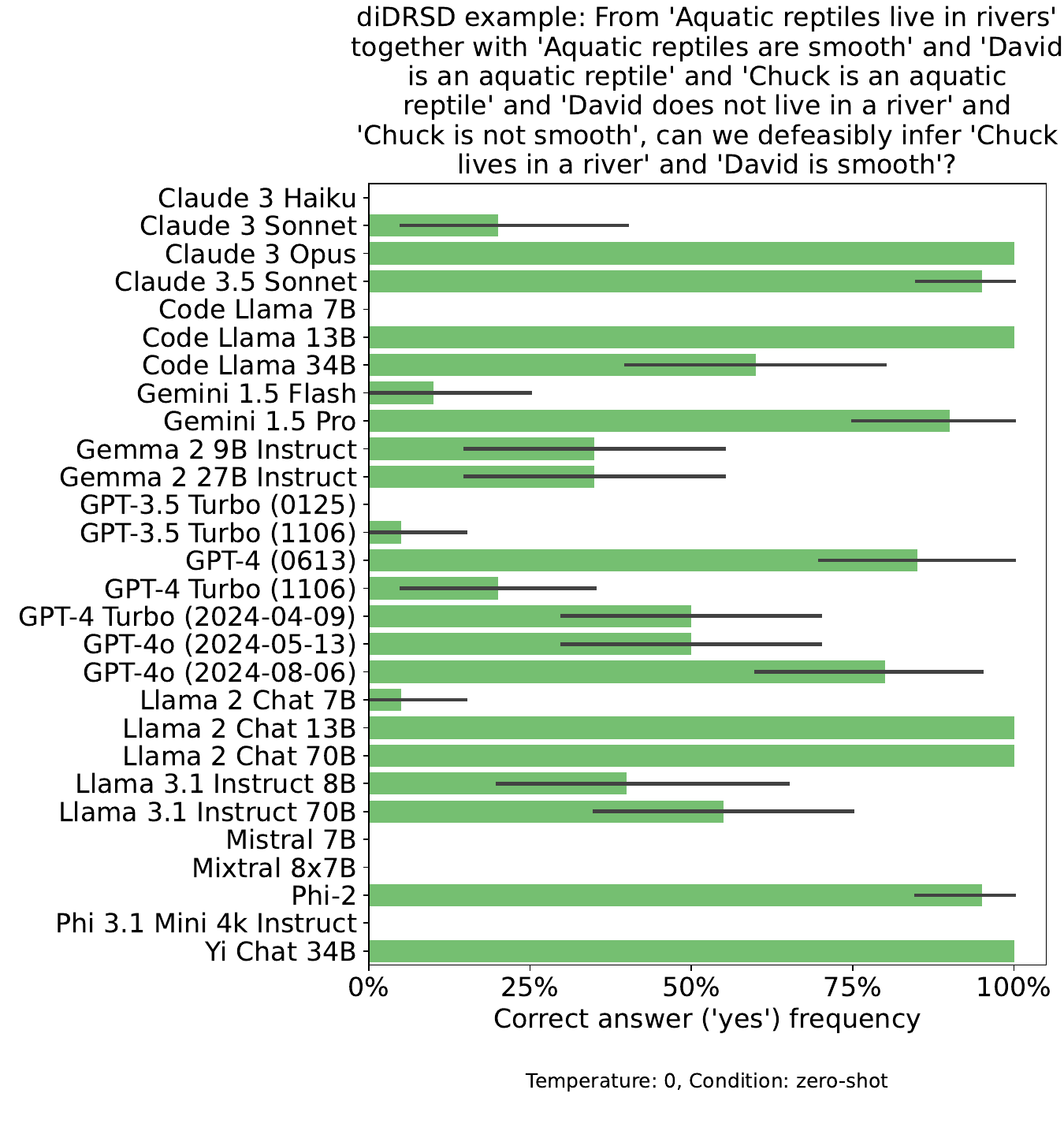}
\includegraphics[scale=.24]{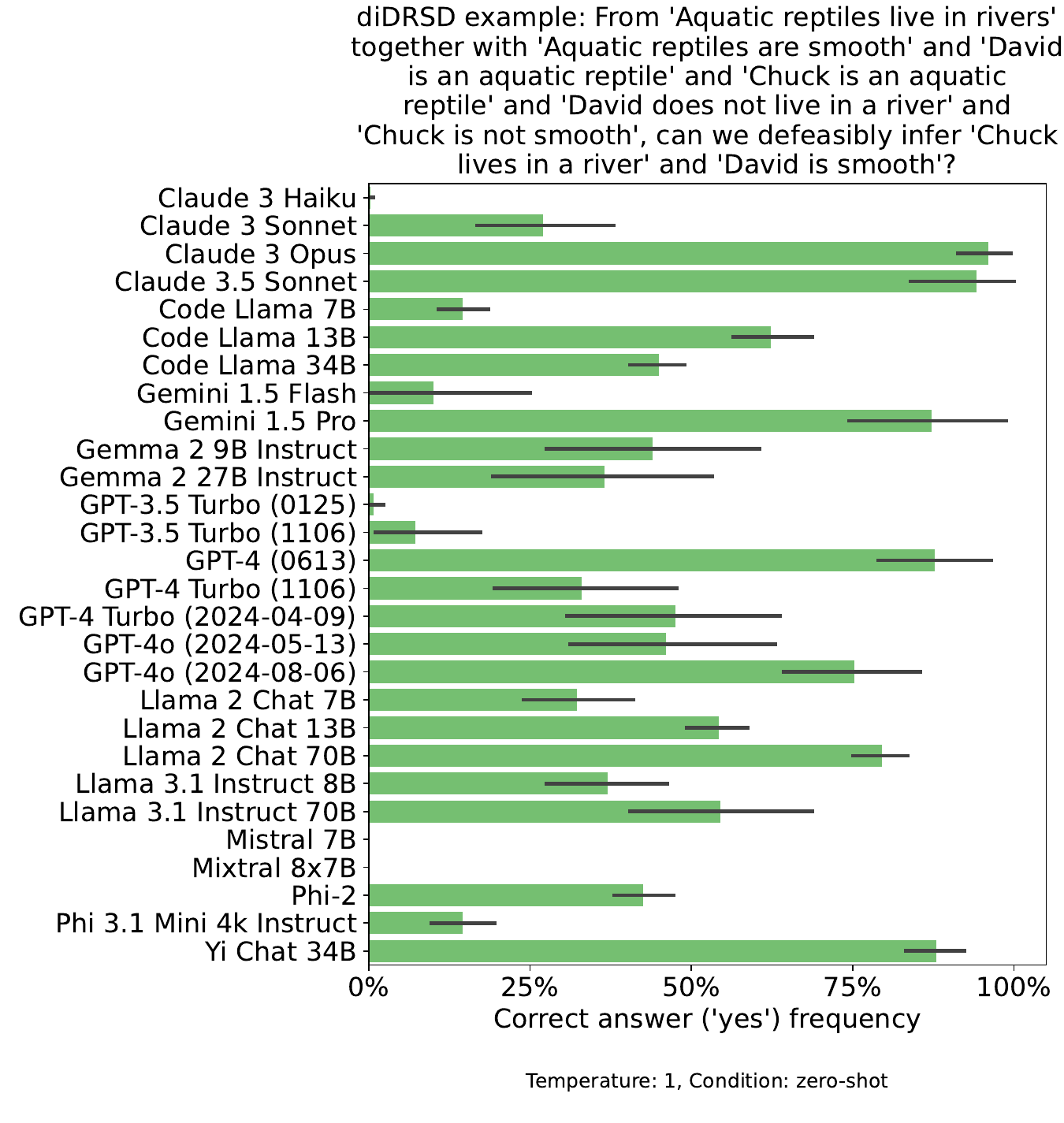}
\end{center}

\subsubsection{Default Reasoning with Disabled Defaults (DRDD)}

\begin{center}
\includegraphics[scale=.25]{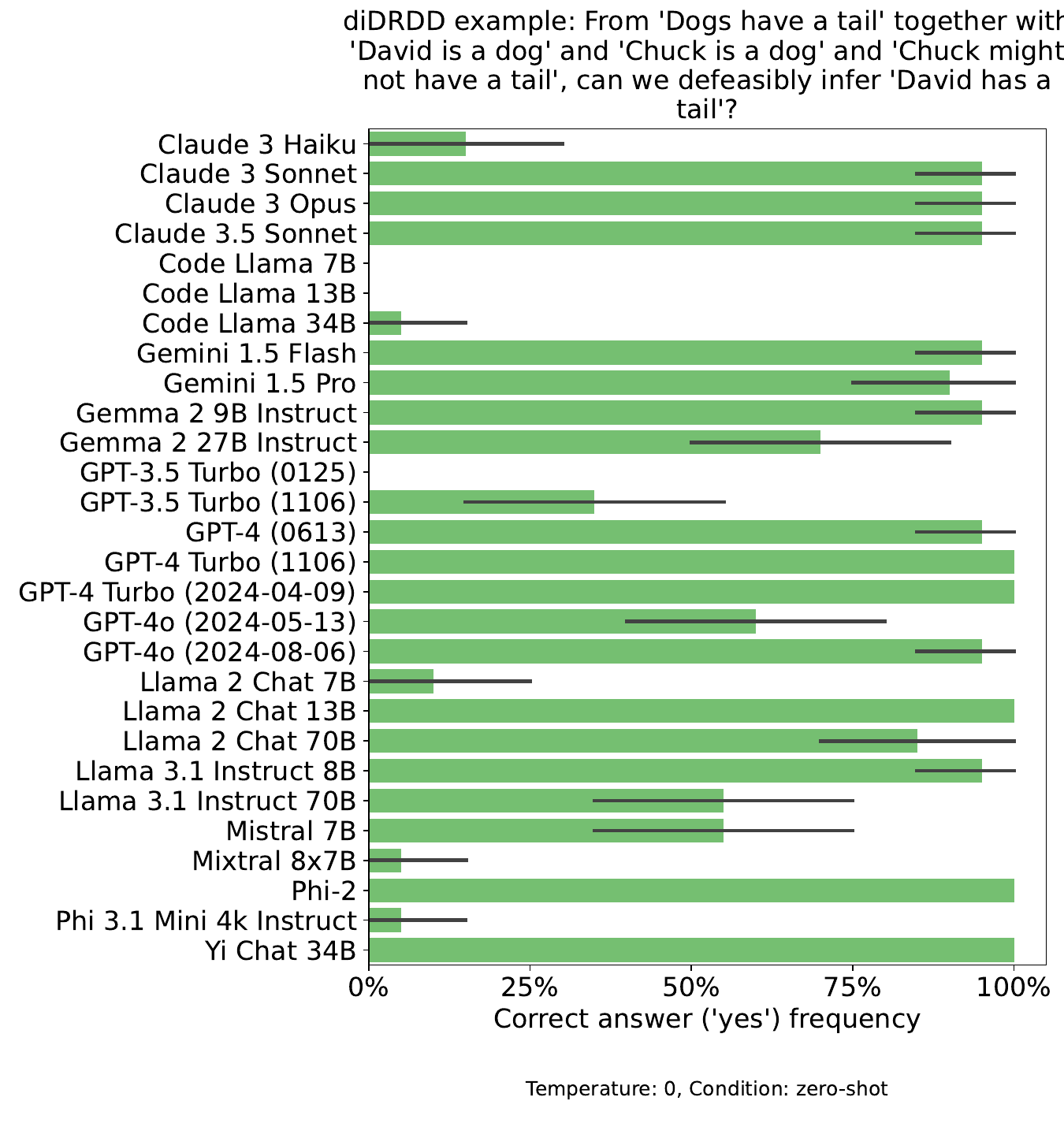}
\includegraphics[scale=.25]{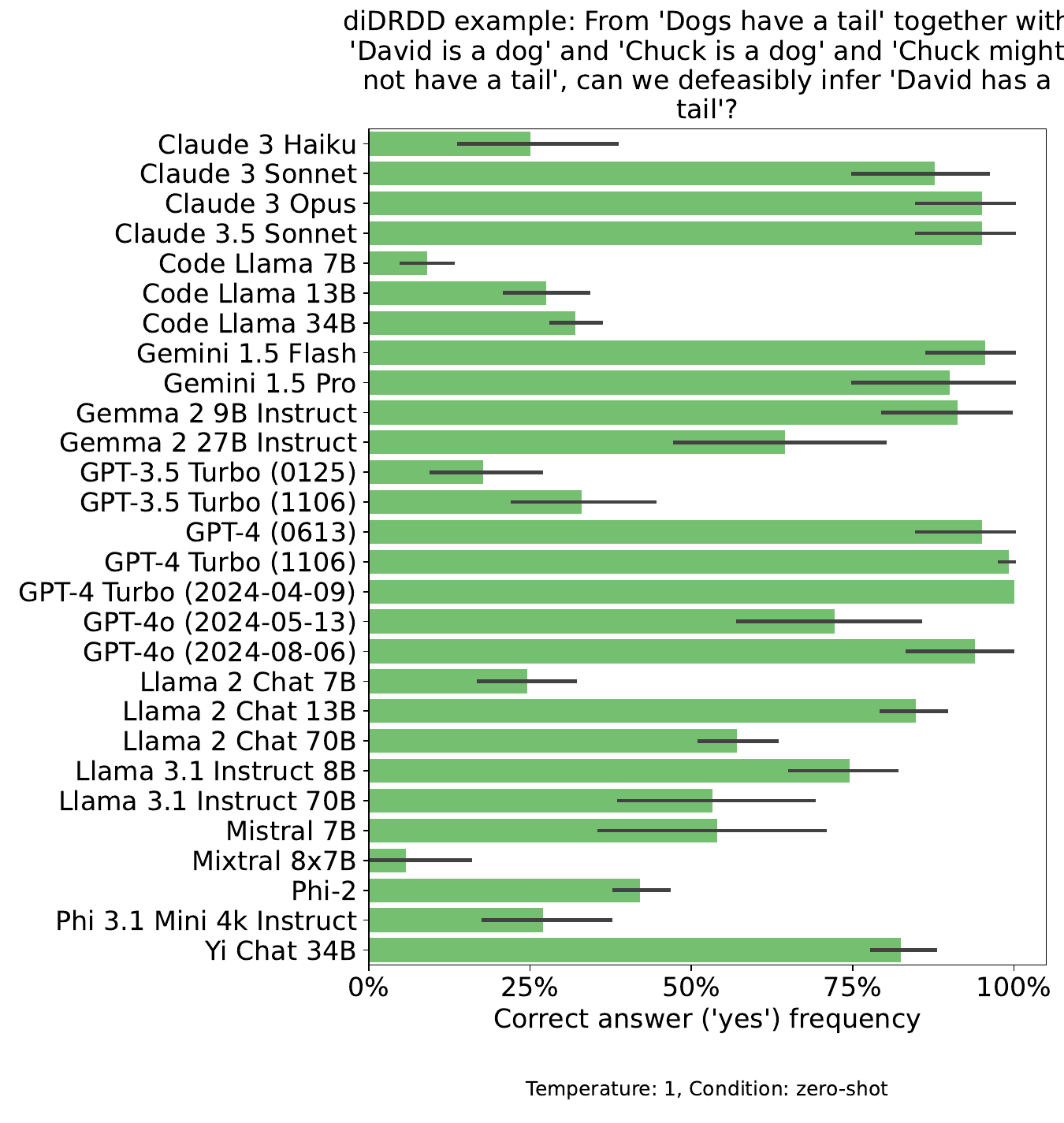}
\end{center}

\subsubsection{Fact Irrelevance (FI)}

\begin{center}
\includegraphics[scale=.25]{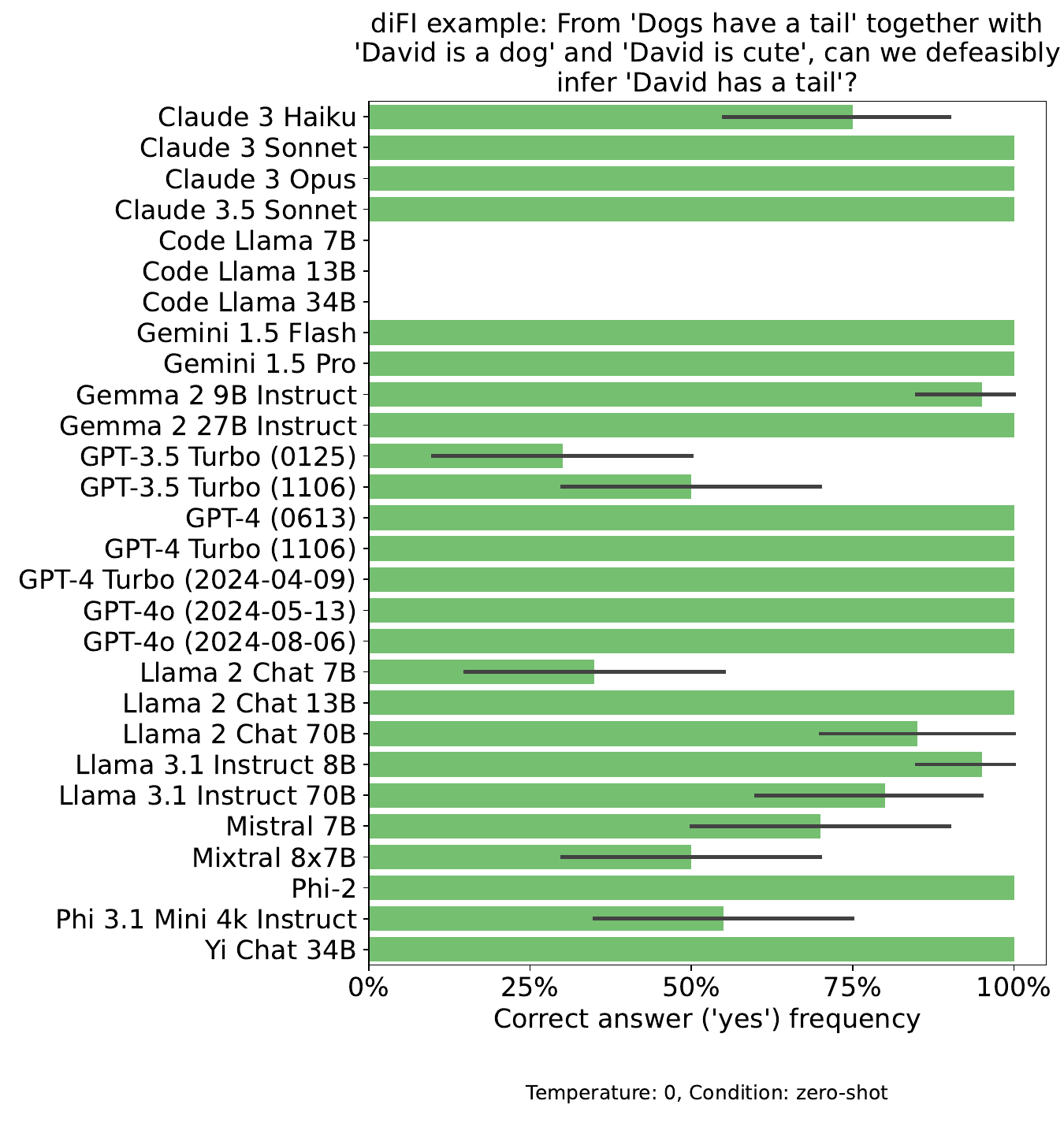}
\includegraphics[scale=.25]{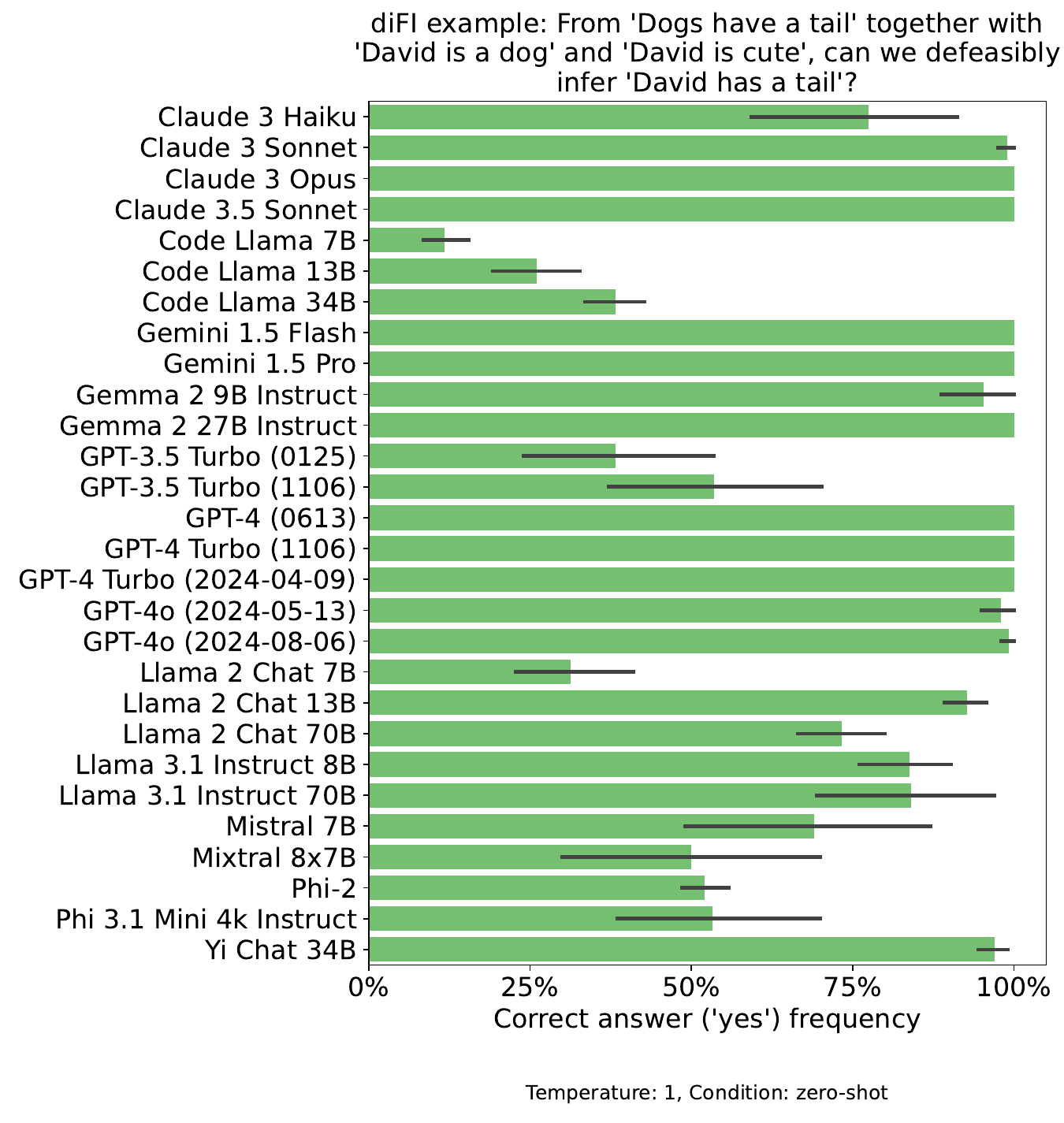}
\end{center}

\subsubsection{Default Irrelevance (DI)}

\begin{center}
\includegraphics[scale=.25]{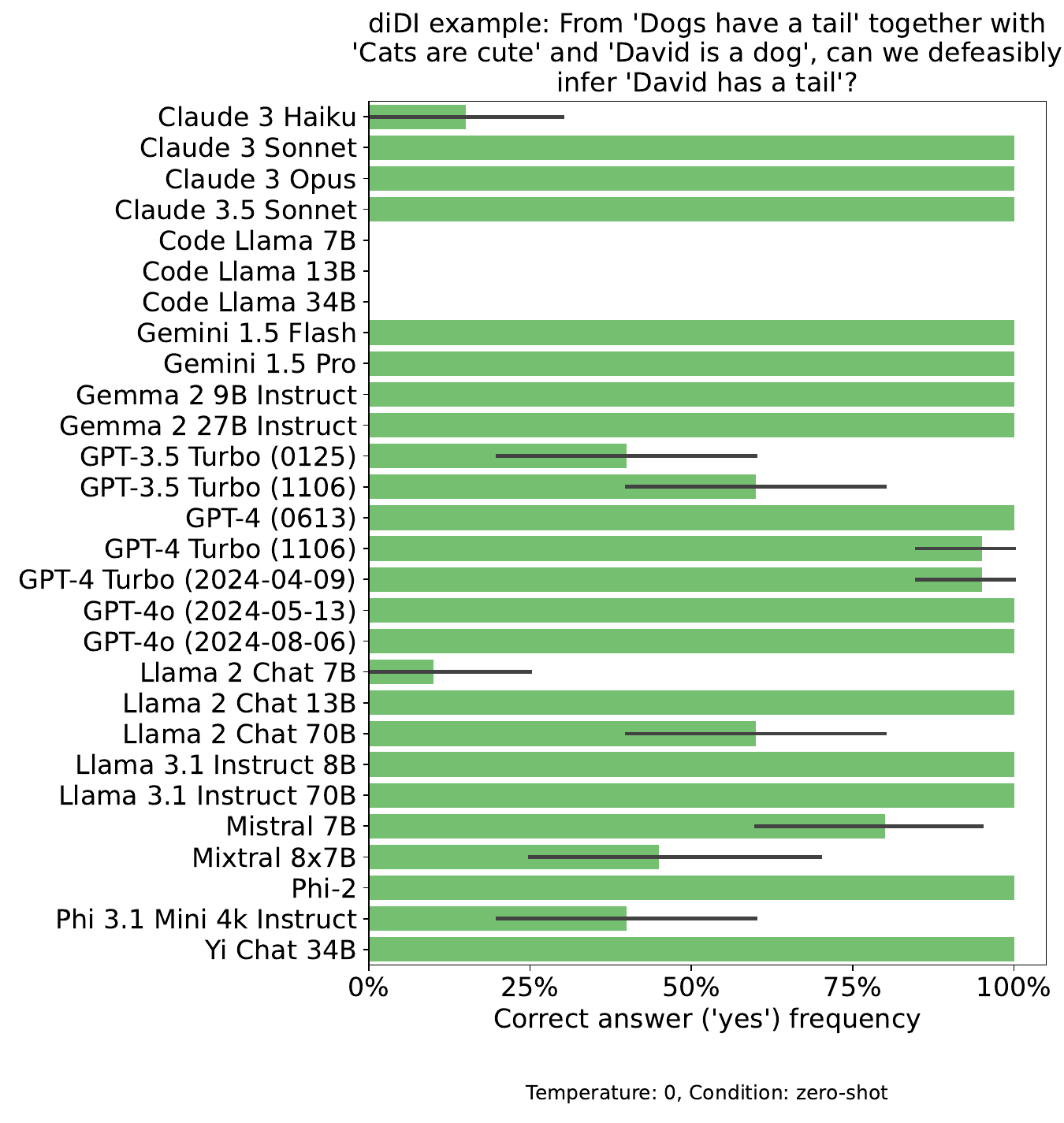}
\includegraphics[scale=.25]{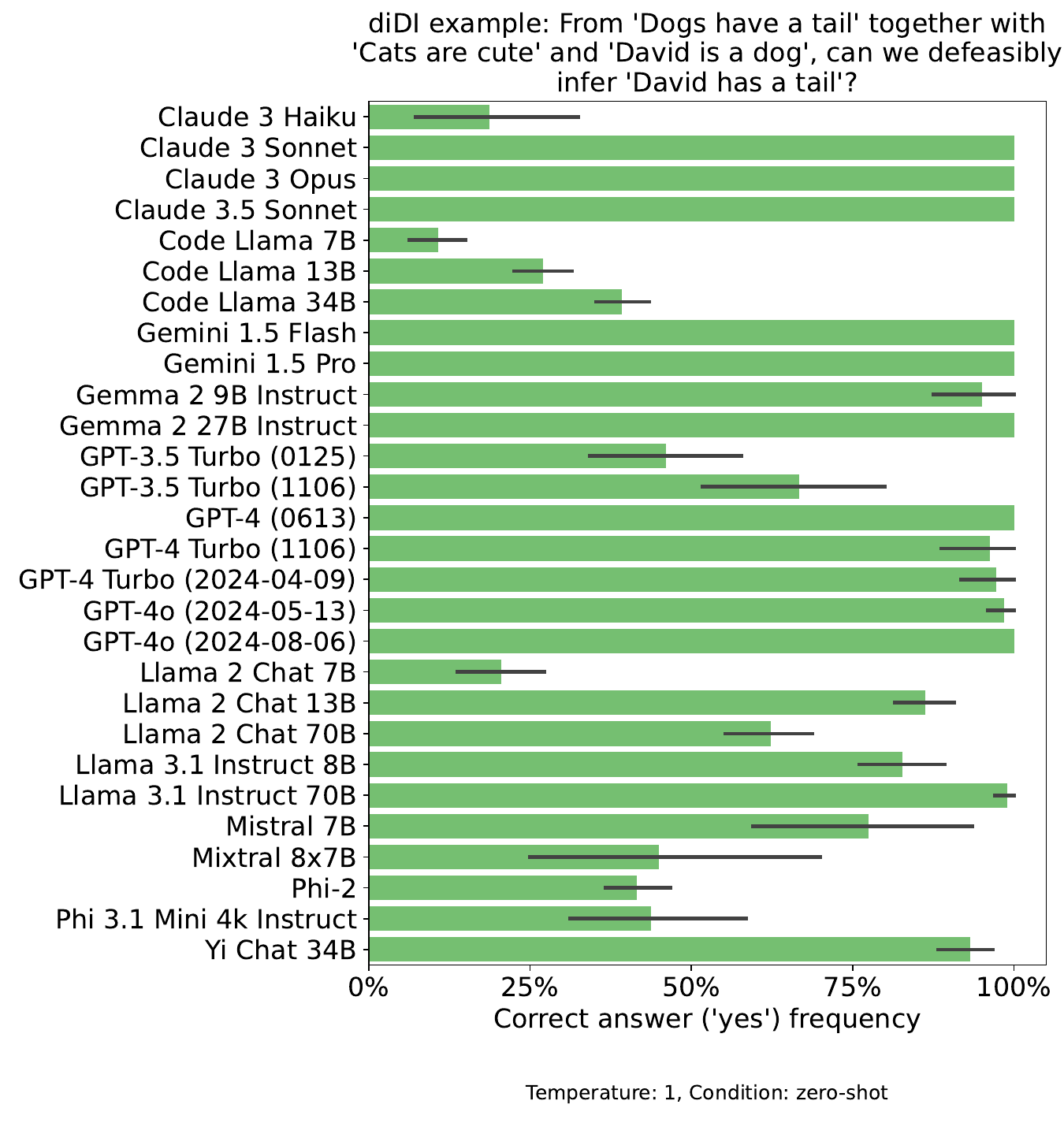}
\end{center}

\subsubsection{Linear Inheritance (LI)}

\begin{center}
\includegraphics[scale=.25]{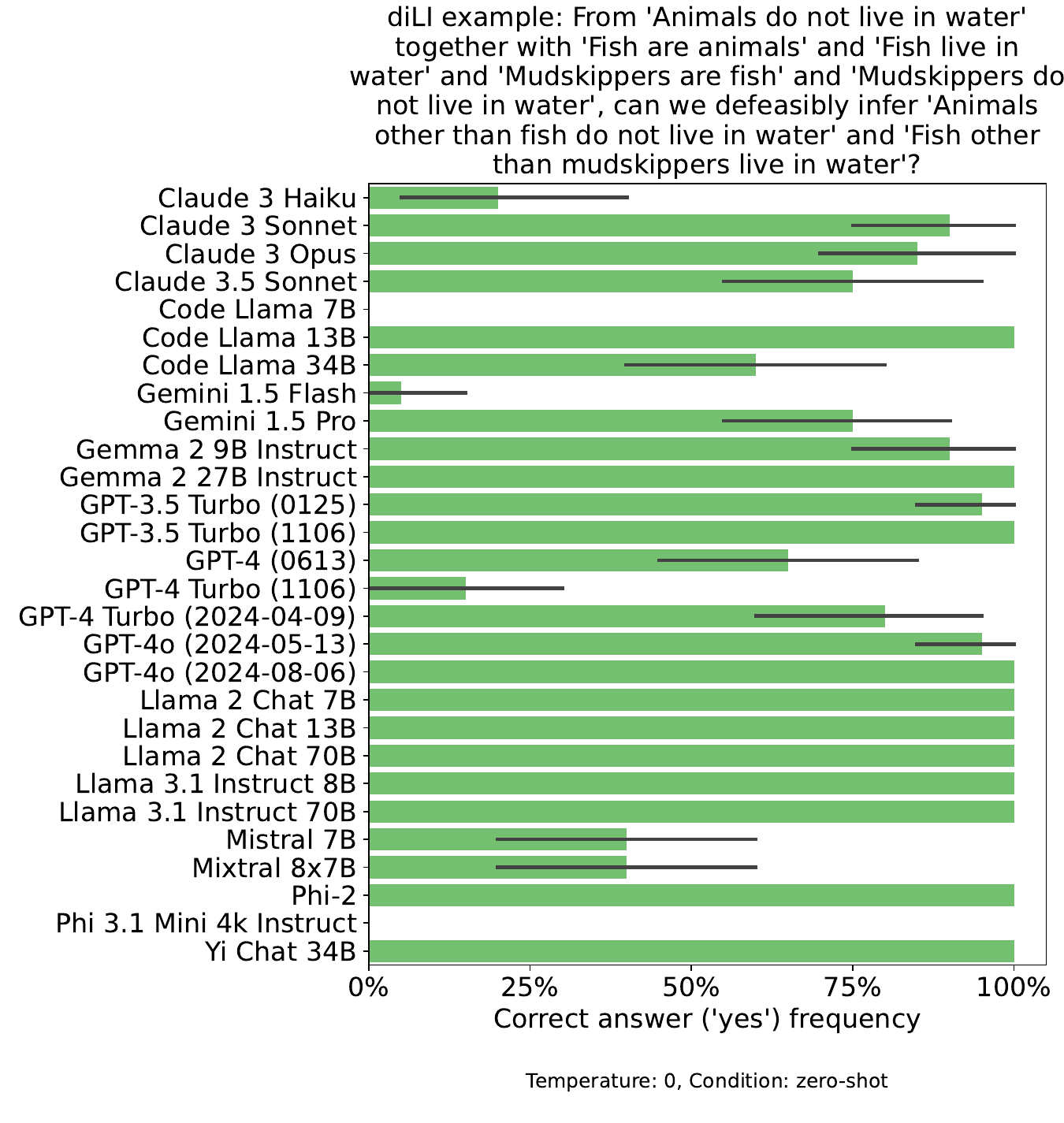}
\includegraphics[scale=.25]{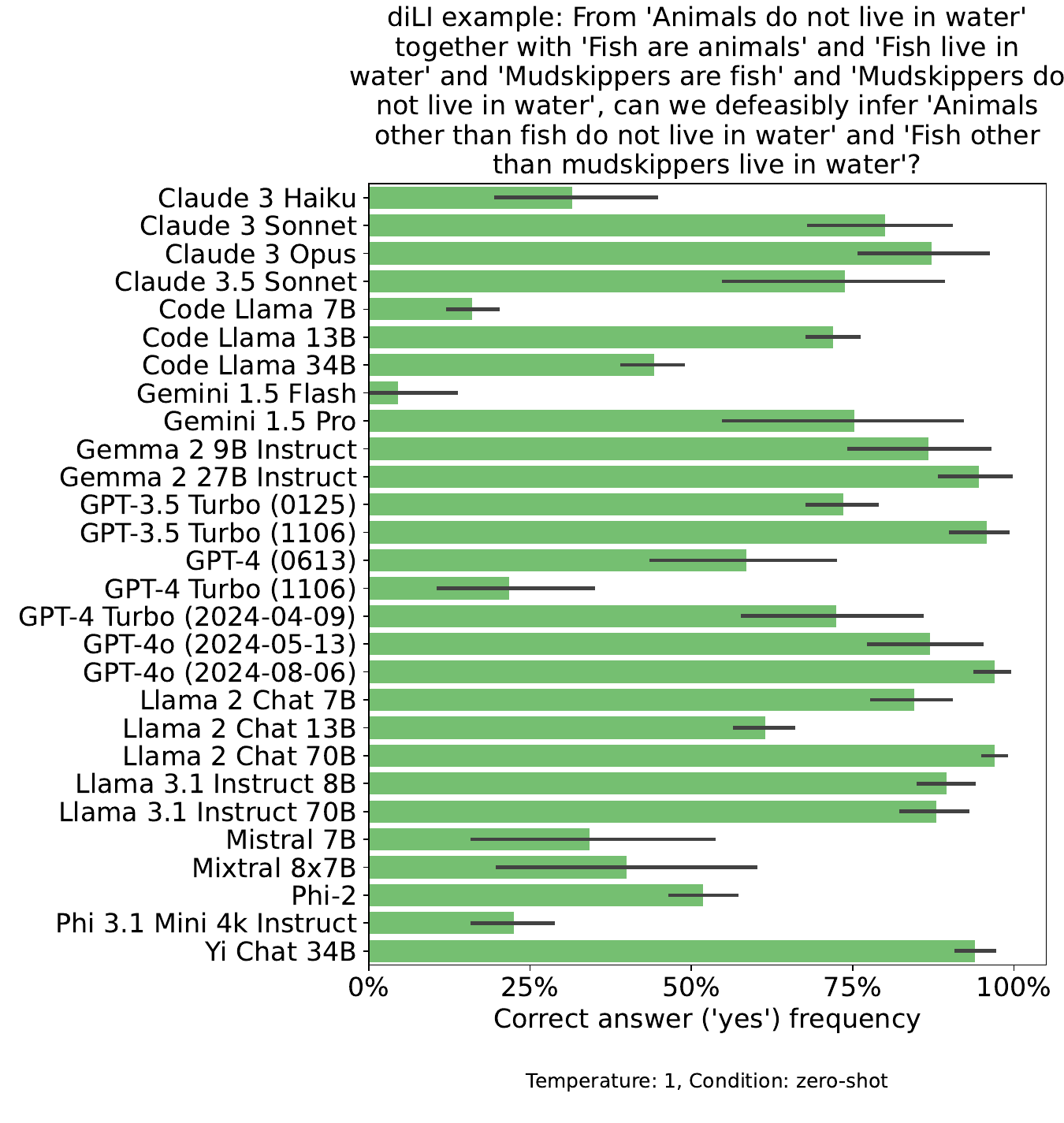}
\end{center}

\subsubsection{Tree-Structure Inheritance (TSI)}

\begin{center}
\includegraphics[scale=.24]{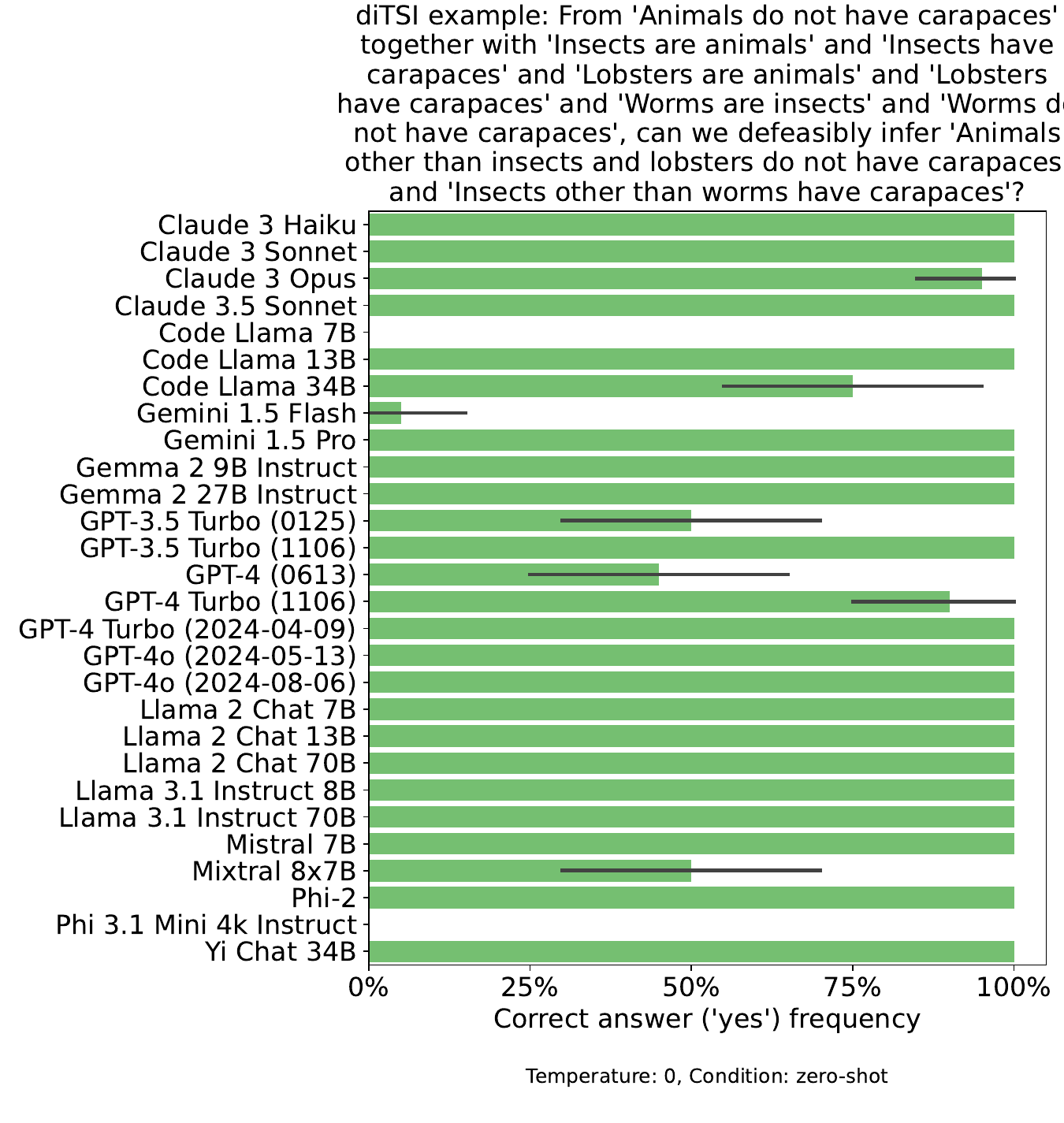}
\includegraphics[scale=.24]{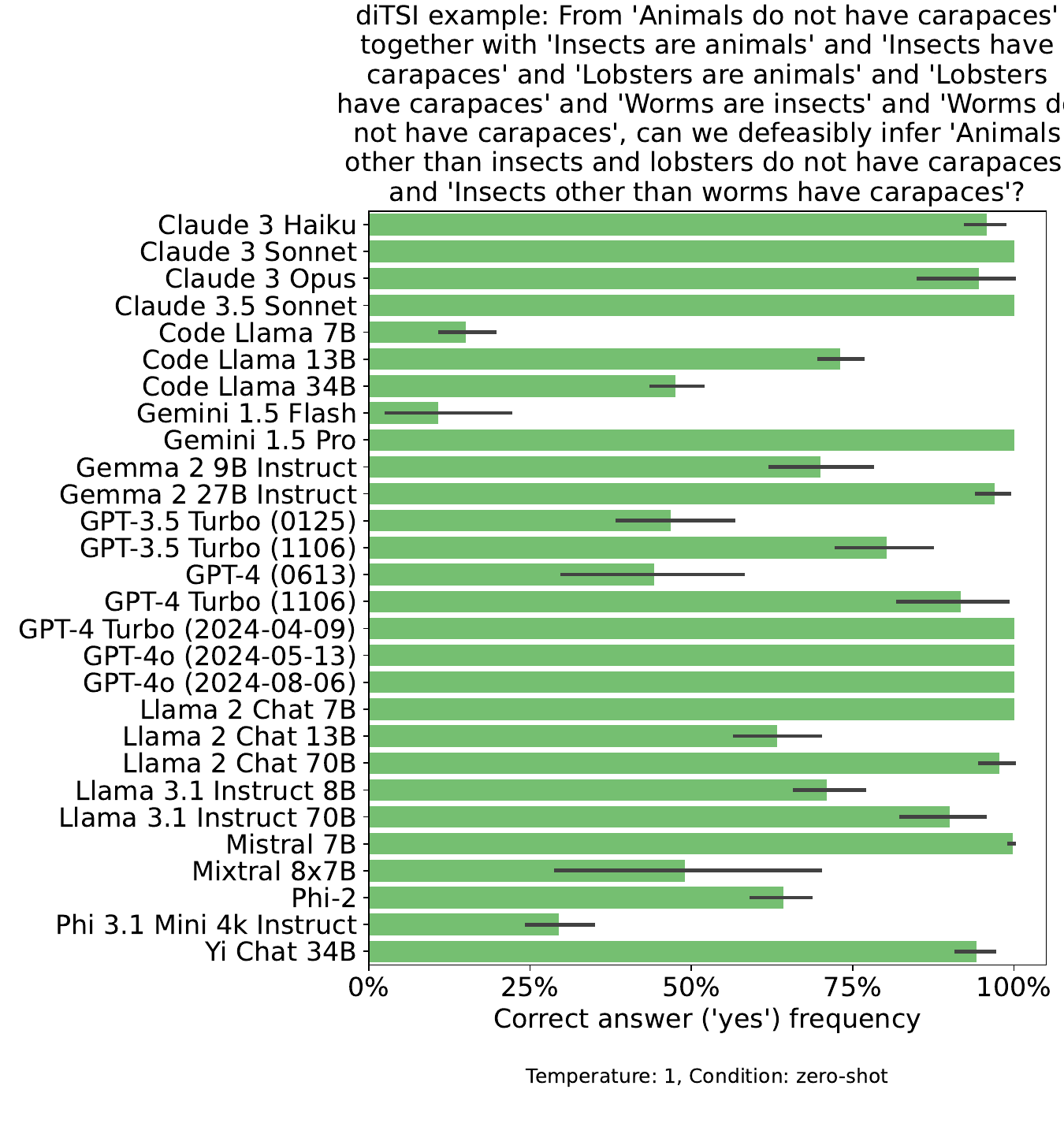}
\end{center}

\subsubsection{One-Step Multiple Inheritance (OSMI)}

\begin{center}
\includegraphics[scale=.24]{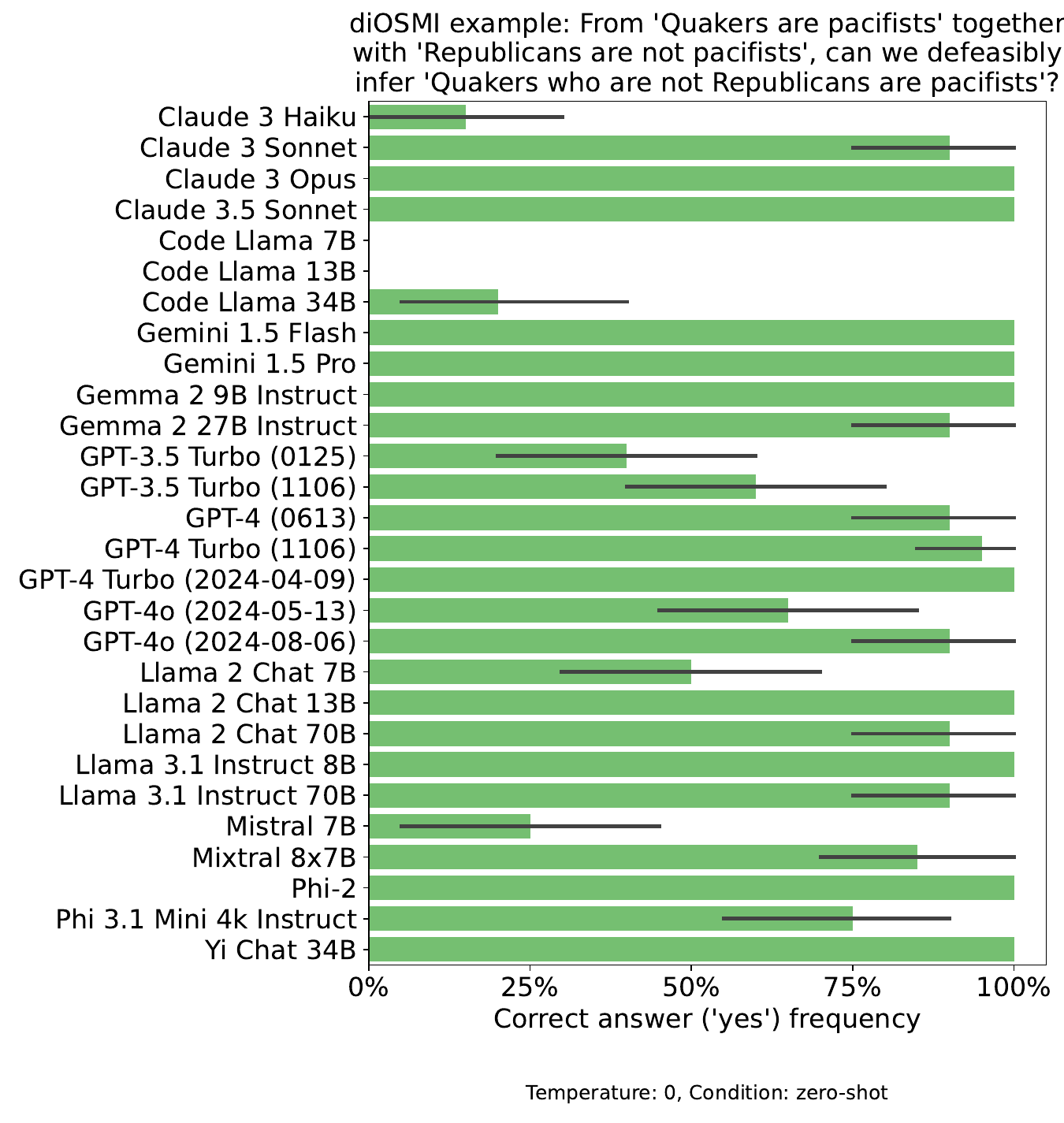}
\includegraphics[scale=.24]{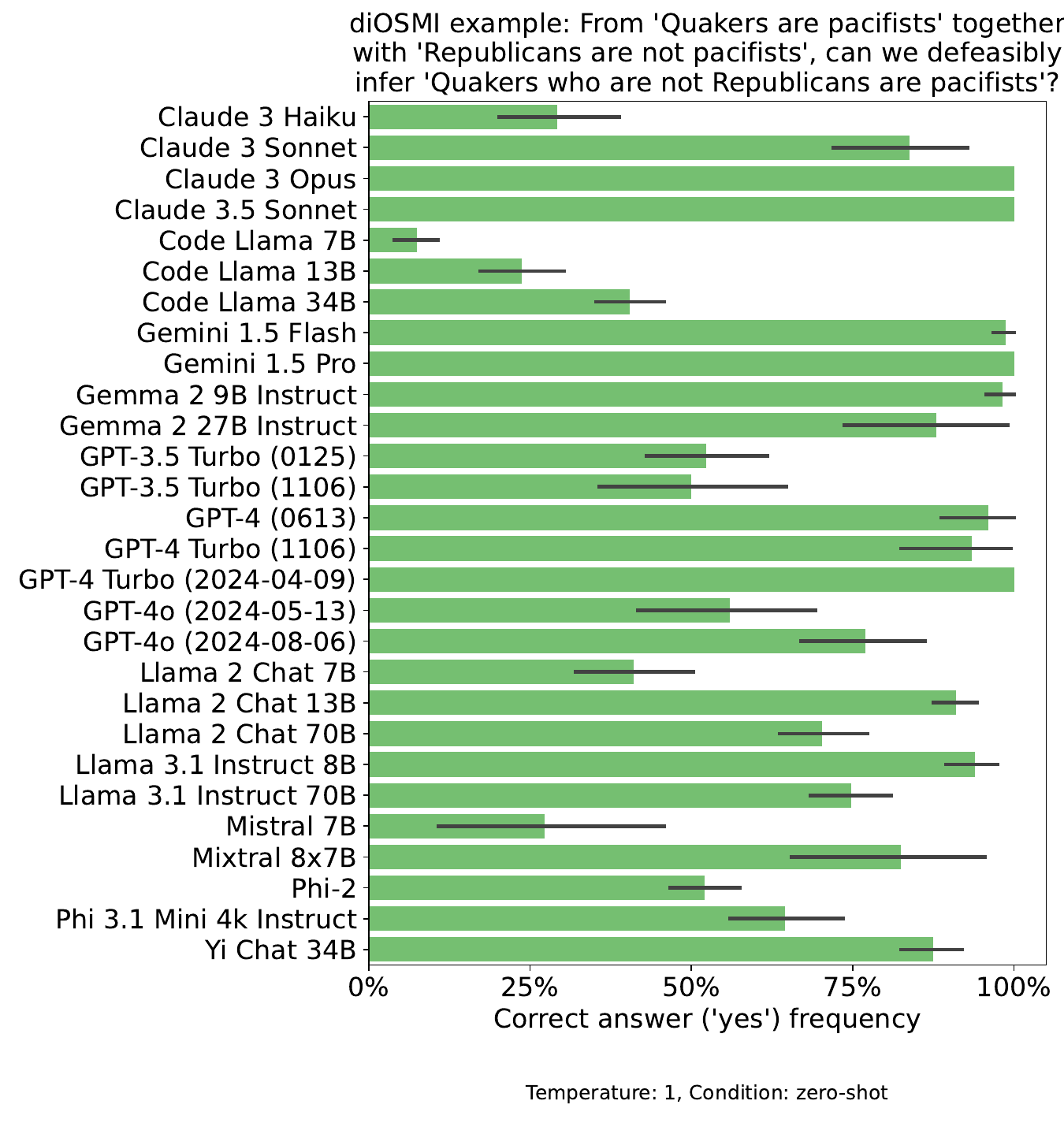}
\end{center}

\subsubsection{One-Step Multiple Inheritance Negative (OSMIn)}

\begin{center}
\includegraphics[scale=.24]{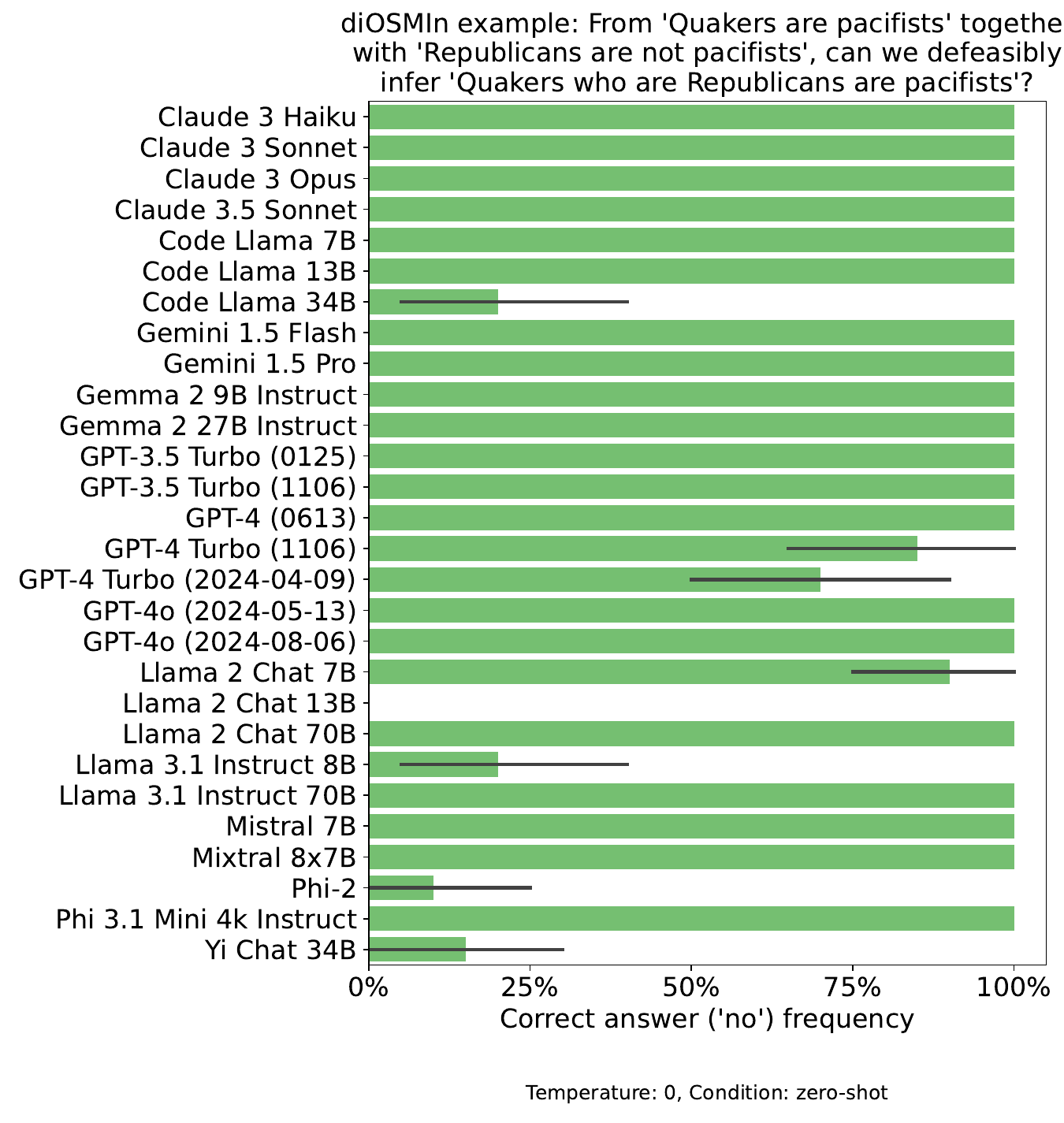}
\includegraphics[scale=.24]{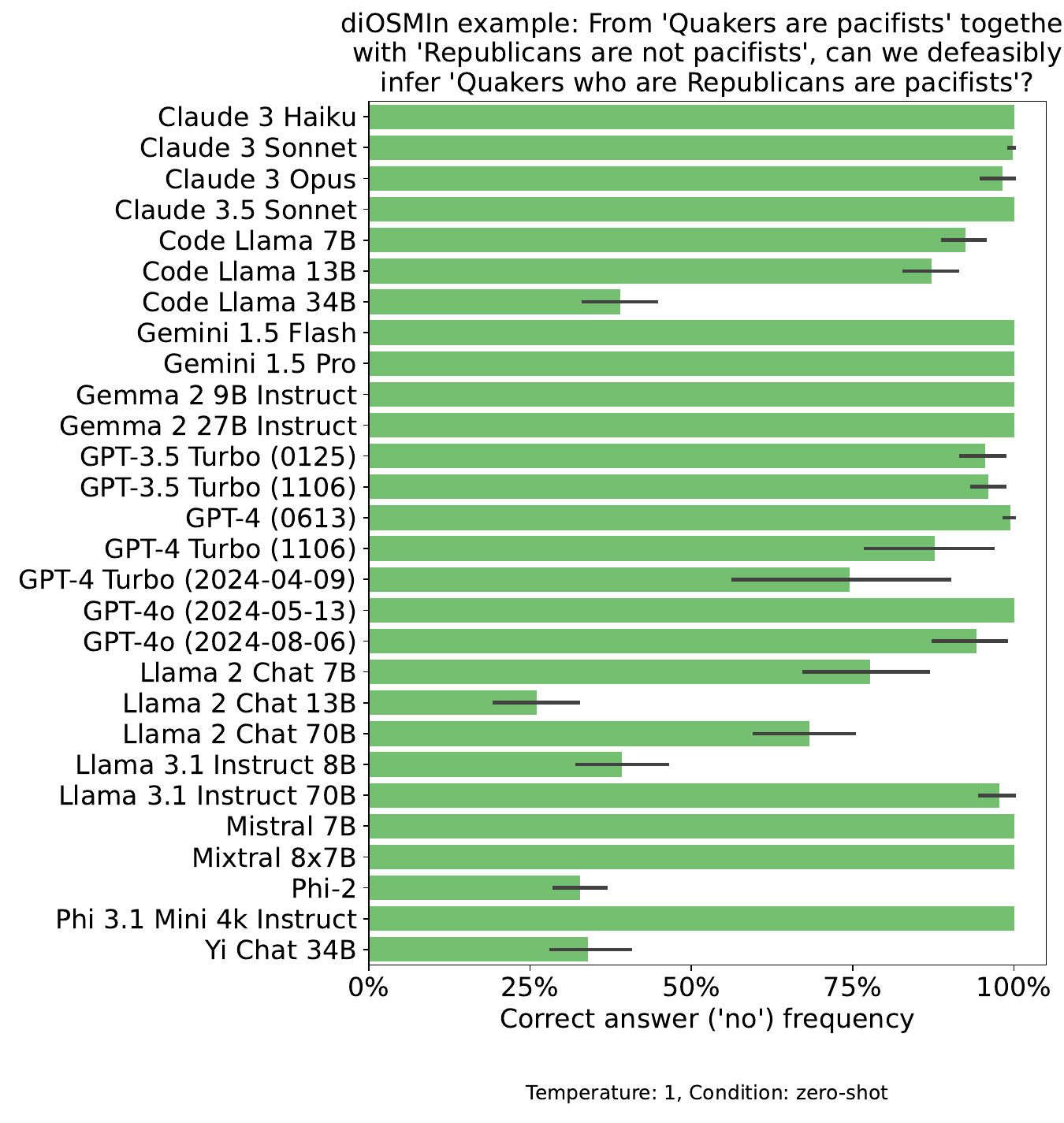}
\end{center}

\subsubsection{Multiple Inheritance (MI)}

\begin{center}
\includegraphics[scale=.24]{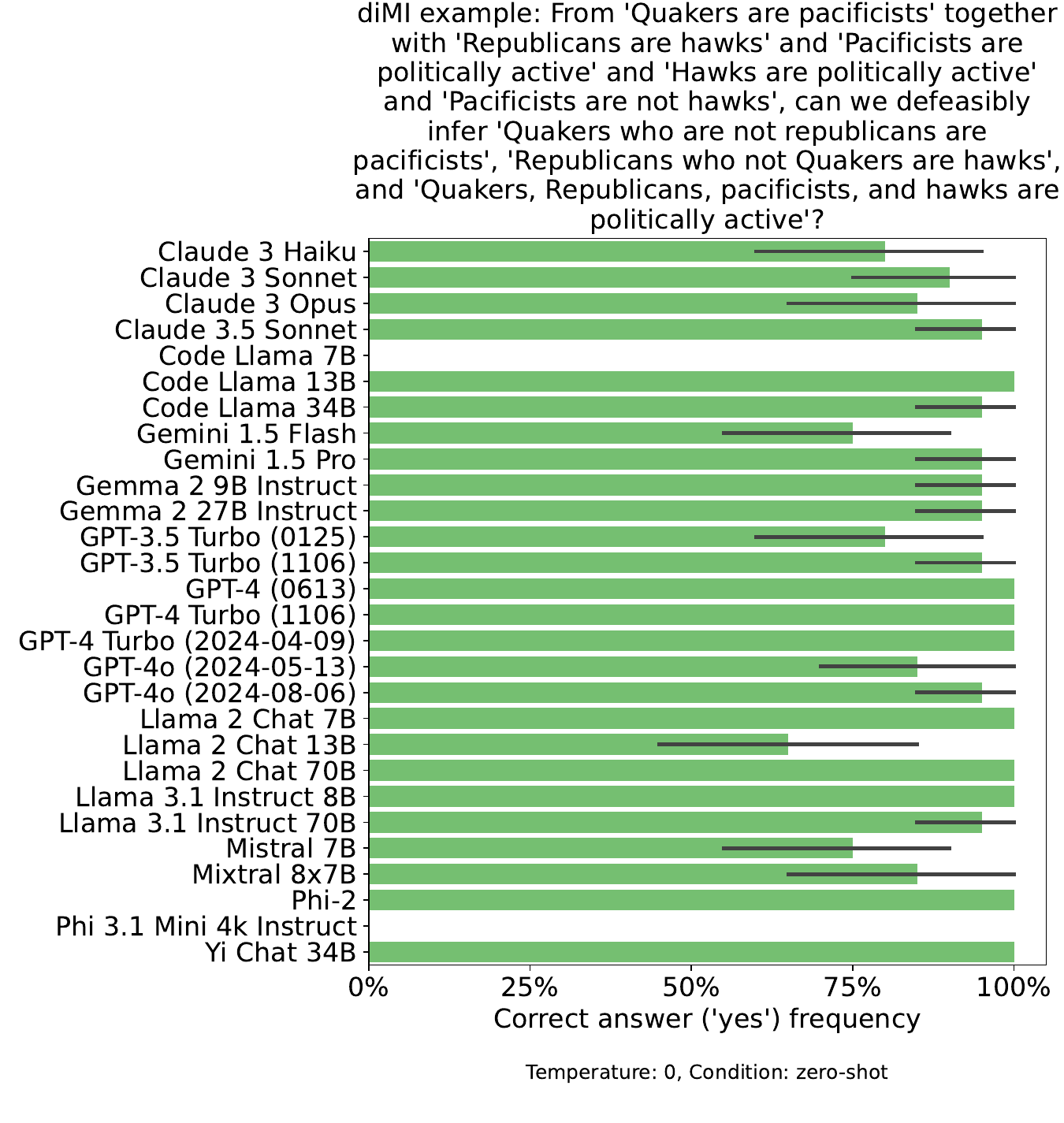}
\includegraphics[scale=.24]{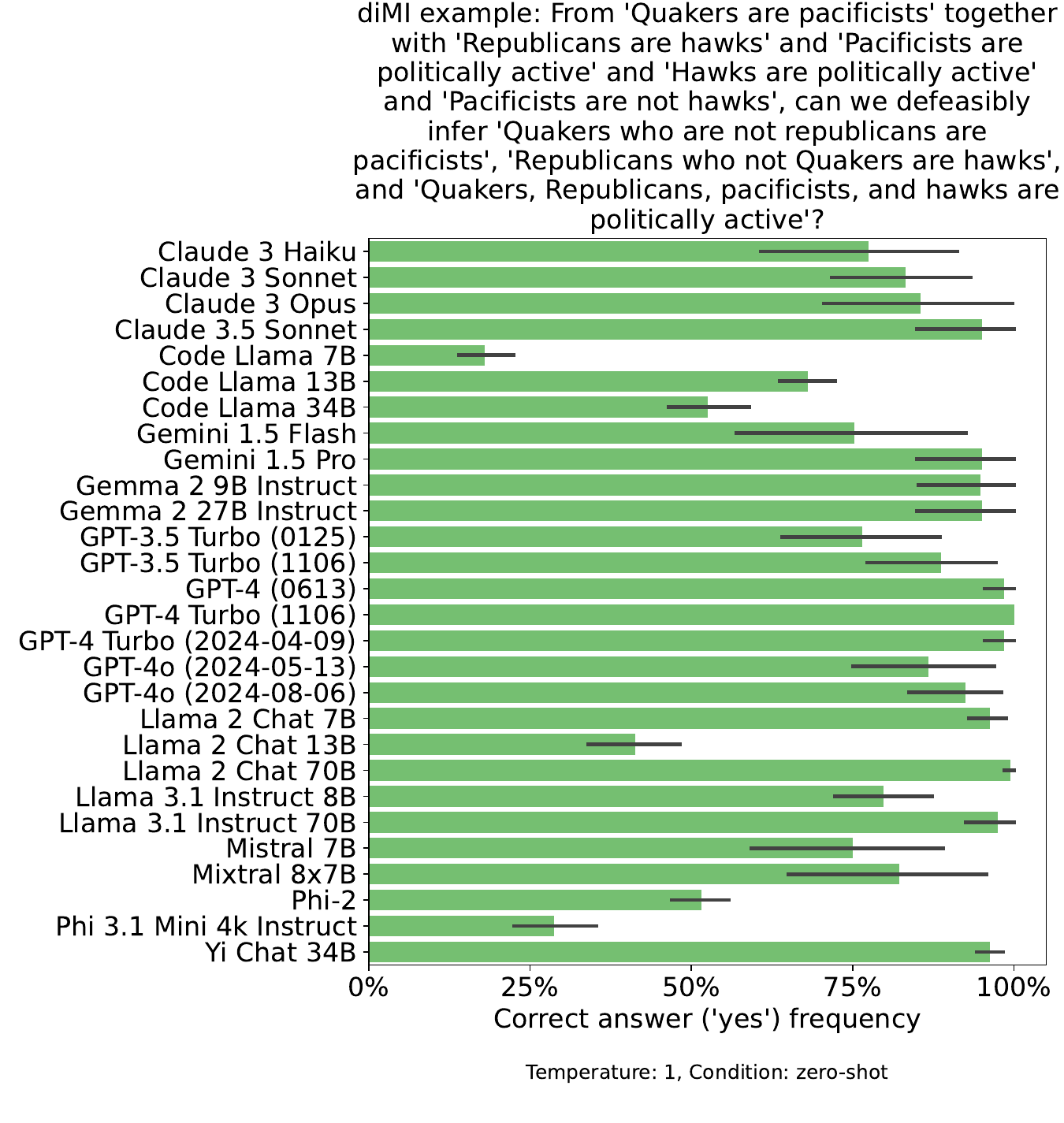}
\end{center}


\subsection{Performance summaries}

\begin{figure}[h!]
\includegraphics[scale=.23]{graphs/zero-shot/summary_graph_multicolor_temp_0_zero-shot.pdf}
\includegraphics[scale=.23]{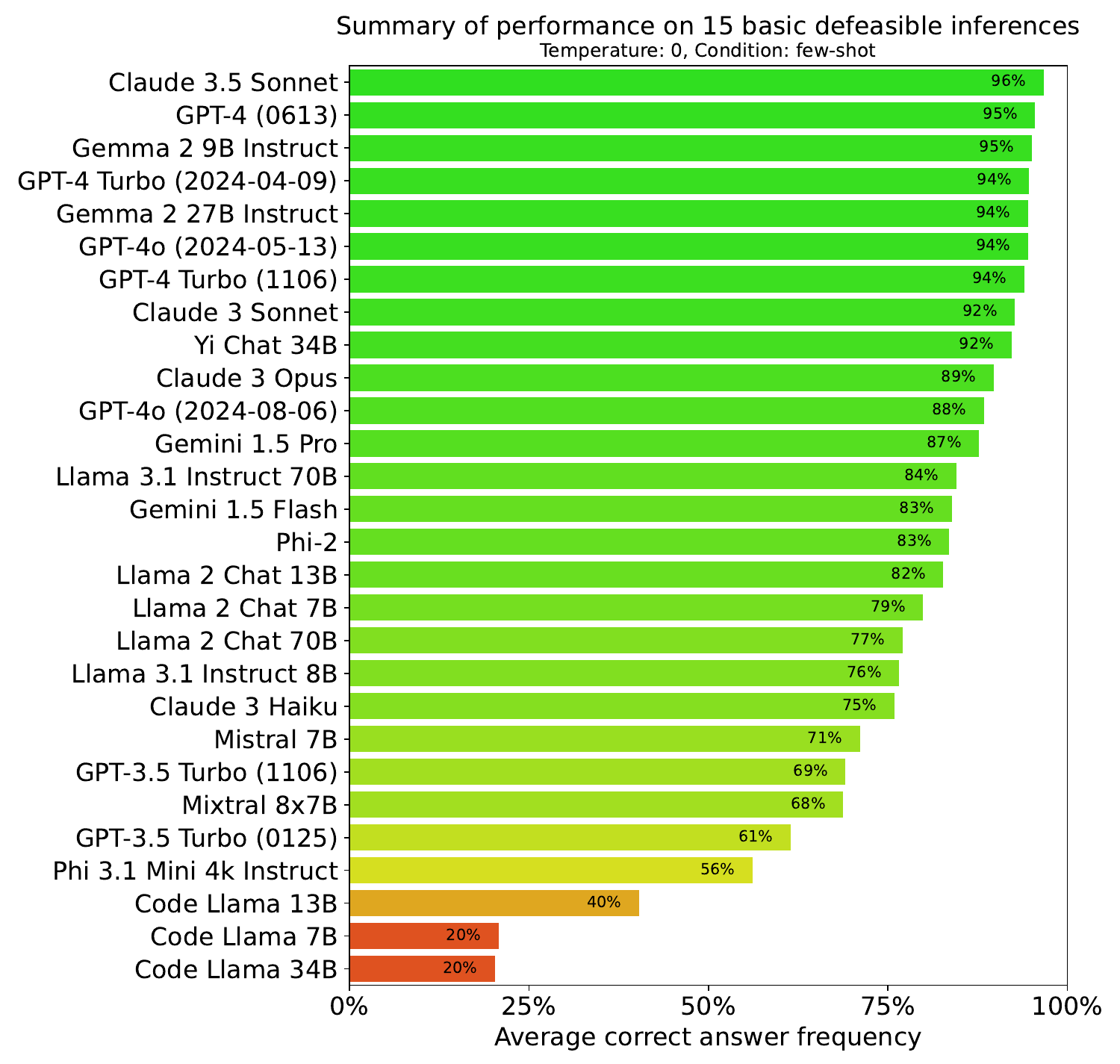}
\includegraphics[scale=.23]{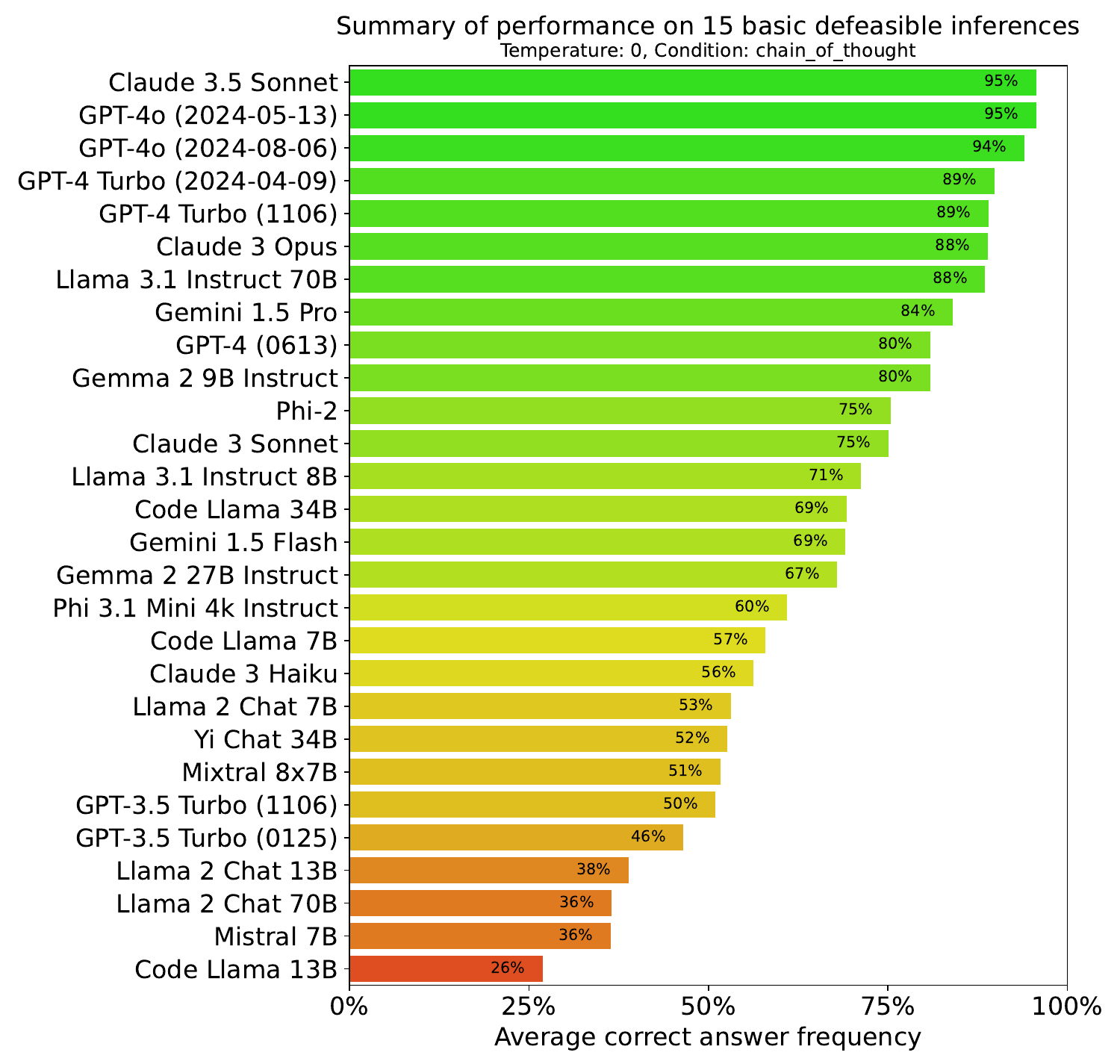}
\caption{Summary of performance on the 15 uncontroversial defeasible inference patterns under different conditions and temperature 0.}

\end{figure}

\begin{figure}[t!]
\includegraphics[scale=.25]{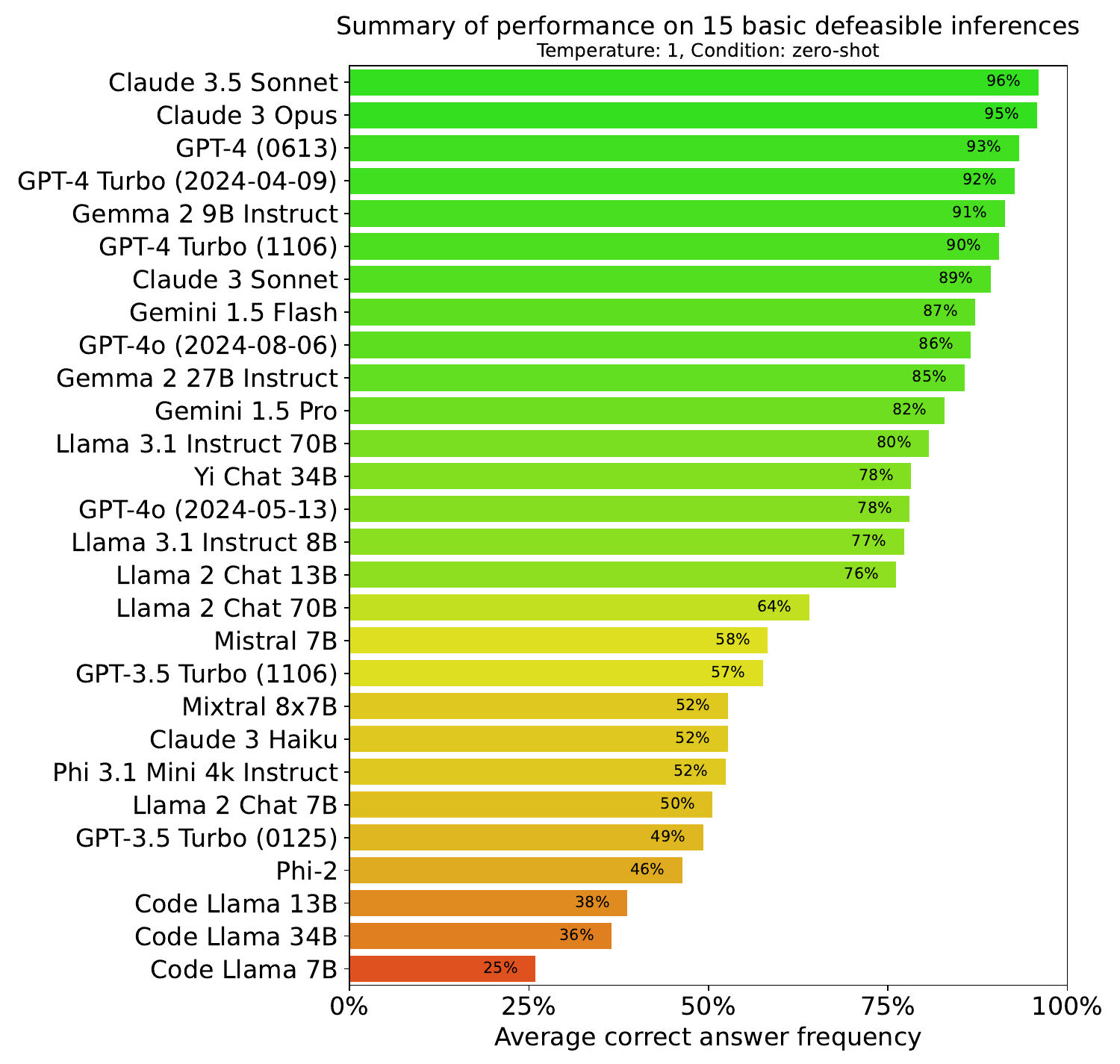}
\includegraphics[scale=.25]{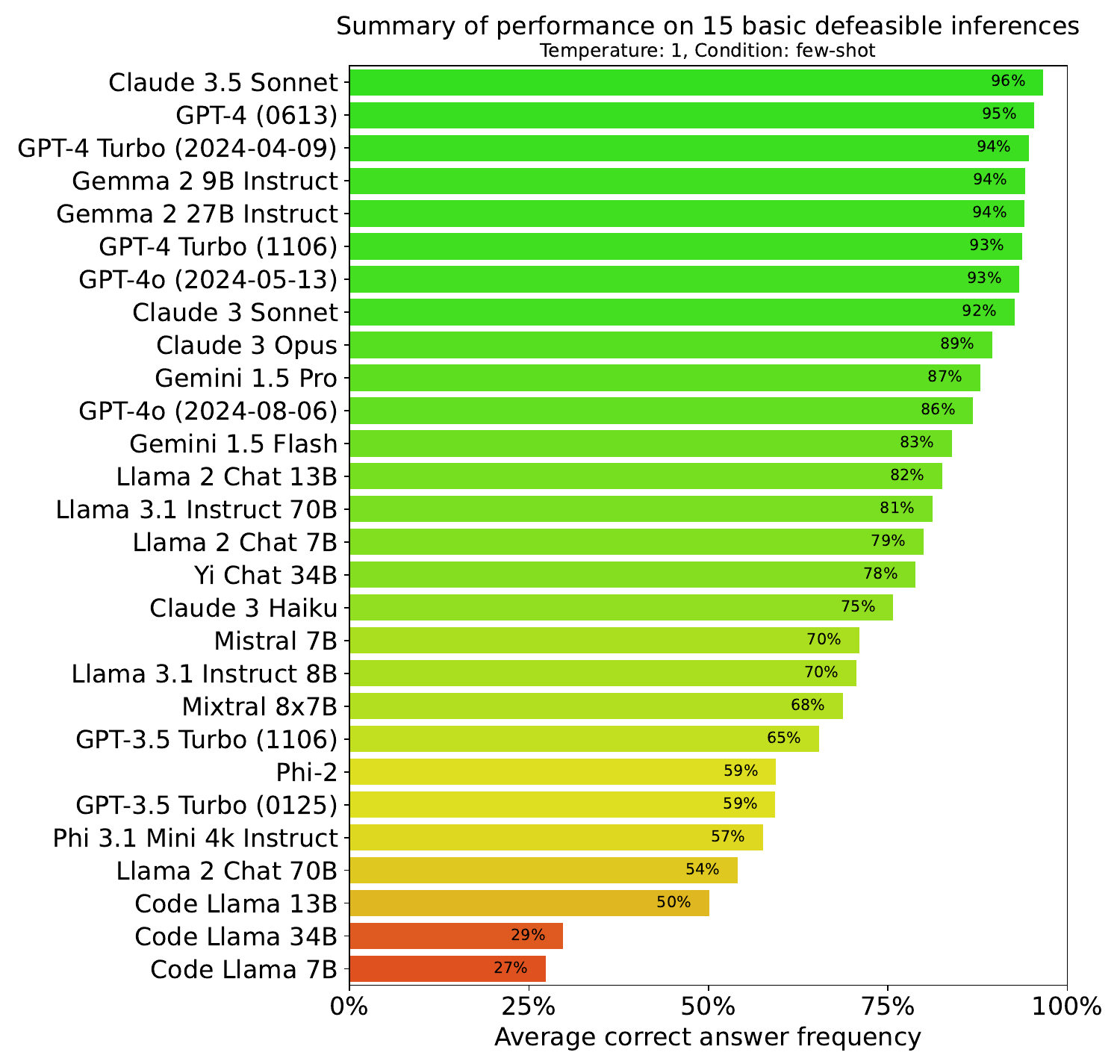}
\caption{Summary of performance on 15 uncontroversial defeasible inference patterns under zero-shot and few-shot conditions with temperature 1.}
\end{figure}

\end{document}